\documentclass{article}
\usepackage[accepted]{icml2021}
\usepackage{amsmath,amsthm, amssymb,mathtools}
\usepackage{caption}

\usepackage{lineno}
\usepackage{microtype}
\usepackage{graphicx}
\usepackage{booktabs} 
\usepackage{hyperref}
\hypersetup{
    colorlinks=true,
    linkcolor=blue,
    filecolor=magenta,       
    urlcolor=cyan,
    pdfpagemode=FullScreen,
    }
\usepackage{cleveref}
\usepackage{physics}
\usepackage{algorithm}
\usepackage{algorithmic} 
\usepackage[caption=false]{subfig} 
\usepackage{booktabs} 
\usepackage{graphicx}
\usepackage[table,xcdraw]{xcolor}
\usepackage{wrapfig}

\usepackage[font=small,skip=3pt]{caption}
\usepackage{titlesec}
\titlespacing*{\section}
{0pt}{0.0ex plus .1ex minus .2ex}{0.0ex plus .2ex}
\titlespacing*{\subsection}
{0pt}{0.0ex plus .1ex minus .2ex}{0.0ex plus .2ex}

\usepackage[utf8]{inputenc}
\usepackage{graphicx}
\usepackage{amsthm}
\usepackage[inline]{enumitem}

\colorlet{myBlue}{blue!60!black}

\usepackage{soul} 
\usepackage{hyperref}
\hypersetup{ 
    colorlinks=true,
    linkcolor=blue, 
    filecolor=magenta,        
    urlcolor=cyan,
    citecolor=myBlue
}
\usepackage{amsmath, amssymb, mathtools, dsfont} 
\usepackage{cleveref} 
\crefname{assumption}{assumption}{assumptions}
\usepackage{bm}

\newtheorem{corollary}{Corollary}
\newtheorem{thm}{Theorem}
\newtheorem{lemma}{Lemma} 
\newtheorem{remark}{Remark}
\newtheorem{proposition}{Proposition}
\newtheorem{definition}{Definition}

\usepackage{tikz}
\usetikzlibrary{positioning,shapes, arrows}
\tikzset{events/.style={ellipse, draw, align=center},}

\usepackage{lipsum}

\usepackage[textwidth=80pt]{todonotes}
\usepackage[ruled,lined, boxed, linesnumbered,commentsnumbered,algo2e]{algorithm2e}    

\SetCommentSty{mycommfont}

\newtheoremstyle{styledef}
  {2pt}
  {0pt}
  {}
  {0em}
  {\bfseries}
  {}
  {1em}
  {}
\theoremstyle{styledef}

\newcommand{\E}{{{\mathbb E}}}
\newcommand{\R}{{{\mathbb R}}}

\newcommand{\cH}{{{\mathcal H}}}
\newcommand{\X}{{{\mathcal X}}}
\newcommand{\Y}{{{\mathcal Y}}}
\newcommand{\Z}{{{\mathcal Z}}}
\newcommand{\W}{{{\mathcal W}}}
\newcommand{\A}{{{\mathcal A}}}
\newcommand{\cA}{{{\mathcal A}}}
\newcommand{\cB}{{{\mathcal B}}}
\newcommand{\cN}{{{\mathcal N}}}

\newcommand{\cL}{{{\mathcal L}}}

\newcommand{\cP}{{{\mathcal P}}}

\newcommand{\tS}{{{\tilde{S}}}}

\newcommand{\tx}{{{\widetilde{x}}}}
\newcommand{\ty}{{{\widetilde{y}}}}
\newcommand{\tz}{{{\widetilde{z}}}}
\newcommand{\ta}{{{\widetilde{a}}}}

\newcommand{\tA}{{{\widetilde{A}}}}
\newcommand{\tZ}{{{\widetilde{Z}}}}

\newcommand{\tX}{{{\widetilde{X}}}}

\newcommand{\Hx}{{{\mathcal H}_{\X}}}

\newcommand{\Hz}{{{\mathcal H}_{\Z}}}
\newcommand{\Hw}{{{\mathcal H}_{\W}}}
\newcommand{\Ha}{{{\mathcal H}_{\A}}}

\newcommand{\kx}{k}
\newcommand{\ka}{k}
\newcommand{\kz}{k}
\newcommand{\kw}{k}

\newcommand{\phix}{{\phi}}
\newcommand{\phia}{\phi}
\newcommand{\phiz}{\phi}
\newcommand{\phiw}{\phi}

\newcommand{\Phix}{\ensuremath {\Phi_{\X}}}
\newcommand{\Phia}{\Phi}
\newcommand{\Phiw}{\Psi}
\newcommand{\Phiz}{\Upsilon}

\makeatletter
\newcommand*\bigcdot{\mathpalette\bigcdot@{.5}}
\newcommand*\bigcdot@[2]{\mathbin{\vcenter{\hbox{\scalebox{#2}{$\m@th#1\bullet$}}}}}
\makeatother

\newcommand{\ps}[1]{\langle #1 \rangle}
\newcommand{\indep}{\perp \!\!\! \perp}
\newcommand{\eqdef}{=\vcentcolon}
\newcommand{\ltwoawx}{\mathcal{L}^2(\mathcal{A\cross W \cross X}, \mathcal{P_{AWX}})}

\newtheorem{assumption}{Assumption}

\DeclareMathOperator*{\U}{\mathcal{U}}
\DeclareMathOperator*{\argmin}{argmin}

\icmltitlerunning{Proximal Causal Learning with Kernels: Two-Stage Estimation and Moment Restriction}

\begin{document}

\twocolumn[
\icmltitle{Proximal Causal Learning with Kernels: \\ Two-Stage Estimation and Moment Restriction}
\icmlsetsymbol{equal}{*}
\icmlsetsymbol{equal2}{$\dagger$}
\begin{icmlauthorlist}
\icmlauthor{Afsaneh Mastouri}{equal,ucl}
\icmlauthor{Yuchen Zhu}{equal,ucl}
\icmlauthor{Limor Gultchin}{ox,turing}
\icmlauthor{Anna Korba}{cres}
\icmlauthor{Ricardo Silva}{ucl}
\icmlauthor{Matt J. Kusner}{ucl}
\icmlauthor{Arthur Gretton}{equal2,ucl}
\icmlauthor{Krikamol Muandet}{equal2,tub}
\end{icmlauthorlist}

\icmlaffiliation{ucl}{University College London, London, United Kingdom}
\icmlaffiliation{ox}{University of Oxford, Oxford, United Kingdom}
\icmlaffiliation{tub}{Max Planck Institute for Intelligent Systems, T\"ubingen, Germany}
\icmlaffiliation{turing}{The Alan Turing Institute, London, United Kingdom}
\icmlaffiliation{cres}{ENSAE/CREST, Paris, France}

\icmlcorrespondingauthor{Afsaneh Mastouri}{afs.mastouri@gmail.com}
\icmlcorrespondingauthor{Yuchen Zhu}{yuchen.zhu.18@ucl.ac.uk}

\icmlkeywords{Machine Learning, ICML}

\vskip 0.3in
]
\printAffiliationsAndNotice{\icmlEqualContribution \icmlSeniorEqualContribution}



\begin{abstract}
We address the problem of causal effect estimation in the presence of unobserved confounding, but where proxies for the latent confounder(s) are observed. We propose two kernel-based methods for nonlinear causal effect estimation in this setting: (a) a two-stage regression approach, and (b) a maximum moment restriction approach. 
We focus on the proximal causal learning setting, but our methods can be used to solve a wider class of inverse problems characterised by a Fredholm integral equation. In particular, we provide a unifying view of two-stage and moment restriction approaches for solving this problem in a nonlinear setting. We provide consistency guarantees for each algorithm, and  demonstrate that these approaches achieve competitive results on synthetic data and data simulating a real-world task. In particular, our approach outperforms earlier methods that are not suited to leveraging proxy variables.
\end{abstract}

\linepenalty=1000

\section{Introduction}

Estimating average treatment effects (ATEs) is critical to answering many scientific questions. From estimating the effects of medical treatments on patient outcomes \cite{connors1996effectiveness,choi2002methotrexate}, to grade retention on cognitive development \cite{fruehwirth2016timing}, 
ATEs are the key estimands of interest. From observational data alone, however, estimating such effects is impossible without further assumptions. This impossibility arises from potential unobserved confounding: one variable may seem to cause another, but this could be due entirely to an unobserved variable causing both of them, e.g., as was used by the tobacco industry to argue against the causal link between smoking and lung cancer \cite{cornfield1959smoking}.

\setlength\abovecaptionskip{0pt}
\begin{figure}[t!]
    \centering
    \includegraphics[width=0.68\columnwidth]{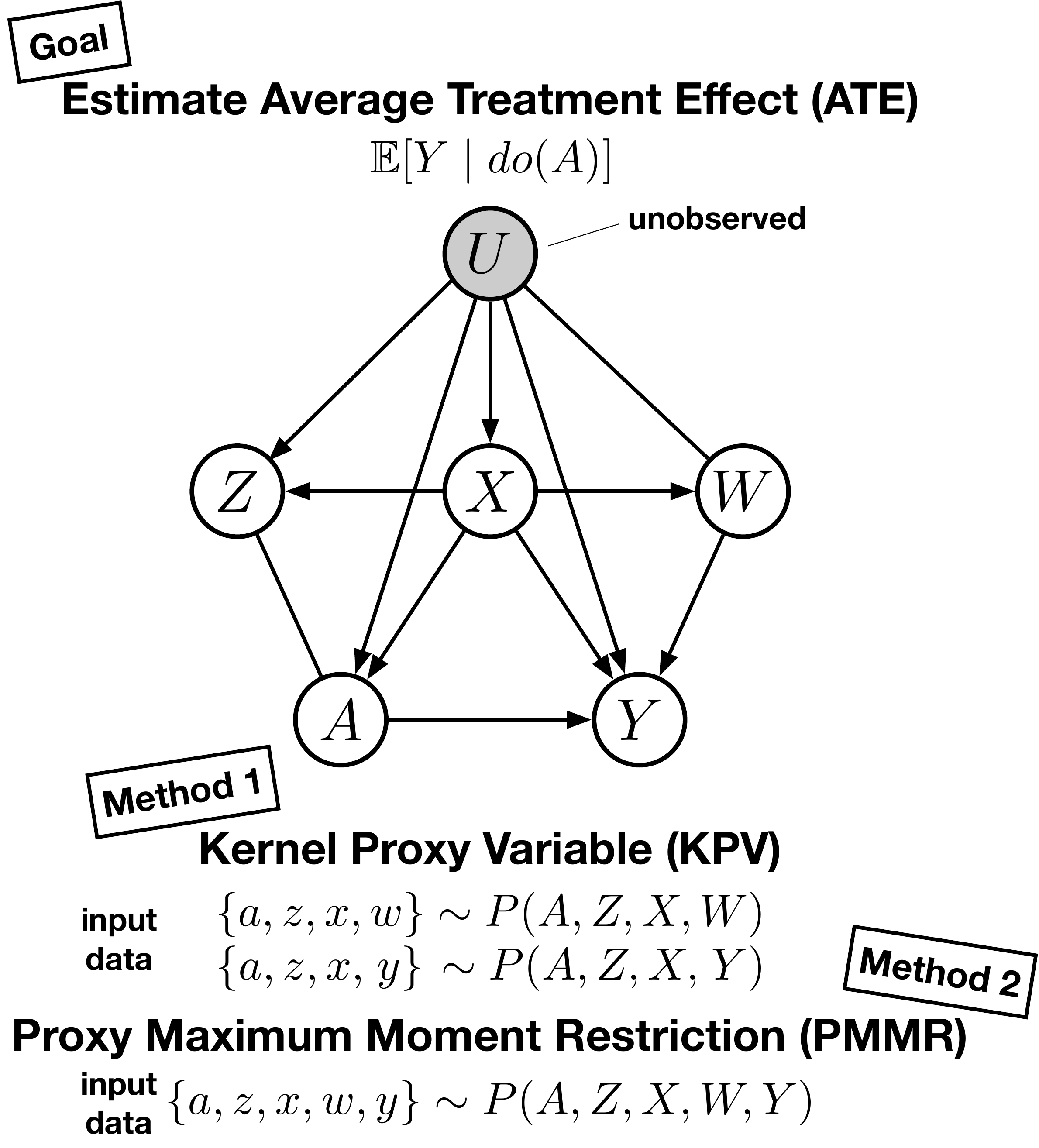}
    \caption{The causal proxy estimation problem, and two methods we introduce to solve it.}
    \label{fig:overall}
    \vspace{-1ex}
    \setlength{\belowcaptionskip}{-10pt}
\end{figure}  
\setlength{\textfloatsep}{8pt}

One of the most common assumptions to bypass this difficulty is to assume that no unobserved confounders exist \cite{imbens2004nonparametric}. This extremely restrictive assumption makes estimation easy: if there are also no observed confounders then the ATE can be estimated using simple regression, otherwise one can use backdoor adjustment \cite{pearl2009causality}. Less restrictive is to assume observation of an \emph{instrumental variable} (IV) that is independent of any unobserved confounders \cite{reiersol1945confluence}. This independence assumption is often broken, however. For example, if medication requires payment, many potential instruments such as educational attainment will be confounded with the outcome through complex socioeconomic factors, which can be difficult to fully observe (e.g., different opportunities afforded by living in different neighborhoods). The same argument regarding unobserved confounding can be made for the grade retention and household expenditure settings.

This fundamental difficulty has inspired work to investigate the relaxation of this independence assumption. An increasingly popular class of models, called \emph{proxy} models, does just this; and  various recent studies incorporate proxies into causal discovery and inference tasks to reduce the influence of confounding bias  \cite{cai2012identifying, tchetgen2014control, schuemie2014interpreting, sofer2016negative, flanders2017new, shi2018multiply}.
Consider the following example from \citet{deaner2021proxy}, described graphically in Figure~\ref{fig:overall}: we wish to understand the effect (i.e., ATE) of holding children back a grade in school (also called `grade retention') $A$ , on their math scores
, $Y$. This relationship is confounded by an unobserved variable $U$ describing students' willingness to learn in school. Luckily we have access to a \emph{proxy} of $U$, student scores from a cognitive and behavioral test $W$. Note that if $W\!=\!U$ we could use backdoor adjustment to estimate the effect of $A$ on $Y$ \cite{pearl2009causality}. In general $W\!\neq\!U,$ however, and this adjustment would produce a biased estimate of the ATE. In this case we can introduce a second proxy $Z$: the cognitive and behavioral test result \emph{after} grade retention $A$. This allows us to form an integral equation similar to the IV setting. The solution to this equation is not the ATE (as it is in the IV case) but a function that, when adjusted over the distribution $P(W)$ (or $P(W,X)$ in the general case) gives the true causal effect \cite{kuroki2014measurement,tchetgen2020introduction}. 
Building on \citet{carroll2006measurement} and \citet{greenland2011bias}, \citet{kuroki2014measurement} were the first to demonstrate the possibility of identifying the causal effect given access to proxy variables. This was generalized by \citet{miao2018confounding}, and \citet{tchetgen2020introduction} recently proved non-parametric identifiability for the general proxy graph (i.e., including $X$) shown in Figure~\ref{fig:overall}. 

The question of how to \emph{estimate} the ATE in this graph for continuous variables is still largely unexplored, however,  particularly in a non-linear setting, and with consistency guarantees. \citet{deaner2021proxy} assume a sieve basis and describe a technique to identify a different causal quantity, the average treatment effect on the treated (ATT), in this graph (without $X$, but the work can be easily extended to include $X$). \citet{tchetgen2020introduction} assume linearity and estimate the ATE. The linearity assumption significantly simplifies estimation, but the ATE in principle can be identified without parametric assumptions \cite{tchetgen2020introduction}. At the same time, there have been exciting developments in using kernel methods to estimate causal effects in the non-linear IV setting, with consistency guarantees \cite{singh2019kernel,Muandet19:DualIV,zhang2020maximum}.
Kernel approaches to ATE, ATT, Conditional ATE, and causal effect estimation under distribution shift, have been explored in various settings \cite{singh2020kernel}.

\citet{singh2020kernel2} has also considered a kernelized proximal setting, using a two-stage least squares approach, however
 the original published method (Algorithm 4.1 in the \href{https://arxiv.org/abs/2012.10315v1}{initial work, December 2020}) is not guaranteed to be consistent, and has related empirical shortcomings: see \Cref{sec:RS_approach} for details.
 \citet{singh2020kernel2}
 subsequently proposed a new algorithm (Algorithm 4.1 in the \href{https://arxiv.org/abs/2012.10315v2}{revised paper, May 2021}, based on \cref{eq:erm2}), although it remains an incomplete instance of the full solution provided by the Representer Theorem: see  \Cref{sec:incomplete} for details.
 A third new algorithm
 (Algorithm 1 in the \href{https://arxiv.org/abs/2012.10315v3}{revised paper, September 2022})
 is different to \citeauthor{singh2020kernel2}'s  previous two algorithms, and provides an alternative implementation to our two-stage approach, as well as covering additional settings such as the Conditional ATE and ATT.
To the best of our knowledge, our approach represents the first  kernel two-stage least squares implementation of proximal causal learning.




In this work, we propose two kernelized estimation procedures for the ATE in the proxy setting, with consistency guarantees: (a) a two-stage regression approach (which we refer to as Kernelized Proxy Variables, or KPV), and (b) a maximum moment restriction approach (which we refer to as Proxy Maximum Moment Restriction, or PMMR). Alongside consistency guarantees, we derive a theoretical connection between both approaches, 
and  show that our methods can also be used to solve a more general class of inverse problems that involve a solution to a Fredholm integral equation. We demonstrate the performance of both approaches on synthetic data, and on data simulating real-world tasks.

\section{Background}

Throughout, a capital letter (e.g. $A$) denotes a random variable on a measurable space, denoted by a calligraphic letter (resp. $\A$). We use lowercase letters to denote the realization of a random variable (e.g. $A=a$). 

\subsection{Causal Inference with Proxy Variables}\label{ch:Problem_setting}

Our goal is to estimate the average treatment effect (ATE) of treatment $A$ on outcome $Y$ in the proxy causal graph of Figure~\ref{fig:overall} (throughout we will assume $Y$ is scalar and continuous; the discrete case is much simpler \cite{miao2018confounding}). To do so, we are given access to proxies of an unobserved confounder $U$: a treatment-inducing proxy $Z$, an outcome-inducing proxy $W$; and optionally observed confounders $X$. 
Formally, given access to samples from either the joint distribution $\rho(A,Z,X,W,Y)$ or from both distributions $\rho(A,Z,X,W)$ and $\rho(A,Z,X,Y)$, we aim to estimate the ATE $\mathbb{E}[Y \mid do(A=a)]$. Throughout we will describe causality using the structural causal model (SCM) formulation of \citet{pearl2009causality}. Here, the causal relationships are represented as directed acyclic graphs. The crucial difference between these models and standard probabilistic graphical models is a new operator: the intervention $do(\cdot)$. This operator describes the process of forcing a random variable to take a particular value, which isolates its effect on downstream variables (i.e., $\mathbb{E}[Y \mid do(A=a)]$ describes the isolated effect of $A$ on $Y$). 
We start by introducing the assumptions that are necessary to identify this causal effect. These assumptions can be divided into two classes: (A) \emph{structural assumptions} and (B) \emph{completeness assumptions}. 

(A) \textit{Structural assumptions} via conditional independences: 
\begin{assumption}\label{ass:struct_ass_1}
 $Y \; \indep Z \mid A,U,X$.
\end{assumption} 
\begin{assumption}\label{ass:struct_ass_2}
\(W \indep (A,Z) \mid  U,X\).
\end{assumption}
These assumptions are very general: they do not enforce restrictions on the functional form of the confounding effect, or indeed on any other effects.
Note that we are not restricting the confounding structure, since we do not make any assumption on the additivity of confounding effect, or on the linearity of the relationship between variables. 

(B) \textit{Completeness assumptions} on the ability of proxy variables 
to characterize the latent confounder:
\begin{assumption} \label{ass:completeness_1} Let $l$ be any square integrable function. Then $\E [l(U)\,|\,a, x, z] \!=\! 0$ for all  $(a,x,z)\in \A \times \X \times \Z$ , if and only if \(l(U) \!=\! 0\) almost surely.
\end{assumption}
\begin{assumption}\label{ass:completeness_2} Let $g$ be any square integrable function.
Then $ \E [g(Z)\,|\,a,x,w] \!=\! 0, \forall(a,x,w)\!\in\! \A \times \X \times \W$ if and only if $g(Z) \!=\! 0$ almost surely.
\end{assumption}
These assumptions guarantee that the proxies are sufficient to describe $U$ for the purposes of ATE estimation. For better intuition we can look at the discrete case: for categorical $U,Z,W$ the above assumptions imply that proxies $W$ and $Z$ have at least as many categories as $U$.
Further, it can be shown that \Cref{ass:completeness_2} along with certain regularity conditions \citep[Appendix, Conditions (v)-(vii)]{miao2018identifying} guarantees that there exists at least one solution to the following 
integral equation:
\begin{equation}\label{eq:integral-eq}
 \E [Y \mid a,x,z] = \int_\mathcal{W} h(a,x,w)\rho(w|a,x,z)\,dw,
\end{equation}
which holds for all $(a,x,z) \in \A \times \X \times \Z$. 
We discuss the completeness conditions in greater detail in \Cref{sec:completeness}.

Given these assumptions, it was shown by \citet{miao2018confounding} that the function $h(a,x,w)$ in \eqref{eq:integral-eq} can be used to identify the causal effect $\mathbb{E}[Y\,|\,do(A=a)]$ as follows, 
\begin{equation}\label{Eq:average_causal_effect}
    \resizebox{0.9\hsize}{!}{$\mathbb{E}[Y\,|\,do(A=a)] = {\int_{\X,\W} h(a,x,w)\rho(x,w) \, dx dw}.$}
\end{equation}




While the causal effect can be identified, approaches for estimating this effect in practice are less well established,  and include
\citet{deaner2021proxy} (via a method of sieves) and \citet{tchetgen2020introduction} (assuming linearity).
The related IV setting has well established estimation methods, however the proximal setting relies on fundamentally different assumptions on the data generating process. 
None of the three key assumptions in the IV setting (namely the relevance condition, exclusion restriction, or unconfounded instrument) are required in proximal setting. 
In particular, we need a set of proxies which are complete for the latent confounder, i.e., dependent with the latent confounder, whereas a valid instrument is independent of the confounder. 
In this respect, the proximal setting is more general than the IV setting, including  the  recent ``IVY'' method of \citet{kuang2020ivy}. 

 Before describing our approach to the problem of estimating the causal effect in \eqref{Eq:average_causal_effect}, we give a brief background on reproducing kernel Hilbert spaces and the additional assumptions we need for estimation.

\subsection{Reproducing Kernel Hilbert Spaces (RKHS)}

For any space $\mathcal{F} \in \{\A,\X,\W,\Z\}$, let $k:\mathcal{F}\times \mathcal{F}\to \R$ be a positive semidefinite kernel. 
We denote by $\phi$ its associated canonical feature map $\phi(x)=k(x,\cdot)$ for any $x\in \mathcal{F}$, and $\cH_{\mathcal{F}}$ its corresponding RKHS of real-valued functions on $\mathcal{F}$. 
The space $\cH_{\mathcal{F}}$ is a Hilbert space with inner product $\ps{\cdot,\cdot}_{\cH_{\mathcal{F}}}$ and norm $\Vert \cdot \Vert_{\cH_{\mathcal{F}}}$. 
It satisfies two important properties: (i) $k(x,\cdot)\in\cH_{\mathcal{F}}$ for all $x\in\mathcal{F}$, (ii) the reproducing property: for all $f \in \cH_{\mathcal{F}}$ and $x\in \mathcal{F}$, $f(x)=\ps{f,k(x,\cdot)}_{\cH_{\mathcal{F}}}$.
We denote the tensor product and Hadamard product by $\otimes$ and $\odot$ respectively. 
For $\mathcal{F},\mathcal{G} \in \{\A,\X,\W,\Z\}$, we will use $\cH_{\mathcal{F}\mathcal{G}}$ to denote the product space $\mathcal{H}_{\mathcal{F}}\times \mathcal{H}_{\mathcal{G}}$.
It can be shown that $\mathcal{H}_{\mathcal{F}\mathcal{G}}$ is isometrically isomorphic to $\mathcal{H}_{\mathcal{F}} \otimes  \mathcal{H}_{\mathcal{G}}$.
For any distribution $\rho$ on $\mathcal{F}$, $\mu_{\rho} := \int k(x,\cdot)d\rho(x)$ is an element of $\mathcal{H}_{\mathcal{F}}$ and is referred to as the kernel mean embedding of $\rho$ \citep{Smola07Hilbert}. 
Similarly, for any conditional distribution $\rho_{X|z}$ for each $z\in\Z$, $\mu_{X|z} := \int k(x,\cdot)d\rho(x|z)$ is a conditional mean embedding of $\rho_{X|z}$ \citep{song2009hilbert,song2013kernel}; see \citet{Muandet17:KME} for a review.

\subsection{Estimation Assumptions}
To enable causal effect estimation in the proxy setting using kernels, we require the following additional assumptions.

\begin{assumption}[Regularity condition]
\label{ass:polish_spaces}
    $\A,\X,\Y,\W,\Z$ are measurable, 
    separable Polish spaces.
\end{assumption}
\Cref{ass:polish_spaces} allows us to define the conditional mean embedding operator and a Hilbert–Schmidt operator. 
\begin{assumption}\label{ass:y_bounded}
\resizebox{0.7\hsize}{!}{$\exists c_Y<\infty$, $|Y|<c_Y$ a.s. and $\E[Y]<c_Y$.}
\end{assumption}
\begin{assumption}[Kernels]
\label{ass:kernel_characteristic}
\begin{enumerate*}[label=(\roman*)]
  \item $\kw (w,\cdot)$ is a characteristic kernel.
  \item $\ka(a,\cdot)$, $\kx(x,\cdot)$, $\kw(w,\cdot)$ and $\kz(z,\cdot)$ are continuous, bounded by $\kappa>0$, and their 
  feature maps 
    are measurable.
\end{enumerate*}
\end{assumption}
The kernel mean embedding of any probability distribution is injective
if a characteristic kernel is used \cite{sriperumbudur2011universality};
this guarantees that a probability distribution can be uniquely represented in an RKHS.
\begin{assumption}\label{as:strictly_positive_measure}
The measure $\mathcal{P_{AWX}}$ is a finite Borel measure with $\mathrm{supp}[P_{\mathit{AWX}}] = \mathcal{A \cross W \cross X}$.
\end{assumption}
We will assume that the problem is well-posed.
\begin{assumption}\label{ass:h_in_rkhs}
Let $h$ be the function defined in \eqref{eq:integral-eq}. We assume that $h \in \cH_{\A\X\W}$. 
\end{assumption}
Finally, given \cref{ass:h_in_rkhs}, we require the following completeness condition.
\begin{assumption}[Completeness condition in RKHS]
\label{as:Completeness_rkhs}
For all $g \in \mathcal{H_{AWX}}$: $\mathbb{E}_{\mathit{AWX}}[g(A,W,X)|A,Z,X] = 0$  $\mathcal{P_{AZX}}$-almost surely if and only if $g(a, w, x) = 0$, $\mathcal{P_{AWX}}-$almost surely. 
\end{assumption}
This condition guarantees the uniqueness of the solution to the integral equation \eqref{eq:integral-eq} in RKHS (see Lemma \ref{lemma4mmr} in Appendix \ref{app:C-PMMR}). 

\section{Kernel Proximal Causal Learning}
\label{sec:methods}

To solve the proximal causal learning problem, we propose two kernel-based methods, \emph{Kernel Proxy Variable} (KPV) and \emph{Proxy Maximum Moment Restriction} (PMMR).
The KPV decomposes the problem of learning function $h$ in \eqref{eq:integral-eq} into two stages: we first learn an empirical representation of $\rho(w|a,x,z),$ and then learn $h$ as a mapping from representation of $\rho(w|a,x,z)$ to $y$, with kernel ridge regression as the main apparatus of learning. This procedure is similar to  Kernel IV regression (KIV)  proposed by \citet{singh2019kernel}. 
 PMMR, on the other hand, employs the Maximum Moment Restriction (MMR) framework \citep{Muandet20:KCM}, which takes advantage of a closed-form solution for a kernelized conditional moment restriction. 
The structural function can be estimated in a single stage with a modified ridge regression objective. 
We clarify the connection between both approaches at the end of this section.

\subsection{Kernel Proxy Variable (KPV)}
\label{sec:Kernel Proxy Variable}
To solve \eqref{eq:integral-eq}, the KPV approach finds $h\in\cH_{\A\X\W}$ that minimizes the following risk functional:
\begin{align}\label{eq:upper_loss_main}
    &\tilde{R}(h) = \mathbb{E}_{AXZY}\left[\left(Y - G_h(A,X,Z)\right)^2\right], \\
    &G_h(a,x,z) := \int_{\mathcal{W}} h(a,x,w)\rho(w\,|\,a,x,z)dw\nonumber
\end{align}
Let $\mu_{W|a,x,z}\in \cH_{\W}$ be the conditional mean embedding of $\rho(W\,|\,a,x,z)$. Then, for any $h \in
\cH_{\A\X\W}$, we have: 
\begin{equation}\label{eq:h_expansion}
G_h(a,x,z)=
 \left\langle h, \phia(a,x) \otimes\mu_{W|a,x,z}\right\rangle_{\mathcal{H}_{\A\X\W}} 
\end{equation}
where $\phi(a,x)=\phia(a)\otimes \phix(x)$. This result arises from the properties of the RKHS tensor space $\cH_{\A\X\W}$ and of the conditional mean embedding.
We denote by $\eta_{AXW}$ the particular function $h$ minimizing \eqref{eq:upper_loss_main}.

The procedure to solve \eqref{eq:upper_loss_main} consists of two ridge regression stages. 
In the first stage, we learn an empirical estimate of $\mu_{W|a,x,z}$ using samples from $P_{\mathit{AXZW}}$. 
Based on the first-stage estimate $\widehat{\mu}_{W|a,x,z}$, we then estimate $\eta_{AXW}$ using samples from $P_{\mathit{AXZY}}$.
The two-stage learning approach of KPV offers flexibility: we can estimate causal effects where samples from the full joint distribution of $\{(a,x,y,z,w)_i\}_{i=1}^n$ are not available, and instead one only has access to samples $\{(a,x,z,w)_i \}_{i=1}^{m_1}$ and $\{(\ta,\tx,\tz, \ty)_j\}_{j=1}^{m_2}$. Note that while there is some similarity with the two-stage regression
used in kernel instrumental variable regression \citep[Section 4]{singh2019kernel}, 
 there is an important difference between the two methods: see \Cref{sec:RS_approach} for details.  
The ridge regressions for these two stages are given in \eqref{eq:erm1} and \eqref{eq:erm2}. 
The reader may refer to \cref{sec:kpv_detailed} for a detailed derivation of the solutions.

\textbf{Stage 1.} From the first sample $\{(a,x,z,w)_i \}_{i=1}^{m_1}$, learn the conditional mean embedding of $\rho(W|a,x,z)$, i.e., $\widehat{\mu}_{W|a,x,z} :=  \widehat{C}_{W|A,X,Z} \left( \phia(a) \otimes \phix(x) \otimes \phiz(z) \right)$ where $\widehat{C}_{W|A,X,Z}$ denotes the conditional mean embedding operator. 
We obtain $\widehat{C}_{W|A,X,Z}$ as a solution to:
\begin{align}\label{eq:erm1}
    &\widehat{C}_{W|A,X,Z} = \argmin_{C \in \cH_{\Gamma}} \; \widehat{E}(C),\text{ with }\\
    &\resizebox{0.9\hsize}{!}{$\widehat{E}(C)=\frac{1}{m_1}\sum\limits_{i=1}^{m_1}\| \phiw(w_i) -C\phi(a_i,x_i,z_i)\|_{\Hw}^2 + \lambda_1\|C\|^2_{\cH_{\Gamma}},$}\nonumber
\end{align}
where $\cH_{\Gamma}$ is the vector-valued RKHS  of operators mapping $\cH_{\A\X\Z}$ to $\Hw$. 
It can be shown that 
$\widehat{C}_{W|A,X,Z}= \Phi({W}) (\mathcal{K}_{AXZ}+ m_1 \lambda_1)^{-1}\Phi^T(A, X,Z) $ where \( \mathcal{K}_{AXZ}=K_{AA} \odot K_{XX} \odot K_{ZZ} \) and $K_{AA}, K_{XX}$ and $K_{ZZ}$ are $m_1\times m_1$ kernel matrices and  $\Phi({W})$ is a vectors of $m_1$ columns, with $\phi(w_i)$ in its $i$th column \citep{song2009hilbert,grunewalder2012conditional, singh2019kernel}. 
Consequently,   $\widehat{\mu}_{W|a,x,z}= \Phi({W}) (\mathcal{K}_{AXZ}+ m_1 \lambda_1)^{-1}\mathcal{K}_{axz} $ with $\mathcal{K}_{axz}=K_{Aa} \odot K_{Xx} \odot K_{Zz} $, where $K_{Aa}$ is a $m_1 \times 1$ vector denoting $k(a_s,a)$ evaluated at all $a_s$ in sample 1. 

\textbf{Stage 2.} From the second sample $\{(\ta,\tx,\tz,\ty)_j\}_{j=1}^{m_2}$, learn $\hat\eta$ via empirical risk minimization (ERM):
\begin{align}\label{eq:erm2}
 &\widehat{\eta}_{AXW}=\argmin_{\eta\in\cH_{\A\X\W}}\; \widehat{L}(\eta) ,\text{ where }\\
    &\resizebox{0.9\hsize}{!}{$\widehat{L}(\eta)= \frac{1}{m_2}\sum\limits_{j=1}^{m_2}( \ty_j- \eta[ \phia(\ta_j,\tx_j)\otimes \widehat{\mu}_{W|\ta_j,\tx_j,\tz_j} ] )^2 \nonumber+\lambda_2 \Vert\eta\Vert^2_{\cH_{\A\X\W}}.$}\nonumber
\end{align}
where 
$
\eta[\phia(\ta)\otimes\phix(\tx)\otimes\widehat{\mu}_{W|\ta,\tx,\tz}] =
  \left\langle \eta, \phia(\ta)\otimes\phix(\tx)\otimes\widehat{\mu}_{W|\ta,\tx,\tz}\right\rangle_{\cH_{\A\X\W}}
$
since $\eta\in\cH_{\A\X\W}.$ 
The estimator $\widehat{\eta}_{\mathit{AXW}}$ given by \eqref{eq:erm2} 
has a closed-form solution \cite{caponnetto2007optimal, smale2007}. 

\begin{thm}\label{thm:closed_form2}
For any $\lambda_2>0$, the solution of \eqref{eq:erm2} exists, is unique, and is given by $\widehat{\eta}_{AXW}=(\boldsymbol{\widehat{T}}_2 +\lambda_2)^{-1} \widehat{g}_2$ where
\begin{align*}
    &\resizebox{0.9\hsize}{!}{$\boldsymbol{\widehat{T}}_2 = \frac{1}{m_2} \sum\limits_{j=1}^{m_2} \left[ \widehat{\mu}_{W|\ta_j,\tx_j,\tz_j}\otimes \phi(\ta_j,\tx_j)\right]\otimes \left[ \widehat{\mu}_{W|\ta_j,\tx_j,\tz_j}\otimes 
    \phi(\ta_j,\tx_j)\right]$}\\
    &\widehat{g}_2 =\frac{1}{m_2}\sum_{j=1}^{m_2} \left[ \widehat{\mu}_{W|\ta_j,\tx_j,\tz_j}\otimes \phi(\ta_j,\tx_j)\right]\ty_j.
\end{align*}
\end{thm}
\begin{remark}\label{rem1:kpv} In the first stage, we learn the functional dependency of an \textit{outcome-induced proxy} on the \textit{cause} and an \textit{exposure induced proxy}. Intuitively, one can interpret the set of $(a,x,z)$ (and not the individual elements of it) as the instrument set for $w,$ with $\mu_{W|a,x,z}$ as a pseudo-structural function capturing the dependency between an instrument set and the target variable. It can be shown that for any fixed $a,x \in \A \times \X$,  $\mu_{W|a,x,z}$ represents the dependency between $W$ and $Z$ due to the common confounder $U$ which is not explained away 
by $a$ and $x$. If $a=u$ for any $a,u \in \A \times \U$, $\rho(W|a,x,z)=\rho(W|a,x)$ and subsequently, $\mu_{W|a,x,z}=\mu_{W|a,x}$  \cite{miao2018confounding}. 
\end{remark}


Theorem \ref{thm:closed_form2} is the precise adaptation of \citet[eq. 3.3, 2021 version]{deaner2021proxy}
to the case of infinite feature spaces, including the use of ridge
regression in Stages 1 and 2, and the use of 
tensor product features. As we deal with infinite feature spaces,
however, we cannot write our solution in terms of explicit feature
maps, but we must express it in terms of feature inner products (kernels),
following the form required by the representer theorem (see \Cref{prop:empiricalFinalForm} below).

As an alternative solution to kernel proximal causal learning, 
one might consider using  the Stage 1 estimate of $\widehat{\mu}_{W,A|A,X,Z}\left(\phi(a)\otimes\phi(x)\otimes\phi(z)\right):=\widehat{\mathbb{E}}\left(\phi(W)\otimes\phi(A)\vert a,x,z\right)$ 
as an input in Stage 2,  
which would allow an unmodified use of the KIV algorithm \cite{singh2019kernel} in the proxy setting.
Unfortunately regression from $\phi(a)$ to $\phi(a)$ is in population limit the identity mapping $I_{\mathcal{H_{A}}}$,
which is not Hilbert-Schmidt for characteristic RKHS, violating the well-posedness assumption for consistency of Stage 1 regression  \citep{singh2019kernel}.
In addition, 
predicting $\phi(a)$ via ridge regression from $\phi(a)$ introduces bias
in the finite sample setting, 
which may impact performance in
the second stage (see  \Cref{sec:RS_approach} for an example).


The KPV algorithm benefits from theoretical guarantees under well-established smoothness assumptions. The main assumptions involve well-posedness (i.e., minimizers belong to the RKHS search space) and source conditions on the integral operators for stage $1$ and $2$, namely \Cref{ass:well_posed1,ass:source1,ass:well_posed2,ass:source2} in \Cref{sec:kpv_detailed}. Specifically, $c_1$ and $c_2$ characterize the smoothness of the integral operator of Stage 1 and 2, respectively, while $b_2$ characterizes the eigenvalue decay of the  Stage 2 operator.

\begin{thm}\label{thm:kpv_final_thm}
Suppose \Cref{ass:polish_spaces,ass:kernel_characteristic,ass:y_bounded,ass:well_posed1,ass:source1,ass:well_posed2,ass:source2} hold. 
Fix $\zeta>0$ and choose $\lambda_1=m_1^{\frac{1}{c_1+1}}$ and $m_1 =m_2^{\frac{\zeta(c_1+1)}{(c_1-1)}}$.

\begin{enumerate}
\setlength{\itemindent}{-0.1in}
    \item If $\zeta\le \frac{b_2(c_2+1)}{b_2 c_2+1}$, choose $\lambda_2={m_2}^{-\frac{\zeta}{c_2+1}}$. Then $\tilde{R}(\widehat{\eta}_{AXW})-\tilde{R}(\eta_{AXW})=O_p\left(m_2^{-\frac{\zeta c_2}{c_2+1}}\right)$.
    \item If $\zeta \ge \frac{b_2(c_2+1)}{b_2 c_2+1}$, choose $\lambda_2={m_2}^{-\frac{b_2}{b_2 c_2+1}}$. Then $\tilde{R}(\widehat{\eta}_{AXW})-\tilde{R}(\eta_{AXW})=O_p\left(m_2^{-\frac{b_2 c_2}{b_2 c_2+1}}\right)$.
\end{enumerate}
\end{thm}
Our proof is adapted from \citet{szabo2016learning, singh2019kernel} and is given in \Cref{sec:kpv_detailed}. 
This leads to the following non-asymptotic guarantees for the estimate of the causal effect \eqref{Eq:average_causal_effect} at test time.
\begin{proposition}\label{prop:rate_kpv_ce}
Consider a sample at test time $\{(x,w)_i\}_{i=1}^{n_t}$. Denote by $\widehat{\mu}_{XW}=\frac{1}{n_t}\sum_{i=1}^{n_t}[\phix(x_i)\otimes\phi(w_i)]$ the empirical mean embedding of $(X,W)$. For any
 $a\in \A$, the KPV estimator of the causal effect \eqref{Eq:average_causal_effect} is given by
$\widehat{\beta}(a)= \widehat{\eta}_{AXW}[\widehat{\mu}_{XW}\otimes \phi(a)]$.
Suppose the assumptions of \Cref{thm:kpv_final_thm} hold.  The estimation error at test time is bounded by 
\begin{enumerate}
\setlength{\itemindent}{-0.1in}
    \item If $\zeta\le \frac{b_2(c_2+1)}{b_2c_2+1}$, $|\widehat{\beta}(a)-\beta(a)|\le O_p(n_t^{-\frac{1}{2}}+ m_2^{-\frac{\zeta (c_2-1)}{c_2+1}})$.
    \item If $\zeta\ge \frac{b_2(c_2+1)}{b_2c_2+1}$,  $|\widehat{\beta}(a)-\beta(a)|\le O_p(n_t^{-\frac{1}{2}}+ m_2^{-\frac{b_2(c_2-1)}{b_2c_2+1}})$.
\end{enumerate}
\end{proposition}
A proof of \Cref{prop:rate_kpv_ce} is provided in \Cref{sec:kpv_detailed}.\footnote{The rates appearing in the ICML 2021 proceedings were faster, since we mistakenly used an $L_2$ norm rather than an RKHS norm in the proof. See  \Cref{sec:proof_rate_kpv_ce} for the correction. We thank Rahul Singh for pointing out the error.} 
Following \citet{singh2020kernel}, this combines \Cref{thm:kpv_final_thm} with a probabilistic bound for the difference between $\widehat{\mu}_{XW}$ and 
its population counterpart.

From \eqref{eq:h_expansion}, a computable kernel solution follows from the representer theorem \citep{Scholkopf01:Representer}, 
\begin{equation*}
\eta_{AXW} \left(\phia({\ta_q,\tx_q}) \otimes \widehat{\mu}_{W|\ta,\tx,\tz} \right)= \sum\limits_{i,s=1}^{m_1}\sum\limits_{j=1}^{m_2}\alpha_{ij} A_{ijsq}
\end{equation*}
with $A_{ijsq}=\{ K_{w_iw_s} \left[\mathcal{K}_{AXZ} + m_1 \lambda_1 \right]^{-1}\mathcal{K}_{\ta_q\tx_q\tz_q}\}\overline{\mathcal{K}}_{\ta_q\tx_q}$ where  $\mathcal{K}_{\ta_q\tx_q\tz_q}=K_{A\ta_q}\odot K_{X\tx_q}\odot K_{Z\tz_q}$ and $ \overline{\mathcal{K}}_{\ta_q\tx_q}=K_{\ta_j\ta_q}\odot K_{\tx_j\tx_q}$ (note that a correct solution requires $m_1 \times m_2$ coefficients, and cannot be expressed by $m_2$ coefficients alone; see  \Cref{sec:discussion_on_eta}). Hence, the problem of learning $\widehat{\eta}_{AXW}$ amounts to learning $\widehat{\alpha} \in \R^{m_1 \times m_2}$, for which classical ridge regression closed form solutions are available, i.e.,
\begin{align}\label{eq:ERMalpha}
   \widehat{\alpha}=\argmin_{\alpha \in \R^{m_1 \times m_2}}  &\frac{1}{m_2} \sum^{m_2}_{q=1}\left (\ty_q- 
    \sum^{m_1}_{i,s=1}\sum^{m_2}_{j=1} \alpha_{ij}A_{ijsq}  \right)^2\nonumber \\ 
   &+\lambda_2  \sum^{m_1}_{i,r=1}\sum^{m_2}_{j,t=1} \alpha_{ij} \, \alpha_{rt} \,B_{ijrt},
\end{align}
with $B_{ijrt}=K_{w_iw_r} \,K_{\ta_j \ta_t} K_{\tx_j \tx_t} $. Obtaining $\widehat{\alpha}$
involves inverting a matrix of dimension $m_1m_2 \times m_1m_2$, which has a complexity of  $\mathcal{O}((m_1m_2)^3)$. Applying the Woodbury matrix identity on a vectorized version of  \eqref{eq:ERMalpha}, we get a cheaper closed-form solution for $\Hat{\nu}=vec(\hat{\alpha})$ below.
\begin{proposition}\label{prop:empiricalFinalForm} Consider the optimization problem of \eqref{eq:ERMalpha}. Let $\nu=vec(\alpha)$, and $\hat{\nu}$ the solution to the vectorized ERM in \eqref{eq:ERMalpha}. We can show that for  $\forall p, q \in \{1,\dots,m_2\}$:
\begin{align*}
    &\Hat{\nu}=\left(\Gamma_{(\tA,\tX,\tZ)} \overline{\otimes} I_{m_2\times m_2}\right) \left( m_2\lambda_2 + \Sigma \right)^{-1}y
    \; \in \R^{m_1m_2}, \\
    &\Sigma= \left(\Gamma^T_{(\ta_q,\tx_q,\tz_q)} K_{WW}\Gamma_{(\ta_p,\tx_p,\tz_p)}\right)(K_{\ta_q\ta_p}K_{\tx_q\tx_p}), \nonumber\\
    &\Gamma_{(\tA,\tX,\tZ)}=(\mathcal{K}_{AXZ}+ m_1 \lambda_1)^{-1}(K_{A\tA}\odot K_{X\tX}\odot K_{Z\tZ}),   \nonumber
\end{align*}    
where $\overline{\otimes}$ represents tensor product of associated columns of matrices with the same number of columns. 
\end{proposition}
 The details of the derivation are deferred to \Cref{sec:kpv_detailed}.
 Following these modifications, we only need to invert an $m_2\times m_2$ matrix.
 


\subsection{Proxy Maximum Moment Restriction (PMMR)}


The PMMR relies on the following result, whose proof is given in \Cref{app:C-PMMR}.

\begin{lemma}\label{lem:proxy-mmr}
A measurable function $h$ on $\mathcal{A} \times \mathcal{W} \times \mathcal{X}$ is the solution to \eqref{eq:integral-eq} if and only if it satisfies the conditional moment restriction (CMR) :   $ \mathbb{E}[Y - h(A,W,X)\,|\,A, Z, X] = 0$, $\mathbb{P}(A,Z,X)$-almost surely.
\end{lemma}

By virtue of Lemma \ref{lem:proxy-mmr}, we can instead solve the integral equation \eqref{eq:integral-eq} using tools developed to solve the CMR \citep{Newey199316EE}.
By the law of iterated expectation, if $\mathbb{E}[Y - h(A,W,X)|A,Z,X] = 0$ holds almost surely, it implies that $\mathbb{E}[(Y - h(A,W,X))g(A,Z,X)] = 0$ for any measurable function $g$ on $\A\times \X\times \Z$. That is, the CMR gives rise to a continuum of conditions which $h$ must satisfy. 

A maximum moment restriction (MMR) framework \citep{Muandet20:KCM} requires that the moment restrictions hold \emph{uniformly} over all functions $g$ that belong to a certain class of RKHS. Based on this framework, \citet{zhang2020maximum} showed, in the context of IV regression, that $h$ can be estimated consistently. 
In this section, we generalize the method proposed in \citet{zhang2020maximum} to the proxy setting by restricting the space of $g$ to a unit ball of the RKHS $\mathcal{H}_{\A\Z\X}$ endowed with the kernel $k$ on $\A\times\Z\times\X$. 

Before proceeding, we note that \citet{miao2018confounding} and \citet{deaner2021proxy} also consider the CMR-based formulation in the proxy setting, but the techniques employed to solve it are different from ours.
The MMR-IV algorithm of \citet{zhang2020maximum} also resembles other recent generalizations of GMM-based methods, notably, \citet{Bennett19:DeepGMM}, \citet{Dikkala20:Minimax}, and \citet{Liao20:NeuralSEM}; see \citet[Sec. 5]{zhang2020maximum} for a detailed discussion. 

\vspace{-7pt}
\paragraph{Objective.}
To solve the CMR, the PMMR finds the function $h\in\mathcal{H}_{\A\X\W}$ that minimizes the MMR objective:
\begin{equation*}
     R_k(h) = \sup_{\substack{g \in \mathcal{H}_{\A\X\Z}\\ \|g\|\leq 1}} \left(\mathbb{E}[(Y - h(A,W,X))g(A,Z,X)]\right)^2 
   \end{equation*}
Similarly to \citet[Lemma 1]{zhang2020maximum}, $R_k$ can be computed in closed form, as stated in the following lemma; see \Cref{app:C-PMMR} for the proof.
\begin{lemma}\label{lem:pmmr_closed_form} 
Assume that $$\E[(Y- h(A,W,X))^2 k((A,Z,X),(A,Z,X))] < \infty$$ 
and denote by  $V'$  an independent copy of the random variable $V$. Then,
$R_k(h) = \mathbb{E}[(Y- h(A,W,X))(Y' - h(A',W',X')) k((A,Z,X), (A',Z',X'))]$.
\end{lemma}

Unlike \citet{zhang2020maximum}, in this work the domains of $h$ and $g$ are not completely disjoint. That is, $(A,X)$ appears in both $h(A,W,X)$ and $g(A,Z,X)$.
Next, we also require that $k$ is integrally strictly positive definite (ISPD).
\begin{assumption} \label{ass:ispd}
The kernel $k: (\mathcal{A} \times \mathcal{Z} \times \mathcal{X})^2 \to \mathbb{R}$ is continuous, bounded, and is integrally strictly positive definite (ISPD), i.e., for any function $f$ that satisfies $0 < \|f\|_2^2 < \infty$, we have $\iint f(v)k(v,v')f(v') \,dv dv' > 0$ where $v := (a,z,x)$ and $v':=(a',z',x')$.
\end{assumption}

\vspace{-7pt}
\paragraph{A single-step solution.}
Under \Cref{ass:ispd}, \citet[Theorem 1]{zhang2020maximum} guarantees that the MMR objective preserves the consistency of the  estimated $h$; $R_k(h) = 0$ if and only if $\mathbb{E}[Y - h(A,W,X)|A,Z,X] = 0$ almost surely. 
Hence, our Lemma \ref{lem:proxy-mmr} implies that any solution $h$ for which $R_k(h) = 0$ will also solve the integral equation \eqref{eq:integral-eq}.

Motivated by this result, we propose to learn $h$ by minimizing the empirical estimate of $R_k$ based on an i.i.d. sample $\{(a,z,x,w,y)_i\}_{i=1}^{n}$ from $P(A,Z,X,W,Y)$.
By Lemma \ref{lem:pmmr_closed_form}, this simply takes the form of a V-statistic \citep{Serfling80:Approximation},
\begin{equation}\label{eq:mmr-vstat}
    \widehat{R}_{V}(h) :=\frac{1}{n^2}\sum\limits_{i, j=1}^{n}(y_{i}-h_i)(y_{j}-h_j) k_{ij},
\end{equation}
where $k_{ij}:= k((a_i,z_{i},x_i), (a_j,z_{j},x_j))$ and $h_i := h(a_i,w_i,x_i)$.
When $i\neq j$, the samples $(a_i,w_i,x_i,y_i)$ and $(a_j,w_j,x_j,y_j)$ are indeed i.i.d. as required by Lemma \ref{lem:pmmr_closed_form}. However, when $i=j$, the samples are dependent. Thus, $\widehat{R}_V$ is a biased estimator of $R_k$.
The PMMR solution can then be obtained as a regularized solution to \eqref{eq:mmr-vstat},
\begin{equation}\label{eq:mmr_obj}
    \hat{h}_{\lambda} = \argmin_{h\in\mathcal{H}_{\A\X\W}}\; \widehat{R}_V(h) + \lambda\|h\|^2_{\mathcal{H}_{\A\X\W}}.
\end{equation}

Similarly to KPV, PMMR also comes with theoretical guarantees under a regularity assumption on $h$, characterized by a parameter $\gamma$, and a well-chosen regularization parameter.
\begin{thm}\label{thm:pmmr_final_thm}
Assume that \Cref{ass:y_bounded,ass:kernel_characteristic,ass:ispd} hold.
If $n^{\frac{1}{2}-\frac{1}{2}\max\left(\frac{2}{\gamma+2},\frac{1}{2}\right)}$ is bounded away from zero and  
$\lambda = n^{-\frac{1}{2}\max\left(\frac{2}{\gamma+2},\frac{1}{2}\right)}$,
then
\begin{equation}
    \| \hat{h}_{\lambda} - h \|_{\mathcal{H}_{\A\X\W}} = O_p\left(n^{-\frac{1}{2}\min\left(\frac{\gamma}{\gamma+2},\frac{1}{2}\right)}\right)
\end{equation}
\end{thm}
This result is new, \citet{zhang2020maximum} did not provide a convergence rate for their solutions. The proof shares a similar structure with that of \Cref{thm:kpv_final_thm}. There follows a bound for the estimate of the causal effect \eqref{Eq:average_causal_effect} at test time. 

\begin{proposition}\label{prop:rate_pmmr_ce}
Consider a sample $\{(x,w)_i\}_{i=1}^{n_t}$ at test time. 
For any
 $a\in \A$, the PMMR estimator of the causal effect \eqref{Eq:average_causal_effect} is given by
$\widehat{\beta}(a) = n^{-1}\sum_{i=1}^n \hat{h}(a, w_i, x_i)$.
Suppose the assumptions of \Cref{thm:pmmr_final_thm} hold. Then, the estimation error at test time is bounded by
\begin{equation*}
    |\widehat{\beta}(a)-\beta(a) | \le O_p\left(n_t^{-\frac{1}{2}}+n^{-\frac{1}{2}\min\left(\frac{\gamma}{\gamma+2},\frac{1}{2}\right)}\right),
\end{equation*}
\end{proposition}
where $\gamma$ can intuitively be  thought of as a regularity parameter characterising the smoothness of $h$; we provide a formal characterisation in Appendix \ref{app:C-PMMR} Def. 4. Similar to \Cref{prop:rate_kpv_ce}, the proof of \Cref{prop:rate_pmmr_ce} combines \Cref{thm:pmmr_final_thm} and a probabilistic bound on the difference between the empirical mean embedding of $(X,W)$ and its population version. The proofs are provided in Appendix \ref{app:C-PMMR}.

\vspace{-7pt}
\paragraph{Closed-form solution and comparison with MMR-IV.}
Using the representer theorem \citep{Scholkopf01:Representer}, we can express any solution of \eqref{eq:mmr_obj} as $\hat{h}(a,w,x) = \sum_{i=1}^n \alpha_i k((a_i, w_i, x_i), (a,w,x))$ for some $(\alpha_i)_{i=1}^{n}\in \R^n$.
Substituting it back into \eqref{eq:mmr_obj} and solving for $\bm{\alpha} := (\alpha_1,\ldots,\alpha_n)$ yields
    $\bm{\alpha} = (LWL + \lambda L)^{-1}L W \bm{y}$
where $\bm{y} := (y_1,\ldots,y_n)^\top$ and $L,W$ are kernel matrices defined by $L_{ij} = k((a_i, w_i, x_i), (a_j,w_j,x_j))$ and $W_{ij} = k\left((a_i,z_{i},x_i), (a_j,z_{j},x_j)\right)$.
Thus, PMMR has time complexity of $\mathcal{O}(n^3)$, 
which can be reduced via the usual Cholesky or Nystr{\"o}m techniques. Moreover, whilst we note that PMMR is similar to its predecessor, MMR-IV, we discuss their differences in Appendix \ref{sec:pmmr_approach}, Remark 5; specifically, we discuss an interpretation of the bridge function $h$, the difference between noise assumptions, and generalisation to a wider class of problems.


\subsection{Connection Between the Two Approaches}
\label{sec:connection}

From a non-parametric instrumental variable (NPIV) perspective,
KPV and PMMR approaches are indeed similar to KIV \citep{singh2019kernel} and MMR-IV \citep{zhang2020maximum}, respectively. 
We first clarify the connection between these two methods in the simpler setting of IV. In this setting, we aim to solve the Fredholm integral equation of the first kind:
\begin{equation}\label{eq:fredholm}
    \mathbb{E}[Y\,|\,z] = \int_{\mathcal{X}} f(x) \,\rho(x|z)dx, \quad \forall z\in\mathcal{Z},
\end{equation}
where, with an abuse of notation, $X$, $Y$, and $Z$ denote endogenous, outcome, and instrumental variables, respectively.
A two-stage approach for IV proceeds as follows. In Stage 1, an estimate of the integral on the rhs of \eqref{eq:fredholm} is constructed. In Stage 2, the function $f$ is learned to minimize the discrepancy between the LHS and the estimated RHS of \eqref{eq:fredholm}. 
Formally, this can be formulated via the unregularized risk
\begin{align}
    L(f) :=& \mathbb{E}_{Z}[(\mathbb{E}[Y|Z] - \mathbb{E}[f(X)\,|\,Z])^2] \label{eq:natural-loss} \\
         &\leq \mathbb{E}_{\mathit{YZ}}[(Y - \mathbb{E}[f(X)\,|\,Z])^2] =: \tilde{L}(f) \label{eq:surrogate-loss}
\end{align}
The risk \eqref{eq:surrogate-loss} is considered in DeepIV \cite{Hartford17:DIV}, KernelIV \cite{singh2019kernel}, and DualIV \cite{Muandet19:DualIV}. 
Both DeepIV and KernelIV directly solve the surrogate loss \eqref{eq:surrogate-loss}, whereas DualIV solves the dual form of \eqref{eq:surrogate-loss}.
Instead of \eqref{eq:fredholm}, an alternative starting point to NPIV is the conditional moment restriction (CMR) $\mathbb{E}[\varepsilon\,|\,Z] = \mathbb{E}[Y - f(X)\,|\,Z] = 0$.  
\citet{Muandet20:KCM} showed that a RKHS for $h$ is sufficient in the sense that the inner maximum moment restriction (MMR) preserves all information about the original CMR.
Although starting from the CMR perspective, \citet{zhang2020maximum} used the MMR in the IV  setting to minimise loss as \eqref{eq:natural-loss}, as shown in \citet[Appendix F]{Liao20:NeuralSEM}.

\vspace{-7pt}
\paragraph{The connection.}
Both DeepIV \citep{Hartford17:DIV} and KernelIV \citep{singh2019kernel} (resp. KPV) solve an objective function that is an upper bound of 
the objective function of MMR \cite{zhang2020maximum} (resp. PMMR) and other GMM-based methods such as AGMM \citep{Dikkala20:Minimax} and DeepGMM \citep{Bennett19:DeepGMM}. This observation reveals the close connection between modern nonlinear methods for NPIV, as well as between our KPV and PMMR algorithms, which minimize $R$ and $\tilde{R},$ respectively, where
\begin{align}\label{eq:jensen}
&\resizebox{.85\hsize}{!}{$R(h) =\mathbb{E}_{AXZ}[(\mathbb{E}[Y|A,X,Z] - \mathbb{E}[h(A,X,W)\,|\,A,X,Z])^2]$} 
\\
&\resizebox{0.85\hsize}{!}{$\leq \mathbb{E}_{AXZY}[(Y - \mathbb{E}[h(A,X,W)\,|\,A,X,Z])^2] = \tilde{R}(h)$}\label{eq:surrogate_loss2} 
\end{align}
We formalize this connection in the following result.

\begin{proposition}\label{th:connection}
    Assume there exists $h \in L^2_{P_{\mathit{AXW}}}$ such that $\E[Y|A,X,Z]=\E[h(A,X,W)|A,X,Z]$. Then,
    \begin{enumerate*}[label=(\roman*)]
    \item $h$ is a minimizer of $R$ and $\tilde{R}$.
    \item Assume $k$ satisfies \Cref{ass:ispd}. Then, $R_k(h)=0$ 
    iff $R(h)=0$.
    \item  Assume $k$ satisfies \Cref{ass:kernel_characteristic}. Suppose  that
        $\E[f(W)|A,X,Z=\cdot]\in \cH_{\A\X\Z}$ for any $f\in\Hw$ and that \Cref{ass:h_in_rkhs} holds.
        Then, $h$ is given by the KPV solution, and is a minimizer of $R$ and $\tilde{R}$.
    \end{enumerate*}
\end{proposition}
See \Cref{sec:connection_proof} for the proof of \Cref{th:connection}. The assumptions needed for the third result ensures that the conditional mean embedding operator is well-defined and characterizes the full distribution $P_{A,X,W|A,X,Z}$, and that the problem is well-posed.

\begin{remark}
The KPV and PMMR approaches minimise risk 
in different ways, and hence they offer two different representations of $h$. 
KPV minimises $\tilde{R}$ by first estimating the empirical conditional mean embedding $\widehat{\mu}_{W|a_i,z_i,x_i}$ from a sub-sample $\{(a,w,x)_i\}_{i=1}^{m_1}$, and then estimating $h$ by minimising $\frac{1}{m_2}\sum_{j=1}^{m_2}(y_j- \E_{W|a_j,x_j,z_j}(h))^2$ over a sub-sample $\{(a,z,x)_j\}_{j=1}^{m_2}$ using $\widehat{\mu}_{W|a_i,z_i,x_i}$ obtained from the first stage.
Hence, 
$h^{\text{KPV}}(a,w,x) = \sum^{m_1}_{i=1} \sum^{m_2}_{j=1} \alpha_{ij} k(a_j,a)k(x_j,x)k(w_i,w)$ for some $(\alpha_{i,j})_{i,j=1}^{m_1,m_2} \in \R^{m_1\times m_2}$. 
In contrast, PMMR directly minimises $R_k$,
resulting in the estimator $\hat{h}^{\text{PMMR}}(a,w,x) = \sum_{i=1}^n \alpha_i k((a_i, w_i, x_i), (a,w,x))$ for some $(\alpha_i)_{i=1}^{n}\in \R^n$, and over the joint distribution of  $\{(a,x,w,z)_i\}_{i=1}^{n}$.


\end{remark}

\section{Experiments}
\setlength\abovecaptionskip{-8pt}
\setlength\belowcaptionskip{-10pt}
 \begin{figure*}[t!]
\begin{minipage}{.30\textwidth}
  \centering
  \includegraphics[width=\linewidth]{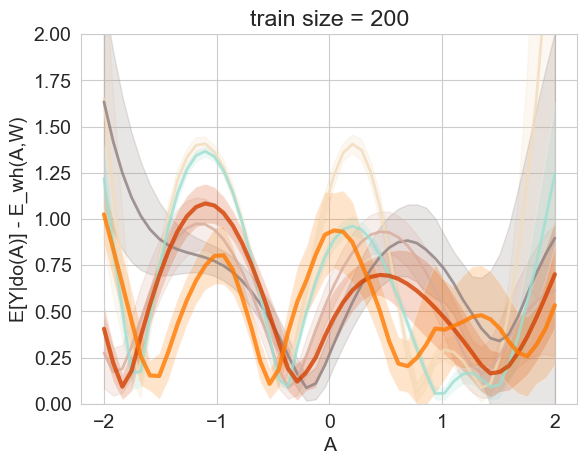}
  \label{fig:Identity}
\end{minipage}%
\begin{minipage}{.30\textwidth}
  \centering
  \includegraphics[width=\linewidth]{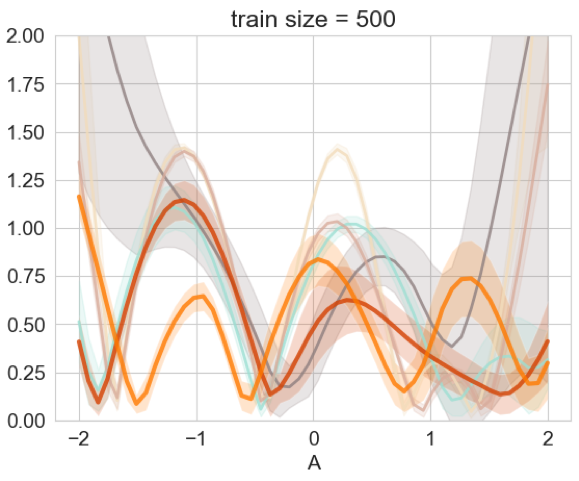}
  \label{fig:test2}
\end{minipage}
\begin{minipage}{.30\textwidth}
  \centering
  \includegraphics[width=\linewidth]{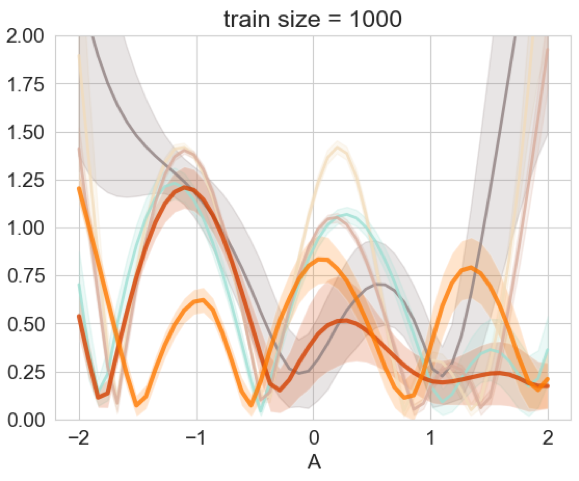}
  \label{fig:test2}
\end{minipage}
\begin{minipage}{.09\textwidth}
  \centering
  \includegraphics[scale=.4]{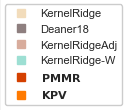}
  \label{fig:test2}
\end{minipage}
\caption{Mean absolute error of ATE estimation for each A of (\ref{eq:synthetic_model_AR}). The lower the mean absolute error the stronger is the performance of the model. Tchetgen-Tchetgen20 has error out of the range so we omit it for clearer plots.}\label{fig:synthetic_res}
\end{figure*}


We evaluate KPV and PMMR against the following baselines: (1) \textit{KernelRidge}: kernel ridge regression $Y\sim A$. (2) \textit{KernelRidge-W}: kernel ridge regression $Y\sim (A,W)$, adjusted over $W$, i.e., $\int \mathbb{E}[Y|A,W]d\rho(W)$. (3)\textit{KernelRidge-{W,Z}}: kernel ridge regression $Y\sim (A,W,Z)$, adjusted over $W$ and $Z$, i.e., $\int \mathbb{E}[Y|A,W,Z]d\rho(W,Z)$. Methods (2) and (3) are implemented in accordance with \cite{singh2020kernel}. (4) \textit{Deaner18}: a two-stage method with a finite feature dictionary \cite{deaner2021proxy}, and (5) \textit{Tchetgen-Tchetgen20}: a linear  two-stage method consistent consistent with \citet{miao2018confounding}.  
The mean absolute error with respect to the true causal effect $\mathbb{E}[Y|do(A)]$, \textit{c-MAE}, is our performance metric. Without loss of generality, we assume that $X$ is a null set. Both KPV and PMMR are capable of handling non-null sets of covariates, $X$. Our codes are publicly available at: \renewcommand\UrlFont{\color{blue}}\url{  https://github.com/yuchen-zhu/kernel_proxies}.
The satisfaction of Assumptions
\ref{ass:struct_ass_1}- \ref{ass:completeness_2} will guarantee identification of $h$ on the support of data, but we still need $(A,W)$ to satisfy an overlap/positivity assumption (similar to classic positivity in causal inference,  e.g. \citet{Westreich_Positivity}) to guarantee empirical identification of $\mathbb{E}[Y|do(a)]$.  As we infer the causal effect from $h$ by $\int \hat{h}(a, w) d\rho(w)$,  we need $\rho(w|a)>0$ whenever $\rho(w) >0$ for $h$ to be well-identified for the marginal support of $W$. 
\subsection{Hyperparameter Tuning}
For both KPV and PMMR, we employ an exponentiated quadratic kernel for continuous variables, as it is continuous, bounded, and characteristic,  thus  meeting all required assumptions. KPV uses a leave-one-out approach with grid search for the kernel bandwidth parameters in both stages and the regularisation parameters (See \cref{sub:D21}). 
PMMR uses a leave-M-out with gradient descent approach to tune for the $L-$bandwidth parameters and the regularisation parameter $\lambda$, where the bandwidth parameters are initialised using the median distance heuristic and the regularisation parameters are initialised randomly. The approach is based on \citet{zhang2020maximum}. We defer the hyperparameter selection details to Appendix \ref{app:D-experiments}.

\subsection{Synthetic Experiment}\label{sec:results_synthetic}
\setlength\abovecaptionskip{-8pt}
\setlength\belowcaptionskip{-10pt}
\begin{figure*}[t!]
\begin{minipage}{.35\textwidth}
  \centering
  \includegraphics[width=\linewidth]{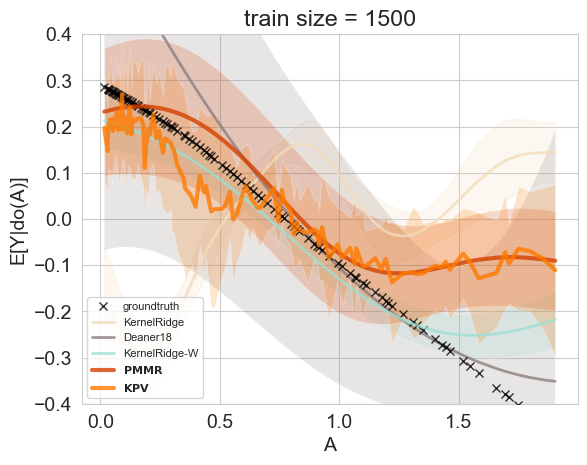}
  \label{fig:edu_im}
\end{minipage}%
\begin{minipage}{.32\textwidth}
  \centering
  \includegraphics[width=\linewidth]{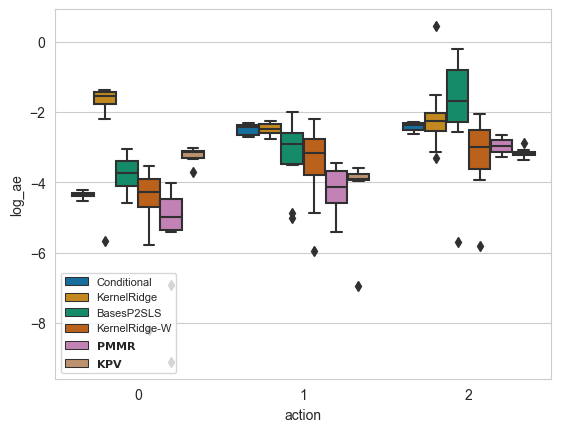}
  \label{fig:edu_im}
\end{minipage}%
\begin{minipage}{.32\textwidth}
  \centering
  \includegraphics[width=\linewidth]{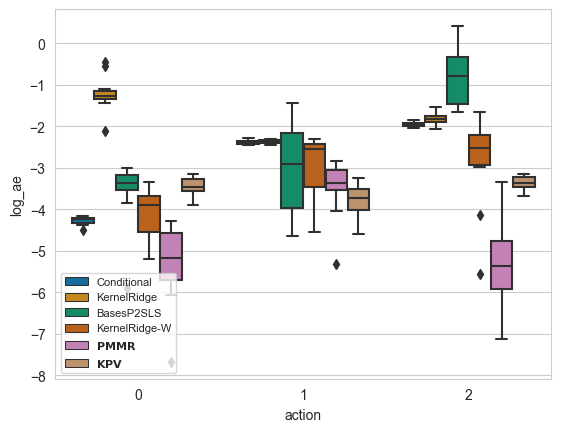}
  \label{fig:edu_ir}
\end{minipage}

\caption{(Right: Abortion and Criminality), ATE comparison of adjustment on W with ground truth and direct regression (Middle: Maths, Left: Reading) Grade retention and cognitive outcome; log MAE $\log|\beta(a) - \hat{\beta}(a)|$ over action values $a=0, 1, 2$; lower is better. 
}\label{fig:abort_and_edu_res}
\end{figure*}

First, we demonstrate the performance of our methods on a synthetic simulation with non-linear treatment and outcome response functions.  In our generative process, $U, W \text{ and  }Z \in \R^2$ and $A$ and $Y$ are scalar. We have defined the latent confounder $U$ such that $U_1$ is dependent on $U_2$. Appendix \ref{app:D-experiments} \cref{fig:synthetic_generative} demonstrates the relationship between $U_1$ and $U_2$. Given $U_2$, we know the range of $U_1$, but the reverse does not hold: knowning $U_1 \in [0,1]$, then $U_2$ is with equal probability in one of the intervals $[-1,0]$ or $[1,2]$. In design of the experiment, we also have chosen $W$ such that its first dimension is highly correlated with $U_1$ (less informative dimension of $U$) with small uniform noise, and its second dimension is a view of $U_2$ with high noise. With this design, it is guaranteed that $\{W\}$ is not a sufficient proxy set for $U$. See  \cref{eq:synthetic_model_AR} for details.
\begin{align}\label{eq:synthetic_model_AR}
U & :=[U_1 ,U_2], \quad U_2\sim  \mathrm{Uniform}[-1,2]\nonumber\\
U_1 &\sim \mathrm{Uniform}[0,1] - \mathds{1}[0\leq U_2\leq 1] \nonumber \\ 
W &:=[W_1, W_2]=[U_1+\mathrm{Uniform}[-1,1], U_2+ \mathcal{N}(0,\sigma^2)]\nonumber\\
Z &:=[Z_1, Z_2]=[U_1 + \mathcal{N}(0,\sigma^2), U_2 + \mathrm{Uniform}[-1,1] ]\nonumber\\
A &:= U_2 + \mathcal{N}(0,\beta^2)\nonumber\\
Y &:= U_2\cos(2(A+0.3U_1+0.2))
\end{align}
where $\sigma^2=3$ and $\beta^2=0.05$.

We use training sets of size 500 and 1000, and average results over 20 seeds affecting the data generation. The generative distribution is presented in  \Cref{app:D-experiments}, \cref{fig:synthetic_generative}.

\setlength\abovecaptionskip{2pt}
\setlength\belowcaptionskip{0pt}
\setlength{\textfloatsep}{0pt plus 1.0pt minus 2.0pt}
\begin{table}[h!]
    \centering
    \resizebox{0.45\textwidth}{!}{\begin{tabular}{|c|c|c|c|}
\hline
  Method & c-MAE(n=200) & c-MAE(n=500) & c-MAE(n=1000)\\
\hline
\textbf{KPV} & \textbf{0.499 $\pm$ 0.310} & \textbf{0.490 $\pm$ 0.285} & \textbf{0.491 $\pm$ 0.290} \\
\hline
\textbf{PMMR} & \textbf{0.533 $\pm$ 0.314} & \textbf{0.494 $\pm$ 0.330} & \textbf{0.472 $\pm$ 0.358} \\
\hline
KernelRidgeAdj & 0.569 $\pm$ 0.317 & 0.577 $\pm$ 0.352 & 0.607 $\pm$ 0.379 \\
\hline
KernelRidge-W & 0.635 $\pm$ 0.428 & 0.695 $\pm$ 0.460 & 0.716 $\pm$ 0.476 \\
\hline
KernelRidge & 0.840 $\pm$ 0.782 & 0.860 $\pm$ 0.709 & 0.852 $\pm$ 0.654 \\
\hline
Deaner18 & 0.681 $\pm$ 0.477 & 1.030 $\pm$ 1.020 & 1.050 $\pm$ 0.867 \\
\hline
Tchetgen-Tchetgen20 & 1.210 $\pm$ 1.070 & 17.60 $\pm$ 85.50 & 1.100 $\pm$ 1.460 \\
\hline
    \end{tabular}}
            \caption{Results comparing our methods to baselines on the simulation studies described in \cref{eq:synthetic_model_AR}}\label{tab:synthetic_res}
\end{table}
Table \ref{tab:synthetic_res} summarizes the results of our experiment with the synthetic data. 
Both KPV and PMMR, as methodologies designed to estimate unbiased causal effect in proximal setting, outperform other methods by a large margin, and have a narrow variance around the results. As expected, the backdoor adjustment for $(W,Z)$, the current common practice to deal with latent confounders (without considering the nuances of the proximal setting), does not suffice to unconfound the causal effect. Related methods, \textit{KernelRidge-(W,Z)} and  \textit{KernelRidge-W}, underperform our methods by large margins. As \cref{fig:synthetic_res} shows, they particulary fail to identify the functional form of the causal effect. \textit{Tchetgen-Tchetgen20} imposes a strong linearity assumption, which is not suitable in this nonlinear case, hence its bad performance. The underperformance of \textit{Deaner18}  is largely related to it only using a finite dictionary of features, whereas the kernel methods use an infinite dictionary.

\subsection{Case studies}\label{sec:results_real}

In the next two experiments,
our aim is to study the performance of our approaches in dealing with real world data.
To have a real causal effect for comparison, we fit a generative model 
to the data, and evaluate against simulations from the model. See \ref{app:D-experiments} for further discussion and for the full procedure.
Consequently,  we refrain from making any policy recommendation on the basis of our results. 
In both experiments, we sample a training set of size 1500, and average results over 10 seeds affecting the data generation.
\subsubsection{Legalized abortion and crime}
We study the data from \citet{donohuelevitt} on the impact of legalized abortion on crime. We follow the data preprocessing steps from \citet{Woody2020EstimatingHE}, removing the state and time variables. We choose the effective abortion rate as treatment ($A$), murder rate as outcome ($Y$), ``generosity to aid families with dependent children'' 
as treatment-inducing proxy ($Z$), and beer consumption per capita, log-prisoner population per capita and concealed weapons law as outcome-inducing proxies ($W$). We collect the rest of the variables as the unobserved confounding variables ($U$).

\textbf{Results.} Table \ref{tab:case_studies_res} includes all results. 
Both KPV and PMMR beat KernelRidge and BasesP2SLS by a large margin, highlighting the advantage of our methods in terms of deconfounding and function space flexibility. KernelRidge-W is the best method overall, beating the second best by a wide margin. We find this result curious, as Figure \ref{fig:abort_and_edu_res} shows that adjustment over $W$ is sufficient for identifying the causal effect in this case, however it is not obvious how to conclude this from the simulation. We leave as future work the investigation of conditions under which proxies provide a sufficient adjustment on their own.

\begin{table}[t!]
\centering
\resizebox{0.45\textwidth}{!}{\begin{tabular}{|c|c|c|c|}
\hline
 \textbf{Metric} &
  \multicolumn{3}{|c|}{\textbf{c-MAE}}  \\ \hline
\textbf{Method/Dataset} &
   Abort. \& Crim. &
  Grd Ret., Maths  &
  Grd. Ret., Reading\\ \hline
KPV &
  0.129 $\pm$ 0.105 &
  {0.036* $\pm$ 0.046}  &
  0.030* $\pm$ 0.051 \\ \hline
 PMMR &
  0.137 $\pm$ 0.101 & 
  0.032 $\pm$ 0.022 & 
  {0.023 $\pm$ 0.022}\\ \hline
Conditional &
  - &
  0.062 $\pm$ 0.036 & 
  0.083 $\pm$ 0.053  \\ \hline
KernelRidge &
  0.330 $\pm$ 0.186 &
  0.200 $\pm$ 0.631 & 
  0.190 $\pm$ 0.308 \\  \hline
KernelRidge-W &
  {0.056 $\pm$ 0.053} &
  0.031 $\pm$ 0.026 & 
  0.024 $\pm$ 0.021  \\ \hline
Deaner18 &
  0.369 $\pm$ 0.284 &
  0.137 $\pm$ 0.223 & 
  0.240 $\pm$ 0.383 \\ \hline
\end{tabular}}
    \caption{Results comparing our methods to baselines on the real-world examples described in \ref{sec:results_real}. \newline{$^*$ We identified a mistake in labeling proxy variables. The mislabeling only affected the causal effect estimated by KPV. We have corrected the mistake and reported the results in this version.   }}\label{tab:case_studies_res}
\end{table}
\subsubsection{Grade retention and cognitive outcome}
We use our methods to study the effect of grade retention on long-term cognitive outcome using data the ECLS-K panel study \citep{deaner2021proxy}. We take cognitive test scores in Maths and Reading at age 11 as outcome variables ($Y$), modelling each outcome separately, and grade retention as the treatment variable ($A$). Similar to \citet{deaner2021proxy}, we take the average of 1st/2nd and 3rd/4th year elementary scores as the treatment-inducing proxy ($Z$), and the cognitive test scores from Kindergarten as the outcome-inducing proxy ($W$). See Appendix \ref{app:D-experiments} for discussion on data. 

 
\textbf{Results.} Results are in Table \ref{tab:case_studies_res}. For both Math grade retention and Reading grade retention, our proposed methods outperform alternatives: KPV does better on the Math outcome prediction, while PMMR exceeds others in estimation for the Reading outcome. KernelRidge-W is still a strong  contender, but all other baselines result in large errors.

\section{Conclusion}

In this paper, we have provided two kernel-based methods to estimate the causal effect in a setting where proxies for the latent confounder are observed. Previous studies mostly focused on characterising identifiability conditions for the proximal causal setting, but lack of methods for estimation was a  barrier to wider implementation. Our work is primarily focused on providing two complementary approaches for  causal effect estimation in this setting. This will hopefully motivate further studies in the area. 

Despite promising empirical results, the hyperparameter selection procedure for both methods can be improved. For KPV, the hyperparameter tuning procedure relies on the assumption that optimal hyperparameters in the first and second stage can be obtained independently, while they are in fact interdependent. For PMMR, there is no systematic way of tuning the hyperparameter of the kernel $k$ that defines the PMMR objective, apart from the median heuristic. Developing a complete hyperparameter tuning procedure for both approaches is an important future research direction. Beyond this, both methods can be employed to estimate causal effect in wider set of problems, where the Average Treatment on the Treated, or Conditional Average Treatment Effect are the quantity of interests.

\section*{Acknowledgments}

RS was partially funded by a ONR grant, award number N62909-19-1-2096; YZ was funded by 
EPSRC with grant number EP/S021566/1.

\newpage
\bibliography{biblio}
\bibliographystyle{icml2021}

\newpage
\appendix
\onecolumn
\section{Completeness conditions}\label{sec:completeness}
\subsection{Completeness condition for continuous and categorical confounder}

The following two completeness conditions are necessary for the existence of solution for equation \eqref{eq:integral-eq} and the consistency of causal effect inference should a solution exist. They are studied as equations (13) and (16) in \citet{tchetgen2020introduction}.

\begin{enumerate}
    \item For all $g \in \mathcal{L}^2_{P_U}$ and for any $a, x$, $\mathbb{E}[g(U)|a,x,z] = 0$ $P_Z-a.s.$ if and only if $g(U) = 0$ $P_U-a.s.$ This condition guarantees the viability of using the solution to \eqref{eq:integral-eq} to consistently estimate the causal effect. Note that since $U$ is unobserved, this condition cannot be directly tested from observational data.
    \item For all $g \in \mathcal{L}^2_{P_Z}$ and for any $a, x$, $\mathbb{E}[g(Z)|a,x,w] = 0$ $P_W-a.s.$ if and only if $g(Z) = 0$ $P_Z-a.s.$ This is a necessary condition for the existence of a solution to \eqref{eq:integral-eq}. 
    With access to joint samples of $(a,x,w,z)$, in practice one can validate whether this condition holds and assess the quality of proxies $W,Z$ with respect to completeness condition. This assessment is beyond the scope of our study. 
\end{enumerate}
For a discrete confounder with categorical proxy variables, the combination of conditions $1$ and $2$ is equivalent to:
\begin{enumerate}
\setcounter{enumi}{2}
\item Both $W$ and $Z$ have at least as many categories as $U$.
\item For all $(a,x)$, the matrix $P$ where $P_{ij} = p(z_i|a,x,w_j)$ is invertible, with $z_i$ and $w_j$ denoting the $i$th and $j$th categories of $Z$ and $W$, respectively. 
Moreover, in the discrete case, this condition is necessary and sufficient for the solvability of Eq.\ref{eq:integral-eq} as studied extensively in \citet{miao2018identifying}, \citet{tchetgen2020introduction}. 
\end{enumerate}

\subsection{Falsifying examples of the completeness condition}\label{sec:comp_examp}
In this section we aim to provide intuition about the completeness conditions by giving examples of distributions which falsify them. For simplicity, we work with the completeness of $Z$ on $U$, which is the statement:

\textit{$Z$ is complete for $U$ if and only if for all $g$ which is square-integrable, $\mathbb{E}[g(u)|z] = 0$ $\mathcal{P}_{Z}-a.s.$ if and only if $g(u) = 0$ $\mathcal{P}_{U}-a.s.$}

 We proceed to provide examples in which the above statement fails to hold true. 

 \begin{itemize}
     \item Trivial example. If $Z \indep U$, then choose any non-zero square integrable $\Tilde{g} \in \mathcal{L}^2(\mathcal{U})$ and define $g = \Tilde{g} - \mathbb{E}[\Tilde{g}(U)]$. Clearly $g \not\equiv 0$, but $\mathbb{E}[g(U)|Z] = \mathbb{E}[g(U)] = \mathbb{E}[\Tilde{g}(U) - \mathbb{E}[\Tilde{g}(U)]] = 0$ 
  \item Merely requiring that $Z$ and $U$ are dependent is not enough. Let $U = (X_1, X_2)$ and let $Z = (X_1, X_1)$ where $X_1 \indep X_2$ and $X_1, X_2 \sim \mathcal{N}(0,1)$. Thus $U$ and $Z$ are dependent. But let $g(U) = X_2$, then clearly $\mathbb{E}[g(U)|Z] = \mathbb{E}[X_2|Z] = \mathbb{E}[X_2] = 0$ for all $Z$ almost surely. Thus $Z$ is not complete for $U$.
    \item The reader might find the above two examples both trivial since they both require some component of $U$ to be independent of all components of $Z$. In the most general setting, the completeness condition is falsified if there is a $g \not\equiv 0 \in \mathcal{L}^2_{\mathcal{P}_Z}$ which is \textit{orthogonal} to $\rho(u|z)$ for all values of $z$. This is equivalent to saying that:
    \begin{equation}
        \int_{\mathcal{U}}g(u)\rho(u|z)du = 0
    \end{equation}
    or, 
    \begin{equation}
        \int_{\mathcal{U}g^+}g^+(u) \rho(u|z) du = \int_{\mathcal{U}g^-}g^-(u) \rho(u|z) du
    \end{equation}
    $\mathcal{P}_Z-a.s.$, where $g^+$ and $g^-$ denotes the function or space restricted where $g$ is positive or negative, respectively. To see an example where this scenario can arise, and where all components of $Z$ are correlated with all components of $U$, consider the following. 
     Let $U\sim \mathcal{N}(0,1)$. $Z = f(U) + \mathcal{N}(0,1) = |U| + \mathcal{N}(0,1)$, where the added gaussian noise is independent of $U$. 
     Let $g$ be a square integrable odd function, that is to say, $g(-x)=-g(x)$. 
    
     Then, we may examine the expectation of $g$ given $z$ as follows:
     \begin{eqnarray}
     \mathbb{E}[g(U)|z] &=& \int_{-\infty}^{\infty} g(u) \rho(u|z) du\\
     &=& \int_{-\infty}^{0} g(u) \rho(u|z)du + \int_{0}^{\infty} g(u) \rho(u|z) du\\
     &=& \int_{\infty}^{0} -g(-v) \rho(-v|z) dv + \int_0^{\infty} g(u)\rho(u|z)du \label{eq: 12}\\
     &=& \int_{0}^{\infty}
     g(-v) \rho(-v|z) dv + \int_0^{\infty} g(u)\rho(u|z)du \label{eq: 13}\\
     &=& \int_{0}^{\infty}
    -g(v) \rho(-v|z) dv + \int_0^{\infty} g(u)\rho(u|z)du \label{eq: 14}\\
    &=& \int_{0}^{\infty}
    -g(u) \rho(-u|z) du + \int_0^{\infty} g(u)\rho(u|z)du \label{eq: 15}
    \end{eqnarray}
    where (\ref{eq: 12}) is by taking substitution $v=-u$, (\ref{eq: 13}) is swapping limit (\ref{eq: 14}) is by oddness of $g$ and (\ref{eq: 15}) is by renaming $v$ as $u$. 
    
    Now, $\rho(u|z)$ is symmetric in $U$, this can be seen by considering $\rho(u|z) = \frac{\rho(z|u) \rho(u)}{\rho(z)} \propto \rho(z|u)\rho(u)$. 
    
    $\rho(z|u)$ is symmetric in $u$ because $f(u) = |u|$ is symmetric; $\rho(u)$ is symmetric because it is a Gaussian; product of symmetric functions is symmetric.
    
    Therefore, 
    \begin{eqnarray}
    (\ref{eq: 15}) &=& \int_{0}^{\infty}
    -g(u) p(u|z) du + \int_0^{\infty} g(u)p(u|z)du = 0
    \end{eqnarray}
    
    Thus no component of $Z$ is independent of $U$ but $Z$ is not complete for $U$. 
    
    Notice that in this case, we were able to construct such a $g$ because $f$ and $\rho$ have the same line of symmetry. Although this is an interesting example of falsification of the completeness condition, it is perhaps an unstable - i.e. we might be able to restore completeness if we slightly perturb the line of symmetry of $f$ and $\rho$.
    \end{itemize}

\begin{remark}
 We note that although the completeness condition can be broken non-trivially by having a non-empty orthogonal set of $p(u|z)$ for almost all $z$, these cases might be unstable i.e. by slightly perturbing the joint distribution $\rho(u,z)$, so we hypothesize that the completeness condition is generically satisfied under mild conditions.
\end{remark}

\section{Kernel Proxy Variable }\label{sec:kpv_detailed}
\subsection{\textbf{Notation}}\label{sec:notation}
\begin{enumerate}
    \item As $\mathcal{H}_{\mathcal{P}} \otimes \mathcal{H}_{\mathcal{Q}}$ is isometrically isomorphic to $\mathcal{H}_{\mathcal{P}\mathcal{Q}}$, we use their features interchangeably, i.e. $\phi(p,q)=\phi(p) \otimes \phi(q)$. 
    \item $k(.,.)$ is a general notation for a kernel function, and  $\phi(\cdot)$ denotes RKHS feature maps.
To simplify notation, the argument of the kernel/feature map identifies it: for instance, $k(a,\cdot)$ and $\phi(a)$ denote the respective kernel and feature map on $\mathcal{A}.$
We  denote $K_{a\tilde a}:=k(a,\tilde a)$.
    \item Kernel functions, their empirical estimates and their associated matrices are symmetric, i.e. $K_{ab}=K_{ba}$ and $K^T_{AA}=K_{AA}$. We use this property frequently in our proofs. 
\end{enumerate}

\subsection{\textbf{Problem setting for RKHS-valued $h$}}\label{sec:approx_G}

Recall that to estimate $h$ in \eqref{eq:integral-eq}, KPV aims at estimating $G_h(a,x,z)$ to minimize the empirical risk as:
\begin{align*}
    \tilde{R}(h) = \mathbb{E}_{AXZY}\left[\left(Y - G_h(A,X,Z)\right)^2\right], \text{ where } 
   \quad G_h(a,x,z) := \int_{\mathcal{W}} h(a,x,w)\rho(w\,|\,a,x,z)dw\nonumber
\end{align*}
Since $h\in\mathcal{H}_{AXW}$ by \Cref{ass:h_in_rkhs}, it follows from the reproducing property and the isometric isomorphism between Hilbert space of tensor products and product of Hilbert spaces that:
\begin{eqnarray}
      G_h(a,x,z) &:=& \int_{\mathcal{W}} h(a,x,w)\rho(w\,|\,a,x,z)dw\nonumber\\
      &=& \int_{\mathcal{W}} \langle h, \phi(a,x,w)\rangle_{\mathcal{H}_{AXW}}\rho(w\,|\,a,x,z)dw \nonumber \\
      &=&  \left\langle h, \int_{\mathcal{W}}\phi(a,x,w) \rho(w\,|\,a,x,z)dw\right\rangle_{\mathcal{H}_{\A\X\W}} \nonumber\\
        &=&  \left\langle h, \int_{\mathcal{W}}[\phia(a)\otimes\phix(x)\otimes\phiw(w)] \rho(w\,|\,a,x,z)dw\right\rangle_{\Ha\otimes\Hx\otimes\Hw} \nonumber\\
        &=& \left\langle h, \phia(a)\otimes\phix(x)\otimes\int_{\mathcal{W}}\phiw(w)\rho(w|a,x,z)dw\right\rangle_{\Ha\otimes\Hx\otimes\Hw}\nonumber\\
        &=& \left\langle h, \phia(a)\otimes\phix(x)\otimes\mu_{W|a,x,z}\right\rangle_{\Ha\otimes\Hx\otimes\Hw}
\end{eqnarray}      

where $\mu_{W|a,x,z}$ denotes a conditional mean embedding of $\rho_{W\,|\,a,x,z}$,
and we used the Bochner integrability  \citep[Definition A.5.20]{steinwart2008support} of the feature map $\phi(w)$ to take the expectation inside the dot product (this holds e.g. for bounded kernels).
The regularised empirical risk minimization problem on $\{(\ta,\tx,\tz,\ty)_j \}_{j=1}^{m_2}$ can be expressed as: 
\begin{align}\label{eq:erm3}
 &\widetilde{\eta}_{AXW}=\argmin_{\eta\in\cH_{\A\X\W}}\; \widetilde{L}(\eta) ,\text{ where }\\
  &\widetilde{L}(\eta)= \frac{1}{m}\sum\limits_{j=1}^{m_2}\left( \ty_j-
\left\langle \eta, \phia(\tilde a_j)\otimes\phix(\tilde x_j)\otimes\mu_{W|\tilde a_j,\tilde x_j,\tilde z_j}\right\rangle_{\Ha\otimes\Hx\otimes\Hw}
  \right)^2 \nonumber+\lambda_2 \Vert\eta\Vert^2_{\cH_{\A\X\W}}\nonumber
\end{align}
with $\mu_{W|\ta_j,\tx_j,\tz_j}$  denoting the (true) conditional mean embedding of $\rho_{W|\ta_j,\tx_j,\tz_j}$.
We will equivalently use the notation
$$
\tilde{\eta}_{AXW}[\phia(\ta)\otimes\phix(\tx)\otimes{\mu}_{W|\ta,\tx,\tz}] =
  \left\langle \tilde{\eta}_{AXW} , \phia(\ta)\otimes\phix(\tx)\otimes{\mu}_{W|\ta,\tx,\tz}\right\rangle_{\cH_{\A\X\W}}
$$
to denote the evalation of $\tilde{\eta}_{AXW}$ at $\phia(\ta)\otimes\phix(\tx)\otimes{\mu}_{W|\ta,\tx,\tz}$.

\subsection{A representer theorem expression for  the empirical solution}\label{sec:discussion_on_eta}
\begin{lemma}\label{lem:general_eta}
  Let $\widehat{\eta}_{AXW}$ be an empirical solution of \eqref{eq:erm2}, where
 the population conditional mean
 $\mu_{W|\ta_{j},\tx_{j},\tz_{j}}$
 is replaced by an empirical estimate $\widehat{\mu}_{W|\ta_{j},\tx_{j},\tz_{j}}$ from  \eqref{eq:final_mu}. Then there exists $\alpha\in \R^{m_1\times m_2}$ such that:
\begin{align}\label{eq:general_eta}
        \widehat{\eta}_{AXW} 
   = \sum_{i=1}^{m_1}\sum_{j=1}^{m_2}\alpha_{ij} \phia(\ta_j)\otimes\phix(\tx_j) \otimes \phiw(w_i).
\end{align}
\end{lemma} 
\begin{proof} 
Consider first the solution $\widetilde{\eta}_{AXW}$ of \eqref{eq:erm3}, where a population estimate of the conditional mean embedding $\mu_{W|\ta_j,\tx_j,\tz_j}$ is used in the first stage.
  By the representer theorem \cite{Scholkopf01:Representer},  there exists $\gamma\in \R^{m_2}$ such that
\begin{align}\label{eq:eta_exp1}
    \widetilde{\eta}_{AXW}=\sum^{m_2}_{j=1}\gamma_j\phia(\ta_j)\otimes\phix(\tx_j)\otimes \mu_{W|\ta_j,\tx_j,\tz_j}. 
\end{align}
In practice, we do not have access to the population embedding  $\mu_{W|a,x,z}$. Thus, we substitute in an empirical estimate from \eqref{eq:gamma},\eqref{eq:final_mu}; see Stage 1 in \Cref{sec:kpv_algo_details} for details. The empirical estimate of $\eta$ remains consistent under this replacement, and converges to its population estimate as {\em both} $m_1$ and $m_2$ increase  (\Cref{thm:kpv_final_thm}): see \Cref{sec:kpv_consistency} for the proof. 

Substituting the empirical estimate $\widehat{\mu}_{W|\ta_{j},\tx_{j},\tz_{j}}$ from  \eqref{eq:final_mu}  in place of the population
$\mu_{W|\ta_{j},\tx_{j},\tz_{j}}$ in the empirical squared loss \eqref{eq:erm3}, then $\eta$ appears in a dot product with
\begin{equation}
\sum_{j=1}^{m_{2}}\phi(\tilde{a}_{j})\otimes\phi(\tilde{x}_{j})\otimes\underset{\widehat{\mu}_{W|\tilde{a}_{j},\tilde{x}_{j},\tilde{z}_{j}}}{\underbrace{\left(\sum_{i=1}^{m_{1}}\Gamma_{i}(\ta_{j},\tx_j,\tz_{j})\phi(w_{i})\right)}}=\sum_{i=1}^{m_{1}}\sum_{j=1}^{m_{2}}\Gamma_{i}(\tilde{a}_{j},\tx_j,\tilde{z}_{j})\phi(\tilde{a}_{j})\otimes\phi(\tilde{x}_{j})\otimes\phi(w_{i})\label{eq:spanTrainingStage2-1}
\end{equation}
 In other words, $ \eta$ in the loss is evaluated at $m_1\times m_2$ samples
$(\tilde{a}_{j},\tilde{x}_{j},w_{i})$.  We know from
the representer theorem \cite{Scholkopf01:Representer} that solutions
$\widehat{\eta}_{AXW}$ are written in the span of $\left(\phi(\tilde{a}_{j})\otimes\phi(\tilde{x}_{j})\otimes\phi(w_{i})\right),$ $i\in \{1,\ldots m_1\},$ $j\in \{1,\ldots m_2\}$.
The Gram matrix of these tensor sample features, appropriately rearranged,
is an $(m_{1}m_{2})\times(m_{1}m_{2})$ matrix,
\[
K_{\mathrm{tot}}:=K_{WW}\otimes\left(K_{AA}\odot K_{XX}\right),
\]
where $K$ is the Kroenecker product. Assuming both $K_{WW}$ and
$K_{AA}\odot K_{XX}$ have full rank, then by \citep[eq. 490]{PetPed08},
the rank of $K_{tot}$ is $m_{1}m_{2}$ (in other words, the sample features used to express the
representer theorem solution span a space of dimension  $m_{1}m_{2}$).

It is instructive to note 
that any empirical solution to \eqref{eq:erm2} can hence be written as a linear combination of features of $\cH_{\A\X\W}$, with features of $w$ from sample of the first stage $\{w_i\}_{i=1}^{m_1}$ and features of $a$ and $z$ from the second stage,  $\{\tilde a_j, \tilde x_j\}_{j=1}^{m_2}$. 
\end{proof}

\begin{lemma}\label{lem:eta_norm}
Let $\widehat{\eta}_{AXW}$ be expressed as \eqref{eq:general_eta}.
Then, its squared RKHS norm can be written as:
    \begin{equation}\label{eq:regularisation_term}
     \Vert \widehat{\eta}_{AXW} \Vert^2_{\cH_{\A\X\W}}
    =\sum^{m_1}_{i,r=1}\sum^{m_2}_{j,t=1} \alpha_{ij} \, \alpha_{rt} \,K_{w_iw_r} \,K_{\ta_j \ta_t} K_{\tx_j \tx_t}.
\end{equation}
\end{lemma}
\begin{proof} By using the reproducing property and tensor product properties, we have:
\begin{align}\label{eq:equality_using_tensor_pd}
    &\Vert \widehat{\eta}_{AXW} \Vert^2_{\cH_{\A\X\W}}
    = \left\langle\widehat{\eta}_{AXW} ,\widehat{\eta}_{AXW} \right\rangle_{\mathcal{F}_{\A\X\W}}\nonumber\\
    &= \left\langle \sum_{i=1}^{m_1}\sum_{j=1}^{m_2}\alpha_{ij} \phia(\ta_j)\otimes\phix(\tx_j) \otimes \phiw(w_i),\sum_{r=1}^{m_1}\sum_{t=1}^{m_2}\alpha_{rt} \phia(\ta_t)\otimes\phix(\tx_t) \otimes \phiw(w_r) \right\rangle_{\mathcal{F}_{\A\X\W}}\nonumber\\
    &= \sum^{m_1}_{i,r=1}\sum^{m_2}_{j,t=1} \alpha_{ij} \, \alpha_{rt} \,K_{w_iw_r} \,K_{\ta_j \ta_t} K_{\tx_j \tx_t},
\end{align}
where ${\mathcal{F}_{\A\X\W}}$ denotes the Frobenius (or Hilbert–Schmidt) inner product. In \eqref{eq:equality_using_tensor_pd}, we have used the known property of tensor product: $\ps{a\otimes b,c\otimes d }_{\mathcal{L}^2(\mathcal{H}_1,\mathcal{H}_2)}=\ps{a,c}_{\cH_1}\otimes\ps{b,d}_{\cH_2}$, where $\mathcal{L}^2(\mathcal{H}_1,\mathcal{H}_2)$ is the space of Hilbert-Schmidt operators from $\cH_1$ to $\cH_2$. Note that since $\widehat{\eta}_{AXW}=\phiw(W)\alpha\otimes\phia(\tA,\tX)$, its squared norm can also be written in trace form, as:
\begin{align}\label{eq:trace_eta}
    &\Vert \widehat{\eta}_{AXW} \Vert^2_{\cH_{\A\X\W}}=\left\langle\widehat{\eta}_{AXW} ,\widehat{\eta}_{AXW} \right\rangle_{\mathcal{F}_{\A\X\W}}\nonumber\\
    &= Tr\left\{\alpha^T K_{WW} \alpha (K_{\tA\tA} \odot K_{\tX\tX})\right\}
\end{align}
using the connection between the $\textit{Trace}$ and $\textit{Hilbert Schmidt}$ or $\textit{Frobenius norm}$ and the reproducing property.
\end{proof}

\subsection{An incomplete solution}\label{sec:incomplete}


In this section, we discuss an alternative kernel proximal algorithm
proposed  by \citet{singh2020kernel2},
(Algorithm 4.1 in the \href{https://arxiv.org/abs/2012.10315v2}{revised paper, May 2021}). 
We demonstrate that
this approach does not represent a valid solution under the Representer Theorem,
unlike the  double sum form of \eqref{eq:general_eta}.

\citeauthor{singh2020kernel2} considers a single joint sample $\{(a,z,x,w,y)_{i}\}_{i=1}^{n},$
so that $m_{1}=m_{2}=n$, and writes the Stage 2 KPV regression solution
as a single sum, rather than a double sum,
\begin{align}
\widehat{\eta}_{\mathrm{inc}} & :=\sum_{i=1}^{n}\alpha_{i}\phi(a_{i})\otimes\phi(x_{i})\otimes\phi(w_{i}).\label{eq:wrongStage2-1-1}
\end{align}
Unfortunately, this solution is \emph{incomplete}, and a double sum
is needed for a correct solution. To see this, consider the subspace
spanned by features making up the incomplete solution $\left(\phi(a_{i})\otimes\phi(x_{i})\otimes\phi(w_{i})\right)_{i=1}^{n}$
in (\ref{eq:wrongStage2-1-1}). The Gram matrix for these sample features 
 is
\[
K_{\mathrm{inc}}=K_{WW}\odot K_{AA}\odot K_{XX},
\]
which has size $n\times n,$ and rank at most $n$ (i.e., these features span a space of dimension at most $n$). Conseqently, the
full Representer Theorem solution $\widehat{\eta}_{AXW}$ cannot be expressed in the form
$\widehat{\eta}_{\mathrm{inc}}$.

\subsection{Kernel Proxy Variable Algorithm}\label{sec:kpv_algo_details}

In \Cref{sec:discussion_on_eta}, we obtained a representer theorem for the form of the solution to \eqref{eq:erm3},
in the event that an empirical estimate $\widehat{\mu}_{W|a,x,z}$  is used for the mean embedding $\mu_{W|a,x,z}$, the (true) conditional mean embedding of $\rho_{W|a,x,z}$.

We have two goals for the present section: first, to provide an explicit form for $\widehat{\mu}_{W|a,x,z}$ (\textit{Stage 1}). Second, in  (\textit{Stage 2}), to learn $\widehat{\eta}_{\mathit{AXW}}$, using the empirical embedding  $\widehat{\mu}_{W|a,x,z}$  learned in stage 1. \Cref{thm:kpv_final_thm} show that the empirical estimate of $\eta$ remains consistent under this replacement and converges to its true value at population level, see \Cref{sec:kpv_consistency} for details. Consistent with the two-stages of the algorithm, we assume that the sample is divided into two sub-samples of size $m_1$ and $m_2$, i.e., $\{(a,x,z,w)_i \}_{i=1}^{m_1}$ and
$\{(\ta,\tx,\ty, \tz)_j\}_{j=1}^{m_2}$. 

\textbf{Stage 1.} Estimating Conditional mean embedding operator $\widehat{C}_{W|A,X,Z}$ from the first sample,  $\{(a,x,z,w)_i \}_{i=1}^{m_1}$.

As stated in \Cref{ass:kernel_characteristic}, $k(a, ·), k(x, ·), k(w, ·) $ and  $k(z, ·)$ are characteristic kernels, and are  continuous, bounded by $\kappa > 0$, and   \(\E[\sqrt{k(\cdot,\cdot)}]<\infty\). We may define the \textit{conditional mean embedding operator} as in \citet{song2009hilbert}:
\[C_{W|A,X,Z}:\mathcal{H}_{\A\X\W} \mapsto \Hw, \quad C_{W|A,X,Z} (\phia(a)\otimes \phix(x)\otimes\phiz(z))=\E[\Phiw (W)|a,x,z].\] 

Following 
\citet[Theorem 1]{singh2019kernel}, it can be shown that 
\begin{equation}\label{eq:CMEO}
    \widehat{C}_{W|A,X,Z} = \Phiw({W}) \left[K_{AA} \odot K_{XX} \odot K_{ZZ} + m_1 \lambda_1 I_{m_1} \right]^{-1} \left[\Phi_{\A\X\Z}(A,X,Z)   \right]^T,
\end{equation}
where $K_{AA}, K_{XX}$ and $K_{ZZ}$ are $m_1\times m_1$ kernel matrices and  $\Phiw({W})$ is a vector of $m_1$ columns, with $\phiw(w_i)$ in its $i$th column. By definition \cite{song2009hilbert},  $\widehat{\mu}_{W|a,x,z} :=  \widehat{C}_{W|A,X,Z} \left( \phia(a) \otimes \phix(x) \otimes \phiz(z) \right)$, and therefore
\begin{align}
     \widehat{\mu}_{W|a,x,z}   &= \left[\Phiw({W}) \left[K_{AA} \odot K_{XX}\odot K_{ZZ} + m_1 \lambda_1 \right]^{-1} \left[\Phia({A}) \otimes \Phix({X}) \otimes \Phiz({Z}) \right]^T \right] \left(\phia(a)\otimes\phix(x)\otimes \phiz(z) \right)\nonumber \\
    &= \Phiw({W}) \Gamma(a,x,z), \label{eq:gamma}
\end{align}
where we applied the reproducing property and used isometric isomorphism between Hilbert space of tensor products and product of Hilbert spaces, i.e. \(\Phi_{\A \otimes \X\otimes \Z}(A,X,Z)=\Phia({A}) \otimes \Phix({X}) \otimes \Phiz({Z}) \). We defined $\Gamma(a,x,z)$ as a column matrix with $m_1$ rows :
\begin{equation}\label{eq:gamma_exp}
    \Gamma(a,x,z)=\left[\mathcal{K}_{AXZ} + m_1 \lambda_1 \right]^{-1}\mathcal{K}_{axz}
\end{equation}
where $\mathcal{K}_{AXZ}=K_{{A}A}\odot K_{XX}\odot K_{{Z}Z}$ and  $\mathcal{K}_{axz}=K_{{A}a}\odot K_{Xx}\odot K_{{Z}z}$ are a $m_1 \times m_1$ matrix and a column matrix with $m_1$ rows, respectively. 
Note that for any given $(a,x,z)$, $\mu_{W|a,x,z} \in \textit{Span} \{\Phiw(W)\}$, 
and its empirical estimate can be expressed as
\begin{equation}\label{eq:final_mu}
    \widehat{\mu}_{W|a,x,z} =
    \sum_{i=1}^{m_1} \Gamma_i (a,x,z) \phiw(w_i),\quad \quad  \forall w_i \in \{(a,x,z,w) \}_{i=1}^{m_1}.
\end{equation}
We now detail the second step where we use $\widehat{\mu}_{W|a,x,z}$ to learn the operator $\eta$ to minimize the empirical loss $\eqref{eq:erm2}$.

\textbf{Stage 2.} Expressing $\widehat{\eta}_{AXW}$ using $\{(\ta,\tx,\ty, \tz)_j\}_{j=1}^{m_2}$ and Stage 1.

It follows from \eqref{eq:general_eta} that for any $\{(\ta,\tx,\tz)_j\}_{j=1}^{m_2}\in(\mathcal{A},\mathcal{X},\mathcal{Z})$,
\begin{align}
    \MoveEqLeft \left\langle \widehat{\eta}_{AXW} , \phia(\ta)\otimes\phix(\tx)\otimes\widehat{\mu}_{W|\ta,\tx,\tz}\right\rangle_{\cH_{\A\X\W}}\nonumber \\
    &= \left\langle \sum_{i=1}^{m_1}\sum_{j=1}^{m_2}\alpha_{ij}  \phia(\ta_j)\otimes\phix(\tx_j) \otimes \phiw(w_i), \phia(\ta)\otimes\phix(\tx)\otimes\widehat{\mu}_{W|\ta,\tx,\tz}\right\rangle_{\cH_{\A\X\W}} \nonumber\\
    &= \left\langle \sum_{i=1}^{m_1}\sum_{j=1}^{m_2}\alpha_{ij} \phia(\ta_j)\otimes\phix(\tx_j) \otimes \phiw(w_i), \phia(\ta)\otimes\phix(\tx)\otimes\left\{ \sum_{s=1}^{m_1}\Gamma_s(\ta,\tx,\tz)\phiw(w_s) \right\}\right\rangle_{\cH_{\A\X\W}} \nonumber\\
    &= \left\langle \sum_{i=1}^{m_1}\sum_{j=1}^{m_2}\alpha_{ij} \phia(\ta_j)\otimes\phix(\tx_j) \otimes \phiw(w_i), \sum_{s=1}^{m_1}\Gamma_s(\ta,\tx,\tz)\phia(\ta)\otimes\phix(\tx)\otimes\phiw(w_s)\right\rangle_{\cH_{\A\X\W}} \nonumber \\
    &= \sum_{i=1}^{m_1}\sum_{s=1}^{m_1}\sum_{j=1}^{m_2} \alpha_{ij}\Gamma_s(\ta,\tx,\tz)\left\langle  \phia(\ta_j)\otimes\phix(\tx_j) \otimes \phiw(w_i), \phia(\ta)\otimes\phix(\tx)\otimes\phiw(w_s)\right\rangle_{\cH_{\A\X\W}}\nonumber\\
    &=\sum_{i=1}^{m_1}\sum_{s=1}^{m_1}\sum_{j=1}^{m_2} \alpha_{ij}\Gamma_s(\ta,\tx,\tz) k(w_i,w_s)k(\ta_j, \ta)k(\tx_j,\tx)
\end{align}
where $k(w_i,w_s), k(\ta_j, \ta) \text{ and  }k(\tx_j,\tx)$ denote associated kernels for variables $w,a \text{ and }x$. The second equation follows from \eqref{eq:h_expansion}.
Substituting the expression of $\Gamma_s(\ta,\tx,\tz)$ from \eqref{eq:gamma_exp}, we have for any $({a,x,z})$:

\begin{multline}\label{eq:h_mu_first}
    \widehat{\eta}_{AXW}[\phia(\ta)\otimes\phix(\tx)\otimes\widehat{\mu}_{W|\ta,\tx,\tz}] =
  \left\langle \widehat{\eta}_{AXW} , \phia(\ta)\otimes\phix(\tx)\otimes\widehat{\mu}_{W|\ta,\tx,\tz}\right\rangle_{\cH_{\A\X\W}} \\
=  \sum^{m_1}_{i=1}\sum^{m_2}_{j=1}\sum^{m_1}_{s=1}\alpha_{ij} K_{w_iw_s} \left\{ \left[K_{AA}\odot K_{XX} \odot K_{ZZ} + m_1 \lambda_1 \right]^{-1}\left[K_{A\ta}\odot K_{X\tx}\odot K_{Z\tz} \right]\right\}_s \left[K_{\ta_j\ta}\odot K_{\tx_j\tx}\right]
\end{multline}

with $K_{AA}, K_{XX} \text{ and } K_{ZZ}$, ${m_1 \times m_1}$ matrices of empirical kernels of $A,X$ and $Z$ estimated from sample 1. 

Equation \eqref{eq:h_mu_first} can be written in matrix format as:
\begin{multline}\label{eq:h_mu_matrix}
  \left\langle \widehat{\eta}_{AXW} , \phia(\ta)\otimes\phix(\tx)\otimes\widehat{\mu}_{W|\ta,\tx,\tz}\right\rangle_{\cH_{\A\X\W}}\nonumber=\\
    \left[K_{\ta\tA}\odot K_{\tx\tX}\right]\alpha^TK_{W W} \{\left[K_{AA}\odot K_{XX} \odot K_{ZZ} + m_1 \lambda_w \right]^{-1}\left[K_{A\ta}\odot K_{X\tx}\odot K_{Z\tz} \right]\} 
\end{multline}
This format will be convenient when deriving the closed-form solution for ERM \eqref{eq:erm2}.
Finally, combining from \cref{eq:h_mu_first} and \eqref{eq:regularisation_term},  the ERM \eqref{eq:erm2} can be written as a minimization over \(\widehat{\alpha}\in \R^{m_1\times m_2}\): 
\begin{eqnarray}\label{eq:ERM_alpha}
   \widehat{\alpha}=\argmin_{\alpha \in \R^{m_1 \times m_2}} \widehat{L}(\alpha), \quad \ \quad
   \widehat{L}(\alpha) =\frac{1}{m_2} \sum^{m_2}_{q=1}\left (\ty^q- 
    \sum^{m_1}_{i=1}\sum^{m_2}_{j=1} \alpha_{ij}A^q_{ij}  \right)^2 +\lambda_2  \sum^{m_1}_{i,s=1}\sum^{m_2}_{j,t=1} \alpha_{ij} \, \alpha_{st} \,B^{st}_{ij},
\end{eqnarray}
denoting \(A^q_{ij}\) and \(B^{st}_{ij}\) as 
\begin{eqnarray}
    A^q_{ij}&=&\{ K_{w_iW} \left[K_{AA}\odot K_{XX} \odot K_{ZZ} + m_1 \lambda_1 \right]^{-1}\left[K_{{A}\ta_q}\odot K_{X\tx_q}\odot K_{{Z}\tz_q} \right]\} \left[K_{\ta_j\ta_q}\odot K_{\tx_j\tx_q}\right]\nonumber,\\
    B^{st}_{ij}&=& \,K_{w_iw_s} \,K_{\ta_j \ta_t} K_{\tx_j \tx_t}.\nonumber
\end{eqnarray}



A solution \(\widehat{\alpha}=[\widehat{\alpha}_{ij}]_{m_1 \times m_2}\) can be derived by solving $\pdv{\widehat{L}(\alpha)}{\alpha}=0 $. As such, $\widehat{\alpha}$ is the solution to the system of the $m_1 \times m_2$ linear equations,
\begin{eqnarray}\label{eq: lineaK_equations}
\forall (i,j) \in m_1\times m_2: 
    &&\sum^{m_2}_q  \ty^q\, A^q_{ij} = 
    \sum^{m_1}_s\sum^{m_2}_t\,\hat{\alpha}_{st} \left[  \sum^{m_2}_q \, A^q_{ij}\, A^q_{st} \, +\, m_2\,\lambda_2\, B^{st}_{ij}\right]
\end{eqnarray}
\begin{remark} While the system of equations \eqref{eq: lineaK_equations} is linear, deriving the solution requires inversion of a $m_1m_2 \times m_1m_2$ matrix. With a memory requirement of complexity  $\mathcal{O} (m_1m_2)^2$ and $\mathcal{O}(m_1m_2)^3$, respectively, this is not possible in practice for even moderate sample sizes. We provide a computationally efficient solution in the next section.
\end{remark}

\subsection{Efficient closed-form solution for $\widehat{\eta}_{AXW}$: Proof of \Cref{prop:empiricalFinalForm}}
As we explained in the previous section, deriving a solution for $\alpha$ -- and consequently empirical estimate of $\eta$ -- involves inverting a matrix $\in \R^{m_1m_2 \times m_1m_2}$,  which is too computationally expensive for most applications. 
In this section, we propose an efficient method for finding $\widehat{\alpha}$.  First, we vectorize the empirical loss \eqref{eq:erm2}; second, we employ a \emph{Woodbudy Matrix Identity}. 

\subsubsection{\textbf{Vectorizing ERM \eqref{eq:erm2}}}
The empirical risk, $\widehat{L}(\alpha)$, is a scalar, and it is a function of $\alpha$, a matrix. The idea of this section is to vectorise $\alpha$ as $v:=vec(\alpha)$, and express empirical loss as a function of $v$. Naturally, this requires manipulation both the total expected loss $\E(Y-\widehat{Y})^2$ and the regularisation. In following sections, we show how to express these terms as functions of $v:=vec(\alpha)$.
\begin{lemma}
Vectorizing $\widehat{\alpha}$ as $\widehat{v}:=vec(\widehat{\alpha})$, the ERM   \eqref{eq:ERM_alpha} can be expressed as:
\begin{eqnarray}\label{eq:EMK_vect}
   \widehat{v}=\argmin_{v \in \R^{m_1m_2}} \widehat{L}(v), \quad 
    \widehat{L}(v) =\frac{1}{m_2} \Vert Y- 
     v^T D\Vert_2^2 +\lambda_2  v^T E v 
\end{eqnarray}
Where:
\begin{eqnarray}
    C &=& K_{W W}\Gamma(\tA,\tX,\tZ)=K_{W W} \left[K_{AA}\odot K_{XX} \odot K_{ZZ} + m_1 \lambda_w \right]^{-1}\left[K_{A\tA}\odot K_{X\tX}\odot K_{Z\tZ} \right] \quad \in \R^{ m_1\times m_2}\quad \label{eq:C}\\
    D &=& C  \, \overline{\otimes} \,\left[K_{\tA\tA}\odot K_{\tX\tX}\right]\qquad    \in \R^{{(m_1  m_2)}\times m_2}\quad \label{eq:D}\\
    E&=& K_{WW} \otimes (K_{\tA\tA} \odot K_{\tX\tX}) \quad \in \R^{{m_1m_2}\times {m_1m_2} }\quad\label{eq:E},
\end{eqnarray}
with $\otimes$ and $\overline{\otimes}$ representing tensor (Kronecker) product and tensor product of associated columns of matrices with the same number of columns, respectively. Vectorization is defined with regards to the rows of a matrix.
\end{lemma}
\begin{proof}

The proof proceeds in two steps. Assume $\eta$ can be written as :
\begin{equation}
    \eta 
   = \sum_{i=1}^{m_1}\sum_{j=1}^{m_2}\alpha_{ij} \phia(\ta_j)\otimes\phix(\tx_j) \otimes \phiw(w_i),
\end{equation}
for $\alpha \in \R^{m_1\times m_2}$. 
We first show vectorized form of $\sum\limits_{q=1}^{m}( y_q- \eta[ \phia(a_q)\otimes \phix(x_q)\otimes \widehat{\mu}_{W|a_q,x_q,z_q} ] )^2,$ and then that of the regularization term  $\Vert\eta\Vert^2_{\cH_{\A\X\W}}$. 

\textbf{Step 1. vectorized form of $\sum\limits_{q=1}^{m}( y_q- \eta[ \phia(a_q)\otimes \phix(x_q)\otimes \widehat{\mu}_{W|a_q,x_q,z_q} ] )^2$}

Let $\widehat{v}:=vec_c(\widehat{\alpha})$, where $\widehat{v}$ is the column-wise vectorization of $\hat{\alpha}$. That is, for \({A={\begin{bmatrix}a&b\\c&d\end{bmatrix}}}\) 
, the vectorization is  \(vec_c(A)={\begin{bmatrix}a\\c\\b\\d\end{bmatrix}}.\) It can be shown that  for column-wise vectorization of compatible matrices $K,L\text{ and }M$, we have:
\begin{eqnarray}\label{eq:Roth}
    vec(KLM)=(M^T \otimes K)vec_c(L)
\end{eqnarray}
This equality is known as \textit{Roth’s relationship} between vectors and matrices. See:\citep[Eq. 82]{macedo2013typing} for proof of column-wise vectorization. Now, if as a specific case we define:
\begin{eqnarray}\label{eq:MLK}
    M&:=& K_{WW} \left[K_{AA}\odot K_{XX} \odot K_{ZZ} + m_1 \lambda_w \right]^{-1}\left[K_{A\ta_q}\odot K_{X\tx_q}\odot K_{Z\tz_q} \right] \nonumber\\
    L &:=& \alpha^T\nonumber\\
    K &:=& K_{\ta_q\tA}\odot K_{\tx_q\tX} = \left(K_{\tA\ta_q}\odot K_{\tX\tx_q}\right)^T; \nonumber
\end{eqnarray}
In this case, $KLM$ is scalar and $\in \R$ and $vec(KLM)=vec([KLM]^T)$. 

Subsequently, we can write: 
\begin{eqnarray}\label{eq:Roth2}
vec(KLM)=(M^T \otimes K)vec_c(L)=vec^T_c(L)(M\otimes K^T)  .
\end{eqnarray}
The second equality uses that transposition and conjugate transposition are distributive over the Kronecker product.

By applying \eqref{eq:Roth} to the matrix form of  \cref{eq:h_mu_first}, we obtain:
\begin{eqnarray}\label{eq:vect_eta}
    &&\eta[\phia(\ta_q)\otimes\phix(\tx_q)\otimes\widehat{\mu}_{W|\ta_q,\tx_q,\tz_q}]\nonumber\\
    &=&vec^T(\alpha)\left[\{K_{WW} \left[K_{AA}\odot K_{XX} \odot K_{ZZ} + m_1 \lambda_w \right]^{-1}\left[K_{A\ta_q}\odot K_{X\tx_q}\odot K_{Z\tz_q} \right]\} \otimes \left(K_{\tA\ta_q}\odot K_{\tX\tx_q}\right)\right]
\end{eqnarray}
Notice, that since the column-wise vectorization of a matrix is equal to the row-wise vectorization of its transpose,   $vec_c(\alpha^T)=vec(\alpha):=v$.

To derive the vectorized form of \eqref{eq:erm2},  \eqref{eq:vect_eta} can be expanded for $(\ta_q, \tx_q,\tz_q)$ for all $q \in \{1,...m_2\}$. Note that in \eqref{eq:MLK}, $M$
is the $q$-th column of $C$ defined in \eqref{eq:C}; and $K^\top$ is the $q$-th column of $K_{\tA\tA} \odot K_{\tX\tX}$. To derive the vectorized form of \cref{eq:erm3}, we expand the results of \eqref{eq:MLK} to all columns of underlying matrices. We introduce operator $\overline{\otimes}$ as a column-wise Kronecker product of matrices\footnote{\(A\overline{\otimes}B= A_i \otimes B_i\) for all $i$s, columns of matrices A and B. This operation is equivalent of Kronecker product of columns and requires matrices A and B to have the same number of columns (but they can have a different number of rows). Note that $\Gamma_{(\tA,\tX,\tZ)_q} =\Gamma_{\ta_q,\tx_q,\tz_q}$ and $\{K_{\tA\tA}\odot O_{\tX\tX}\}_q = K_{\tA\ta_q}\odot O_{\tX\tx_q}$, respectively. This operator allows us to express empirical loss in matrix-vector form.}. Note that this operator is in fact the \textit{column-wise Khatri–Rao product}. 

Finally,
\begin{equation}
    \sum_{q=1}^{m_2} \left( y_q - \eta[\phia(\ta_q)\otimes\phix(\tx_q)\otimes\widehat{\mu}_{W|\ta_q,\tx_q,\tz_q}]\right)^2  
=\Vert Y- v^T \, C \overline{\otimes}\left[K_{\tA\tA}\odot K_{\tX\tX}\right]\Vert^2_2 
    =\Vert Y- v^T D\Vert^2_2,
\end{equation}
with $C$ and $D$ defined by \eqref{eq:C} and \eqref{eq:D}.

\textbf{Step 2. Expressing  $\Vert\eta\Vert^2_{\cH_{\A\X\W}}$ in terms of the  vector $v$}

For the regularization term in \eqref{eq:erm2}, we use the expression of the norm of $\eta$ in matrix terms as presented in \eqref{eq:trace_eta}: 
\begin{align}
     \Vert \eta \Vert^2_{\cH_{\A\X\W}}= &Tr\left\{\alpha^T K_{WW} \alpha (K_{\tA\tA} \odot K_{\tX\tX})\right\}    \nonumber\\
     &= vec(\alpha)^T vec(K_{WW} \alpha (K_{\tA\tA} \odot K_{\tX\tX}))\nonumber\\
     &= vec(\alpha)^T \{K_{WW} \otimes (K_{\tA\tA} \odot K_{\tX\tX})^T \}vec(\alpha)\nonumber\\
     & =v^T \{K_{WW} \otimes (K_{\tA\tA} \odot K_{\tX\tX}) \}\, v \\
     & := v^T E v.
\end{align}
Note that the vectorization is row-wise. In the second equality, we used that $Trace(A^TB)=vec(A)^T vec(B)$ for two square matrices $A$ and $B$ of the same size.
The third equality is the row-wise expression of \textit{Roth's relationship} between vectors and matrices (see \citet{macedo2013typing}).\end{proof}


\subsubsection{\textbf{Derivation of the closed form solution for $\widehat{\eta}$}}
We presented the vectorized form of ERM \cref{eq:erm2} in \Cref{eq:EMK_vect}. Its minimizer $\widehat{v}$ is the solution to a ridge regression in $\R^{m_1m_2}$ and its closed-form is easily available through:
\begin{equation}\label{Eq:v_close1}
    \widehat{v} = \left\{\left[D D^T + m_2 \lambda_2 E\right]^{-1} D \right\}y ,
\end{equation}
with $D$ and $E$ given by \eqref{eq:D} and \eqref{eq:E}, respectively. The solution still requires inversion of $D D^T$, an $m_1m_2\times m_1m_2$ matrix, however. In the following, we use the Woodbury identity  to derive an efficient closed-form solution for \cref{eq:erm2}.

\begin{lemma}\label{lem:close_v}
The closed form solution in \cref{Eq:v_close1} can be rearranged as:
\begin{eqnarray}
  \widehat{v} &=&  \left(\Gamma_{(A,X,Z)} \overline{\otimes} I\right) \left( m_2\lambda_2 I + \Sigma \right)^{-1}
    y
    \quad \quad \in \R^{m_1m_2}\label{eq:st2}\\
    \text{  and }\Sigma&=& \left[\left(\Gamma^T_{(\ta_q,\tx_q,\tz_q)} K_{WW}\Gamma_{(\ta_p,\tx_p,\tz_p)}\right)(K_{\ta_q\ta_p}K_{\tx_q\tx_p})
     \right]_{m_2\times m_2}, \text{ for } p, q \in \{1,\dots,m_2\},\label{eq:st2bis}
    \end{eqnarray}
\end{lemma}
where $\Gamma_{(a,x,z)}:=\Gamma(a,x,z)$ is defined in \eqref{eq:gamma_exp}. 
Hence,  the closed-form solution for $v:=vec(\alpha)$  only involves the inversion of an $m_2 \times m_2$ matrix $\Sigma$.  

\begin{proof} We start by applying Woobudy identity to eq. \eqref{Eq:v_close1}:
\begin{eqnarray}\label{Eq:alpha_close}
    \widehat{v} &=& \left\{\left[D D^T + m_2 \lambda_2 E\right]^{-1} D \right\}y    \quad \nonumber\\
    &=& E^{-1} D \left[ m_2 \lambda_2 I + D^T E^{-1}D\right]^{-1}y \label{eq:woodbury}\\
        &=& \left(\Gamma_{(A,X,Z)} \overline{\otimes} I\right) \left( m_2\lambda_2 I + \Sigma \right)^{-1}
    y
    \quad \quad \in \R^{m_1m_2}\label{eq:st2}\\
    \text{  and }\Sigma&=& \left[\left(\Gamma^T_{(\ta_q,\tx_q,\tz_q)} K_{WW}\Gamma_{(\ta_p,\tx_p,\tz_p)}\right)(K_{\ta_q\ta_p}K_{\tx_q\tx_p})
     \right]_{m_2\times m_2}, \text{ for } p, q \in \{1,\dots,m_2\}\label{eq:sigma} 
\end{eqnarray}
The final equality, \eqref{eq:st2}, is the outcome of \cref{lem:st1}.  
\end{proof}

\begin{lemma}\label{lem:st1} We may write
 $D^TE^{-1}D =\Sigma$, where $\Sigma= \left[\left(\Gamma^T_{(\ta_q,\tx_q,\tz_q)} K_{WW}\Gamma_{(\ta_p,\tx_p,\tz_p)}\right)(K_{\ta_q\ta_p}K_{\tx_q\tx_p})
     \right]_{m_2\times m_2}$ , for  $p, q \in \{1,\dots,m_2\}$.
\end{lemma}
\begin{proof}
 We first show that: \(E^{-1}D = \Gamma_{(\tA,\tX,\tZ)}\, \overline{\otimes}\, I_{m_2\times m_2}\)
\begin{eqnarray}\label{Eq:proof_t2}
     E^{-1}D &=& \left(K_{WW} \otimes (K_{\tA\tA} \odot K_{\tX\tX})\right)^{-1}
     \left(K_{WW} \Gamma_{(\tA,\tX,\tZ)}\,\overline{\otimes} \left(K_{\tA\tA}\odot K_{\tX\tX}\right) \right)\nonumber\\
     &=& \left(K_{WW}^{-1} \otimes (K_{\tA\tA} \odot K_{\tX\tX})^{-1}\right)\left(K_{WW} \Gamma_{(\tA,\tX,\tZ)}\,\overline{\otimes} \left(K_{\tA\tA}\odot K_{\tX\tX}\right) \right)\nonumber\\
     &=&\left[
     \dots, 
     \left(K_{WW}^{-1} \otimes (K_{\tA\tA} \odot K_{\tX\tX})^{-1}\right)
     \left(K_{WW} \Gamma_{(\ta_q,\tx_q,\tz_q)}\,\otimes \left(K_{\tA\ta_q}\odot K_{\tX\tx_q}\right) \right)
     , \dots     \right]\nonumber\\
     &=& \left[
     \dots,
     \left( K_{WW}^{-1}K_{WW} \Gamma_{(\ta_q,\tx_q,\tz_q)}\right)
     \,\otimes\, 
     \left((K_{\tA\tA} \odot K_{\tX\tX})^{-1} \left(K_{\tA\ta_q}\odot K_{\tX\tx_q}\right)
     \right)
     ,\dots
     \right]\nonumber\\&=&
     \left[\dots,
     \Gamma_{(\ta_q,\tx_q,\tz_q)} 
     \,\otimes\,
     I_q
     ,\dots\right]= \Gamma_{(\tA,\tX,\tZ)}     \overline{\otimes}\, I_{m_2\times m_2}\nonumber
\end{eqnarray}
For the third equality, we expand $\overline{\otimes}$ in terms of the Kronecker product of associated columns of matrices. We then use the property of the Kronecker product $(A\otimes B)(C \otimes D)=(AC) \otimes (BD)$ for compatible $A,B,C, \text{ and } D$.

In the second step, we replace $E^{-1}D$ with its equivalent derived in step one, and show that:  \(D^T(E^{-1}D)= D^T\left(\Gamma_{(\tA,\tX,\tZ)}\, \overline{\otimes}\, I_{m_2\times m_2}\right)=
     \left[\left(\Gamma^T_{(\ta_q,\tx_q,\tz_q)} K_{WW}\Gamma_{(\ta_p,\tx_p,\tz_p)}\right)(K_{\ta_q\ta_p}K_{\tx_q\tx_p})
     \right]_{m_2\times m_2}\). First,

    \begin{eqnarray}
    &&D^T\left(\Gamma_{(\tA,\tX,\tZ)}\, \overline{\otimes}\, I_{m_2\times m_2}\right)=\left\{C^T \overline{\otimes} \left(K_{\tA\tA}\odot K_{\tX\tX}\right)\right\} \left(\Gamma_{(\tA,\tX,\tZ)}\, \overline{\otimes}\, I_{m_2\times m_2}\right)\nonumber\\
    &&=\left\{\left(K_{W W}\Gamma_{(\tA,\tX,\tZ)}\right)^T \overline{\otimes}\left(K_{\tA\tA}\odot K_{\tX\tX}\right)\right\} \left(\Gamma_{(\tA,\tX,\tZ)}\, \overline{\otimes}\, I_{m_2\times m_2}\right)\nonumber
    \end{eqnarray}
    
Next, let's take a closer look at individual elements of the matrix $[.]_{qp}$, the $q$th row of $p$th column.
    \begin{eqnarray}
     &&\left[D^T\left(\Gamma_{(\tA,\tX,\tZ)}\, \overline{\otimes}\, I_{m_2\times m_2}\right)\right]_{qp}=\left\{\left(\Gamma^T_{(\ta_q,\tx_q,\tz_q)}K_{W W}\right)  {\otimes}\left(K_{\tA\ta_q}\odot K_{\tX\tx_q}\right)\right\} \left(\Gamma_{(\ta_p,\tx_p,\tz_p)}\, \otimes\, I_{p}\right)\label{eq:36}\\
    &=&\left(\Gamma^T_{(\ta_q,\tx_q,\tz_q)}K_{W W}\Gamma_{(\ta_p,\tx_p,\tz_p)}\right)  \otimes \left(\left(K_{\tA\ta_q}\odot K_{\tX\tx_q}\right) I_{p}\right)\\
    &=&\left(\Gamma^T_{(\ta_q,\tx_q,\tz_q)}K_{W W}\Gamma_{(\ta_p,\tx_p,\tz_p)}\right)(K_{\ta_q\ta_p}K_{\tx_q\tx_p})
     \end{eqnarray}
In \eqref{eq:36} we have used the property of Kronecker product $(A\otimes B)(C \otimes D)=(AC) \otimes (BD)$ for compatible $A,B,C, \text{ and } D$. 
\end{proof}
\subsection{Estimating the causal effect}
Recall that the causal effect \eqref{Eq:average_causal_effect} is written \(\beta(a) = {\int_{\X,\W} h(a,x,w)f(x,w) dx dw}\) . Since $h\in \cH_{\A\X\W}$ by \Cref{ass:h_in_rkhs}, and using the reproducing property, we can write:
 \begin{eqnarray}
    \beta(a) &=& {\int_{\X,\W} h(a,x,w)\rho(x,w) dx dw}\nonumber \\
    &=& \int_{\X,\W}\langle h, \phia(a)\otimes \phix(x)\otimes \phiw(w)\rangle_{\cH_{\A\X\W}} \rho(x,w) dx dw\nonumber\\
    &=& \ps{h, \phia(a) \otimes \int_{\X\W}\phix(x)\otimes \phiw(w)\rho(x,w) dx dw}_{\cH_{\A\X\W}}.
\end{eqnarray}
 Consequently,  $\widehat{h}$ can be expressed as: \(\widehat{h}=\sum\limits^{m_1}_{i=1}\sum\limits^{m_2}_{j=1} \widehat{\alpha}_{ij} \phia(\ta_j) \otimes \phix(\tx_j)\otimes \phiw(w_i)\). We can further replace $\int_{\X\W}\phix(x)\otimes \phiw(w)\rho(x,w) dx dw$ by its empirical estimate $\frac{1}{n}\sum_{k=1}^n \phix(x_k)\otimes \phiw(w_k)$ from the sample $\{(x,w)_k\}_{k=1}^n$. This leads to the following estimator of the causal effect:
 \begin{eqnarray}
    \widehat{\beta}(a) 
   &=&\ps{ \widehat{h},\phia(a) \otimes \int_{\X\W}\frac{1}{n}\sum_{k=1}^n \phix(x_k)\otimes \phiw(w_k)}_{\cH_{\A\X\W}}\nonumber\\ 
    &=& \langle \sum^{m_1}_{i=1}\sum^{m_2}_{j=1}
    \widehat{\alpha}_{ij} \phia(\ta_j) \otimes \phix(\tx_j)\otimes \phiw(w_i), \phia(a)\otimes \frac{1}{n}\sum_{k=1}^n \phix(x_k)\otimes \phiw(w_k)\rangle_{\cH_{\A\X\W}}\nonumber\\
    &=& \frac{1}{n}\sum^{m_1}_{i=1}\sum^{m_2}_{j=1}\sum^{n}_{k=1}\widehat{\alpha}_{ij}K_{a\ta_j}K_{x_k\tx_j}K_{w_kw_i}.
\end{eqnarray}

\subsection{Algorithm}
See full implementation of the Kernel Proxy Method at \renewcommand\UrlFont{\color{blue}}\url{ https://github.com/Afsaneh-Mastouri/KPV}.

\subsection{An alternative two-stage solution, and its shortcomings}\label{sec:RS_approach}


\citet{singh2020kernel2}
proposed an alternative solution to kernel proximal causal learning 
(Algorithm 4.1 in the \href{https://arxiv.org/abs/2012.10315v1}{initial work, December 2020}), which directly employs \citep[Algorithm 1]{singh2019kernel}. 
\citeauthor{singh2020kernel2}
directly used the Stage 1 estimate of $\widehat{\mu}_{W,A|A,X,Z}\left(\phi(a)\otimes\phi(x)\otimes\phi(z)\right):=\widehat{\mathbb{E}}\left(\phi(W)\otimes\phi(A)\vert a,x,z\right)$, obtained by ridge regression, 
as an input in Stage 2,
which would allow an unmodified use of the KIV algorithm \cite{singh2019kernel} in the proxy setting.
We now show that this method is incorrect, as it does not satisfy the required conditions for consistency,
and suffers from related shortcomings in practice.


Theoretically, regression from $\phi(a)$ to $\phi(a)$ is, in population limit, the identity mapping $I_{\mathcal{H_{A}}}$ from $\mathcal{H_{A}}$ to $\mathcal{H_{A}}$. This operator is not Hilbert-Schmidt for characteristic RKHSs, and violates the well-posedness assumption for consistency of Stage 1 regression  \citep{singh2019kernel}.


\begin{figure}
\begin{minipage}{.5\textwidth}
  \centering
  \includegraphics[width=\linewidth]{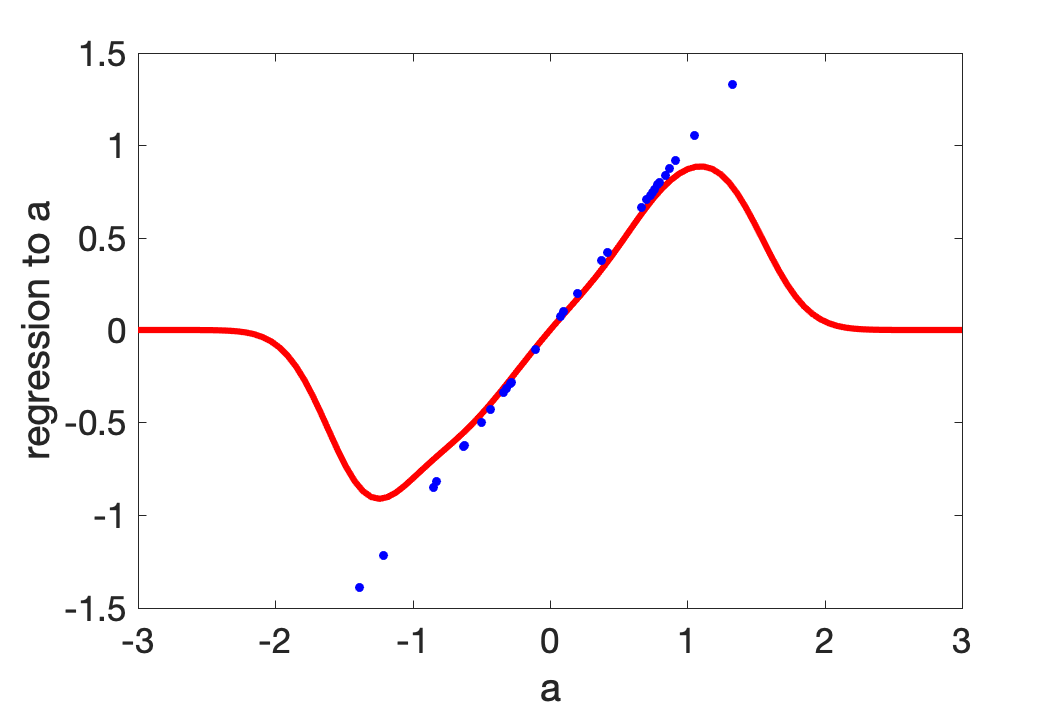} 
  \captionof{figure}{Learning an identity map on a non-compact 
   \\ domain using  (Gaussian) kernel ridge regression.}
  \label{fig:Identity}
\end{minipage}%
\begin{minipage}{.5\textwidth}
  \centering
  \includegraphics[width=\linewidth]{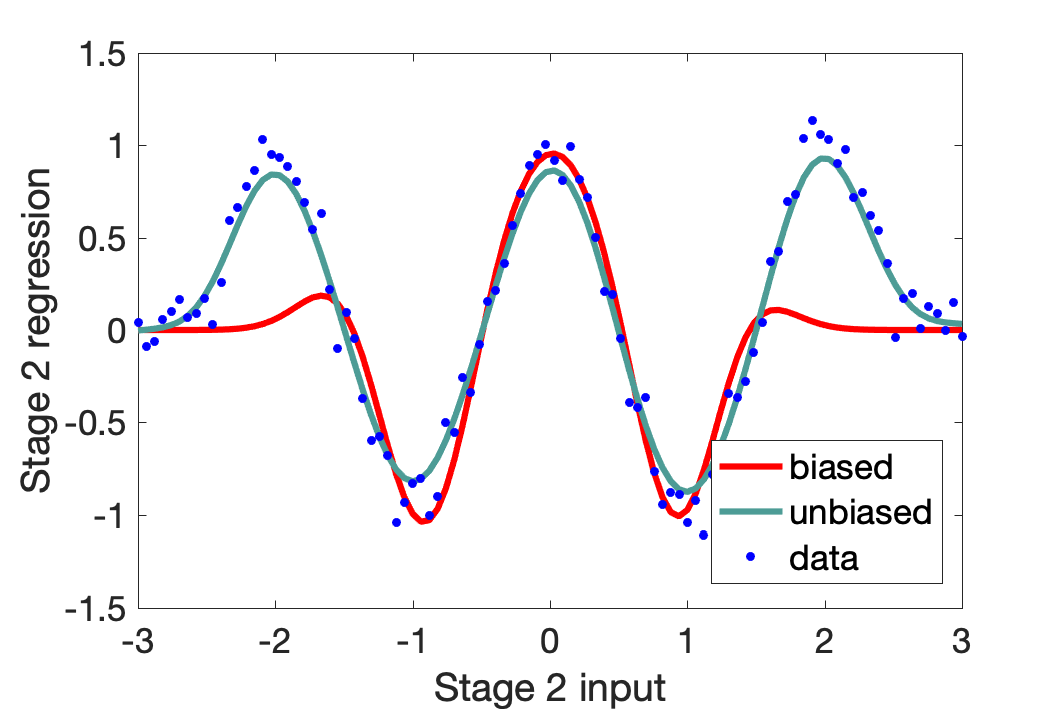}  
  \caption{Bias in second stage as a result of using $\hat\mu_{A|a}$ 
    in Stage 1 (``biased'') vs regressing on $\phi(a)$  (``unbiased'').}
  \label{fig:test2}
\end{minipage}
\end{figure}

In practice, predicting $\phi(a)$ via ridge regression from $\phi(a)$ introduces bias in the finite sample setting. This is shown in
an example in Figures \ref{fig:Identity} and \ref{fig:test2}.
In a first stage (\Cref{fig:Identity}), the identity map is approximated by ridge regression, where the distribution  $\rho_{\mathcal{A}}(a)$ is Gaussian centred at the origin. 
This distribution is supported on the entire real line, but for finite samples, few points are seen at the tails, and bias is introduced (the function reverts to zero). The impact of this bias will reduce as more training samples are observed (although the identity map will never be learned perfectly, as discussed earlier).
 This bias affects the second stage.
 In  \Cref{fig:test2}, the distribution of $a$ for the second stage is uniform on the interval $[-3,3]$. This is a {\em subset} of the stage 1 support of $\rho_{\mathcal{A}}(a)$, 
 yet due to the limited number of samples from stage 1, bias is nonetheless introduced
 near the boundaries of that interval. This bias can be more severe as the dimension of $a$ increases.
As seen in \Cref{fig:test2}, this bias impacts the second stage, where we compare regression from $\hat\mu_{A|a}$  to $y$  ({\em biased}) with regression from $\phi(a)$ to $y$ ({\em unbiased}).
This bias is avoided in our KPV setting by using the Stage 2 input
 $\mu_{W|a,z}\otimes \phi(a)$ instead of $\mu_{W,A|a,z}$ (ignoring $x$ for simplicity). 

\subsection{Consistency}\label{sec:kpv_consistency}

In this section, we provide consistency results for the KPV approach. 
For any Hilbert space $\mathcal{F}$, we denote $\cL(\mathcal{F})$ the space of bounded linear operators from $\mathcal{F}$ to itself.
For any Hilbert space $\mathcal{G}$, we denote by $\cL^2(\mathcal{F},\mathcal{G})$ the space of Hilbert-Schmidt operators from $\mathcal{F}$ to $\mathcal{G}$. 
We denote by $L^2(\mathcal{F},\rho)$ the space of square integrable functions on $\mathcal{F}$  with respect to measure $\rho$.

\subsubsection{\textbf{Theoretical guarantees for Stage 1}}\label{sec:proof_stage_1}


The optimal $C_{W|X,A,Z}$ minimizes the expected discrepancy:
\begin{equation*}
    C_{W|X,A,Z} = \argmin_{C \in \cL^2( \cH_{AXZ},\Hw)} E(C), \text{ where }E(C)=\E_{WAXZ}\| \phiw(W) -C\phi(A,X,Z)\|_{\Hw}^2
\end{equation*}
We now provide a non-asymptotic consistency result for Stage 1.
This directly follows the Stage 1 IV proof of \citet{singh2019kernel}, based in turn on the regression result of \citet{smale2007}, and we simply state the main results as they apply in our setting, referencing the relevant theorems from the earlier work as needed. 

The problem of learning $C_{W|A,X,Z}$ is transformed into a vector-valued regression, where the search space is the vector-valued RKHS $\cH_{\Gamma}$ of operators mapping  $\cH_{\A\X\Z}$ to $\Hw$. A crucial result is that $ \cH_{\A\X\Z}\otimes \Hw$ is isomorphic to $\mathcal{L}^2(\cH_{\A\X\Z},\Hw)$. Hence, by choosing the vector-valued kernel $\Gamma$ with feature map : $(a,x,z,w)\mapsto [ \phia(a)\otimes \phix(x)\otimes \phiz(z)\otimes \phiw(w)]:=\phia(a)\otimes \phix(x)\otimes \phiz(z) \ps{\phiw(w),\cdot}_{\Hw}$, we have $\cH_{\Gamma}=\mathcal{L}^2(\cH_{\A\X\Z},\Hw)$ and they share the same norm. We denote by $L^2(\A\times\X\times\Z,\rho_{\A\X\Z})$ the space of square integrable functions from $\A\times\X\times \Z$ to $\W$ with respect to measure $\rho_{\A\X\Z}$, where $\rho_{\A\X\Z}$ is the restriction of $\rho$ to $\A\times\X\times\Z$.
\begin{assumption}\label{ass:well_posed1}
Suppose that $C_{W|X,A,Z}\in \cH_{\Gamma}$, i.e. $C_{W|X,A,Z}=\argmin_{C \in \cH_{\Gamma}}E(C)$.
\end{assumption}

\begin{definition}[Kernel Integral operator for Stage 1]
Define the integral operator :
\begin{align*} 
S_1 \colon L^2(\A\times\X\times \Z,\rho_{\A\X\Z})& \longrightarrow \cH_{\A\X\Z}\\
g&\longmapsto \int \phi(a,x,z)g(a,x,z)d\rho_{\A\X\Z}(a,x,z).
\end{align*}
The uncentered covariance operator is defined by $T_1=S_1\circ S_1^*$, where $S_1^*$ is the adjoint of $S_1$.
\end{definition}

\begin{assumption}\label{ass:source1}
Fix $\gamma_1<\infty$. For given $c_2 \in (1, 2]$, define the prior $\cP(\gamma_1, c_1)$ as the set of probability
distributions $\rho$ on $\A \times \X \times \Z \times \W$ such that a range space assumption is satisfied : $\exists G_1 \in \cH_{\Gamma}$ s.t. $C_{W|A,X,Z} =T_1^{\frac{c_1-1}{2}}\circ G_1$ and $\|G_1\|^2_{\cH_{\Gamma}}\le \gamma_1$.
\end{assumption}

Our estimator for $C_{W|AXZ}$ is given by ERM \eqref{eq:erm1} based on $\{(a,x,z,w)_i \}_{i=1}^{m_1}$.
The following theorem provides the closed-form solution of \eqref{eq:erm1}.

\begin{thm}\citep[Theorem 1]{singh2019kernel}
For any $\lambda_1>0$, the solution of \eqref{eq:erm1} exists, is unique, and is given by:
\begin{align*}
    \widehat{C}_{W|A,X,Z} = (\boldsymbol{T}_1 +\lambda_1)^{-1}g_1, &\text{ where } \boldsymbol{T}_1=\frac{1}{m_1}\sum_{i=1}^{m_1}\phi(a_i,x_i,z_i)\otimes \phi(a_i,x_i,z_i),\\ &\text{ and } g_1 = \frac{1}{m_1} \sum_{i=1}^{m_1} \phi(a_i,x_i,z_i)\otimes \phiw(w_i);
\end{align*}
and for any $(a,x,z)\in \A\times\X\times\Z$, we have $\widehat{\mu}_{W|a,x,z} =  \widehat{C}_{W|A,X,Z} \left( \phia(a) \otimes \phix(x) \otimes \phiz(z) \right)$.
\end{thm}

Under the assumptions provided above, we can now derive a non-asymptotic bound in high probability for the estimated conditional mean embedding, for a well-chosen regularization parameter.

\begin{thm}\label{thm:kpv_s1_bound}
Suppose \Cref{ass:polish_spaces,ass:kernel_characteristic,ass:well_posed1,ass:source1} hold. Define $\lambda_1$ as:
\begin{equation*}
    \lambda_1=\left(\frac{8\kappa^3(\kappa+ \kappa^3 \|C_{W|A,X,Z} \|_{\cH_{\Gamma}})\ln(2/\delta)}{\sqrt{m_1\gamma_1(c_1-1)}}\right)^{\frac{2}{c_1+1}}
\end{equation*}
Then, for any $x,a,z\in \A\times\X\times\Z$ and any $\delta \in (0,1)$, the following holds with probability $1-\delta$:
\begin{equation*}
    \| \widehat{\mu}_{W|{a,x,z}}  - \mu_{W|{a,x,z}}\|_{\Hw} \le \kappa^3 r_{C}(\delta,m_1,c_1) \eqdef \kappa^3 \frac{\sqrt{\gamma_1}(c_1+1)}{4^{\frac{1}{c_1+1}}}\left(\frac{4\kappa^3(\kappa+\kappa^3\|C_{W|A,X,Z} \|_{\cH_{\Gamma}})\ln(2/\delta)}{\sqrt{m_1\gamma_1}(c_1-1)}\right)^{\frac{c_1-1}{c_1+1}},
\end{equation*}
where $\widehat{\mu}_{W|a,x,z} =  \widehat{C}_{W|A,X,Z} \left( \phia(a) \otimes \phix(x) \otimes \phiz(z) \right)$ and $\widehat{C}_{W|A,X,Z}$ is the solution of \eqref{eq:erm1}.
\end{thm}

\begin{proof} Under \Cref{ass:polish_spaces} and \ref{ass:kernel_characteristic}, $\Ha,\Hx,\Hz$ are separable (see Lemma 4.33 of \citet{steinwart2008support}). Hence, for any $(a,x,z)\in \A\times\X\times\Z$, we have :
$\| \phia(a)\otimes \phix(x)\otimes \phiz(z)\|_{\cH_{\A\X\Z}} =\| \phia(a)\|_{\cH_{\A}}\|  \phix(x)\|_{\cH_{\X}}\|  \phiz(z)\|_{\cH_{\Z}}\le \kappa^3$ by \Cref{ass:kernel_characteristic}.
Then, we can write:
\begin{align*}
    \|\widehat{\mu}_{W|a,x,z}-\mu_{W|{a,x,z}}\|_{\Hw} &=  \|(\widehat{C}_{W|A,X,Z}-C_{W|A,X,Z}) \left( \phia(a) \otimes \phix(x) \otimes \phiz(z) \right)\|_{\Hw}\\
    &\le \|\widehat{C}_{W|A,X,Z}-C_{W|A,X,Z}\|_{\cH_{\Gamma}}\|\phia(a) \otimes \phix(x) \otimes \phiz(z)\|_{\cH_{\A\X\Z}}\\
    &\le \kappa^3 r_{C}(\delta,m_1,c_1)
\end{align*}
where the last inequality results from \citet[Theorem 2]{singh2019kernel}. 
\end{proof}

\subsubsection{\textbf{Theoretical guarantees for Stage 2}}\label{sec:proof_stage_2}


The optimal $\eta$ minimizes the expected discrepancy:
\begin{equation*}
  \eta_{AXW} = \argmin_{\eta \in \cH_{\A\X\W}}\tilde{R}(\eta), \text{ where }\tilde{R}(\eta)=\E_{AXZY}\left\{ Y- \eta[ \phia(a,x)\otimes \mu_{W|a,x,z} ] \right\}^2.
\end{equation*}

Similarly to Stage 1, the problem of learning $\eta_{AXW}$ is transformed into a ridge regression, where the search space is the RKHS $\cH_{\A\X\W}$ of $\Y$-valued functions ($\Y\subset \R$). We now provide our assumptions to derive non asymptotic results for Stage 2.
The approach builds  on the Stage 2 proof of \citet{singh2019kernel}, based in turn on \citep{caponnetto2007optimal,szabo2016learning},
with modifications made to account for the difference in setting, since the input to our Stage 2
differs from the case of instrumental variable regression (see proofs for details).

\begin{assumption}\label{ass:well_posed2}
Suppose that $\eta_{AXW}\in \cH_{\A\X\W}$, i.e. $\eta_{AXW} =\argmin_{\eta \in \cH_{\A\X\W}}\tilde{R}(\eta)$.
\end{assumption}

\begin{definition}[Kernel integral operator for Stage 2] Define the integral operator :
\begin{align*} 
  S_2 \colon \cH_{\A\X\W} & \longrightarrow \cH_{\A\X\W}\\
  \eta&\longmapsto \int [\mu_{W|a,x,z}\otimes \phi(a,x)]\eta[\phi(a,x)\otimes \mu_{W|a,x,z}]d\rho_{\cH_{\W}\times\A \times\X}(\mu_{W|a,x,z},a,x).
\end{align*}
The uncentered covariance operator is defined by $T_2=S_2\circ S_2^*$, where $S_2^*$ is the adjoint of $S_2$.
\end{definition}

\begin{assumption}\label{ass:source2}
Fix $\gamma_2<\infty$. For given $c_2 \in (1,2]$, define the prior $\cP(\gamma_2,b_2,c_2)$ 
as the set of probability distributions $\rho$ on $\cH_{\A\X\W}\times \Y$ such that:
\begin{itemize}
    \item A range space assumption is satisfied : $\exists G_2\in \cH_{\A\X\W}$ s.t. $\eta_{AXW} = T_2^{\frac{c_2-1}{2}}\circ G_2$ and $\|G_2\|_{\cH_{\A\X\W}}\le \gamma_2$
    \item The eigenvalues $(l_k)_{k\in \mathbb{N}^*}$of $T_2$ satisfy $\alpha_2\le l_k  k^{-b_2}\le \beta_2$ for $b_2>1$, $\alpha_2,\beta_2>0$.
    \end{itemize} 
\end{assumption}

\begin{thm}\label{thm:kpv_consistency}
Assume \Cref{ass:polish_spaces,ass:kernel_characteristic,ass:y_bounded,ass:well_posed1,ass:source1,ass:well_posed2,ass:source2} hold. 
Assume the assumptions of \Cref{thm:kpv_s1_bound} hold and define $\lambda_1$ accordingly. Assume also that $m_1,m_2$ are large enough (see \Cref{prop:bound_T2_T2hat}) and that $\lambda_2\le \|T_2\|_{\cL(\cH_{\A\X\W})}$. Then, for any $\epsilon,\delta\in (0,1)$, the following holds w.p. $1-\epsilon-\delta$:
\begin{multline*}
    \tilde{R}(\widehat{\eta}_{AXW})-\tilde{R}(\eta_{AXW})\le r_H(\delta,m_1,c_1,\epsilon, m_2, b_2,c_2):= 5 \left\{ \frac{4\kappa^{10}  c_Y^2}{\lambda_2}r_C(\delta,m_1,c_1)^2 \right.\\ 
    +\frac{4\kappa^{10}  c_Y^2}{\lambda_2}r_C(\delta,m_1,c_1)^2.4\left( \frac{32\ln^2(6/\epsilon)}{\lambda_2} \left[\frac{(c_Y + \| \eta_{AXW}\|_{\cH_{\A\X\W}})^2(4+m_2\lambda_2( \beta_2^{\frac{1}{b_2}} \frac{\pi/b_2}{sin(\pi/b_2)}\lambda_2^{-\frac{1}{b_2}})}{m_2^2\lambda_2}\right]+\right.\\
    \left.\frac{32 \ln^2(6/\epsilon)}{\lambda_2}\left[ \frac{4\gamma_2\lambda_2^{c_2-1}+m_2\gamma_2\lambda_2^{c_2}}{m_2^2\lambda_2}\right] +  \gamma_2\lambda_2^{c_2-1}+ \|\eta_{AXW}\|^2_{\cH_{\A\X\W}} \right) +\gamma_2\lambda_2^{c_2}    \left.\right.\\
    \left.+32 \ln^2(6/\epsilon) \left[\frac{(c_Y + \| \eta_{AXW}\|_{\cH_{\A\X\W}})^2(4+m_2\lambda_2( \beta_2^{\frac{1}{b_2}} \frac{\pi/b_2}{sin(\pi/b_2)}\lambda_2^{-\frac{1}{b_2}}))}{m_2^2\lambda_2}\right]+ 8 \ln^2(6/\epsilon)\left[ \frac{4\gamma_2\lambda_2^{c_2-1}+m_2\gamma_2\lambda_2^{c_2}}{m_2^2\lambda_2}\right]
    \right\}.
\end{multline*}
\end{thm}
\begin{proof}
By \Cref{prop:risk_decomposition}, we have:
\begin{equation*}
   \tilde{R}(\widehat{\eta}_{AXW})-\tilde{R}(\eta_{AXW})\le 5\left[ S_{-1} + S_0 + \A(\lambda_2) + S_1+S_2 \right]. 
\end{equation*}
Then, by \Cref{prop:bound_S_minus1_0}, w.p. $1-\frac{\epsilon}{3}-\delta$, we have:
\begin{equation*}
    S_{-1}\le \frac{4}{\lambda_2} \kappa^{10} r_C(\delta,m_1,c_1)^2  c_Y^2,\quad 
 S_0 \le \frac{4}{\lambda_2}\kappa^{10}r_C(\delta,m_1,c_1)^2 \|\widetilde{\eta}_{AXW}\|^2_{\cH_{\A\X\W}}
\end{equation*}
where by \Cref{prop:bound_eta_tilde}, w.p. $1-\frac{2\epsilon}{3}$:
\begin{align*}
        &\|\widetilde{\eta}_{AXW}\|^2_{\cH_{\A\X\W}}\\
        &\le
        4\left( \frac{32\ln^2(6/\epsilon)}{\lambda_2} \left[\frac{(c_Y + \| \eta_{AXW}\|_{\cH_{\A\X\W}})^2(4+m_2\lambda_2\cN(\lambda_2))}{m_2^2\lambda_2}\right]+
    \frac{32 \ln^2(6/\epsilon)}{\lambda_2}\left[ \frac{4\cB(\lambda_2)+m_2\cA(\lambda_2)}{m_2\lambda_2}\right] +  \cB(\lambda_2)+ \|\eta_{AXW}\|^2_{\cH_{\A\X\W}} \right).
\end{align*}
Also, by \Cref{prop:bound_S1_S2}, w.p. $1-\frac{2\epsilon}{3}$, we have:
\begin{equation*}
    S_1\le 32 \ln^2(6/\epsilon) \left[\frac{(c_Y + \| \eta_{AXW}\|_{\cH_{\A\X\W}})^2(4+m_2\lambda_2\cN(\lambda_2))}{m_2^2\lambda_2}\right],\quad
    S_2\le 8 \ln^2(6/\epsilon)\left[ \frac{4\cB(\lambda_2)+m_2\cA(\lambda_2)}{m_2^2\lambda_2}\right].
\end{equation*}
Finally, by \Cref{prop:bound_operator_quantities},
\begin{equation*}
    \cA(\lambda_2)\le \gamma_2 \lambda_2^{c_2},\quad\cB(\lambda_2)\le \gamma_2 \lambda_2^{c_2-1}, \quad \cN(\lambda_2)\le \beta_2^{\frac{1}{b_2}} \frac{\pi/b_2}{sin(\pi/b_2)}\lambda_2^{-\frac{1}{b_2}}.
\end{equation*} 
Combining all the probabilistic bounds yields the final result.
\end{proof}

\textbf{Proof of \Cref{thm:kpv_final_thm}}.
\begin{proof}
Ignoring constants in \Cref{thm:kpv_consistency}, we have:
\begin{align*}
    S_{-1}&=O\left(\frac{r_C(\delta,m_1,c_1)^2}{\lambda_2}\right),\\
    &S_0=O\left(\frac{r_C(\delta,m_1,c_1)^2}{\lambda_2}.\left(\frac{1}{m_2^2\lambda_2^2}+\frac{1}{m_2\lambda_2^{1+1/b_2}}+\frac{1}{m_2^2\lambda_2^{3-c_2}}+\frac{1}{m_2\lambda_2^{2-c_2}}+\lambda_2^{c_2-1}+1\right)\right)\\
    &\mathcal{A}(\lambda_2)=O(\lambda_2^{c_2}),\quad S_1=O\left(\frac{1}{m_2^2\lambda_2}+\frac{1}{m_2\lambda_2^{1/b_2}}\right),\quad S_2=O\left(\frac{1}{m_2^2\lambda_2^{2-c_2}}+\frac{1}{m_2\lambda_2^{1-c_2}}\right).
\end{align*}
The last term in $S_0$ indicates that $S_0$ dominates $S_{-1}$. Moreover, since $b_2>1$ and $c_2\in (1,2]$, we have that $\frac{1}{m_2}$ dominates $\frac{1}{m_2\lambda_2^{3-c_2}}$; that $\frac{1}{m_2\lambda_2^{1+1/b_2}}$ dominates $\frac{1}{m_2\lambda_2^{2-c_2}}$; and that $1$ dominates $\lambda_2^{c_2-1}$ (since $\lambda_2\to 0$). For the same reasons, $S_1$ dominates $S_2$. 

Hence, we have:
\begin{equation*}
    \tilde{R}(\widehat{\eta}_{AXW})-\tilde{R}(\eta_{AXW})=O\left( \frac{r_C(\delta,m_1,c_1)^2}{\lambda_2} \left[ \frac{1}{m_2^2\lambda_2^2} +\frac{1}{m_2\lambda_2^{1+1/b_2}}+1\right] +\lambda_2^{c_2}+\frac{1}{m_2^2\lambda_2}+\frac{1}{m_2\lambda_2^{1/b_2}}\right).
\end{equation*}

By \Cref{thm:kpv_s1_bound}, and by choosing $m_1=m_2^{\zeta \frac{c_1+1}{c_1-1}}$ as stated in \Cref{thm:kpv_final_thm}, we have successively:
\begin{equation*}
    r_C(\delta,m_1,c_1)^2=O\left(m_1^{-\frac{c_1+1}{c_1-1}} \right)=O(m_2^{-\zeta}),
\end{equation*}
which leads to:
\begin{equation*}
    \tilde{R}(\widehat{\eta}_{AXW})-\tilde{R}(\eta_{AXW})=O\left( \frac{1}{m_2^{2+\zeta}\lambda_2^3} +\frac{1}{m_2^{1+\zeta}\lambda_2^{2+1/b_2}}+\frac{1}{m_2^{\zeta}\lambda_2}
    +\lambda_2^{c_2}+\frac{1}{m_2^2\lambda_2}+\frac{1}{m_2\lambda_2^{1/b_2}}\right).
\end{equation*}
The final result is from \citet[Theorem 5]{szabo2016learning}.
\end{proof}

We next introduce analogous results  which will be used in proving \Cref{prop:rate_kpv_ce} (see \Cref{sec:proof_rate_kpv_ce}).  The relations in \Cref{cor:kpv_consistency} and \Cref{thm:kpv_final_thm_rkhs} provide convergence rates  in the RKHS norm, rather than the $L_2$ norm. Since the RKHS norm gives stronger guarantees (namely, that norm convergence implies pointwise convergence), we pay  a penalty, which takes the form of an additional $\lambda_2^{-1}$ appearing in certain of the terms (as compared with the \Cref{thm:kpv_final_thm} proof).

\begin{corollary}
    \label{cor:kpv_consistency}
Suppose the assumptions of \Cref{thm:kpv_consistency} hold. Then, for any $\epsilon,\delta\in (0,1)$, the following holds w.p. $1-\epsilon-\delta$:
\begin{multline*}
    \|\widehat{\eta}_{AXW}-\eta_{AXW}\|^2_{\cH_{\A\X\W}}\le \tilde{r}_H(\delta,m_1,c_1,\epsilon, m_2, b_2,c_2):= 5 \left\{ \frac{16\kappa^{10}  c_Y^2}{\lambda_2^2}r_C(\delta,m_1,c_1)^2 \right.\\ 
    +\frac{16\kappa^{10}  c_Y^2}{\lambda_2^2}r_C(\delta,m_1,c_1)^2.4\left( \frac{32\ln^2(6/\epsilon)}{\lambda_2} \left[\frac{(c_Y + \| \eta_{AXW}\|_{\cH_{\A\X\W}})^2(4+m_2\lambda_2( \beta_2^{\frac{1}{b_2}} \frac{\pi/b_2}{sin(\pi/b_2)}\lambda_2^{-\frac{1}{b_2}})}{m_2^2\lambda_2}\right]+\right.\\
    \left.\frac{32 \ln^2(6/\epsilon)}{\lambda_2}\left[ \frac{4\gamma_2\lambda_2^{c_2-1}+m_2\gamma_2\lambda_2^{c_2}}{m_2^2\lambda_2}\right] +  \gamma_2\lambda_2^{c_2-1}+ \|\eta_{AXW}\|^2_{\cH_{\A\X\W}} \right) +\gamma_2\lambda_2^{c_2-1}    \left.\right.\\
    \left.+32 \ln^2(6/\epsilon) \left[\frac{(c_Y + \| \eta_{AXW}\|_{\cH_{\A\X\W}})^2(4+m_2\lambda_2( \beta_2^{\frac{1}{b_2}} \frac{\pi/b_2}{sin(\pi/b_2)}\lambda_2^{-\frac{1}{b_2}}))}{m_2^2\lambda_2^2}\right]+ 32 \ln^2(6/\epsilon)\left[ \frac{4\gamma_2\lambda_2^{c_2-1}+m_2\gamma_2\lambda_2^{c_2}}{m_2^2\lambda_2^2}\right]
    \right\}.
\end{multline*}
\end{corollary}
\begin{proof}
By \Cref{cor:rkh_norm_decomposition}, we have:
\begin{equation*}
   \|\widehat{\eta}_{AXW}-\eta_{AXW}\|^2_{\cH_{\A\X\W}}\le 5\left[ \tS_{-1} + \tS_0 + \mathcal{B}(\lambda_2) + \tS_1+\tS_2 \right]. 
\end{equation*}
Then, by \Cref{cor:bound_tS_minus1_0}, w.p. $1-\frac{\epsilon}{3}-\delta$, we have:
\begin{equation*}
    \tS_{-1}\le \frac{16}{\lambda_2^2} \kappa^{10} r_C(\delta,m_1,c_1)^2  c_Y^2,\quad 
 \tS_0 \le \frac{16}{\lambda_2^2}\kappa^{10}r_C(\delta,m_1,c_1)^2 \|\widetilde{\eta}_{AXW}\|^2_{\cH_{\A\X\W}}
\end{equation*}
where by \Cref{prop:bound_eta_tilde},
w.p. $1-\frac{2\epsilon}{3}$:
\begin{align*}
        &\|\widetilde{\eta}_{AXW}\|^2_{\cH_{\A\X\W}}\\
        &\le
        4\left( \frac{32\ln^2(6/\epsilon)}{\lambda_2} \left[\frac{(c_Y + \| \eta_{AXW}\|_{\cH_{\A\X\W}})^2(4+m_2\lambda_2\cN(\lambda_2))}{m_2^2\lambda_2}\right]+
    \frac{32 \ln^2(6/\epsilon)}{\lambda_2}\left[ \frac{4\cB(\lambda_2)+m_2\cA(\lambda_2)}{m_2\lambda_2}\right] +  \cB(\lambda_2)+ \|\eta_{AXW}\|^2_{\cH_{\A\X\W}} \right).
\end{align*}
Also, by \Cref{cor:bound_tS1_tS2}, w.p. $1-\frac{2\epsilon}{3}$, we have:
\begin{equation*}
    \tS_1\le 32 \ln^2(6/\epsilon) \left[\frac{(c_Y + \| \eta_{AXW}\|_{\cH_{\A\X\W}})^2(4+m_2\lambda_2\cN(\lambda_2))}{m_2^2\lambda_2^2}\right],\quad
    \tS_2\le 32 \ln^2(6/\epsilon)\left[ \frac{4\cB(\lambda_2)+m_2\cA(\lambda_2)}{m_2^2\lambda_2^2}\right].
\end{equation*}
Finally, by \Cref{prop:bound_operator_quantities},
\begin{equation*}
    \cA(\lambda_2)\le \gamma_2 \lambda_2^{c_2},\quad\cB(\lambda_2)\le \gamma_2 \lambda_2^{c_2-1}, \quad \cN(\lambda_2)\le \beta_2^{\frac{1}{b_2}} \frac{\pi/b_2}{sin(\pi/b_2)}\lambda_2^{-\frac{1}{b_2}}.
\end{equation*} 
Combining all the probabilistic bounds yields the final result.
\end{proof}

\begin{thm}\label{thm:kpv_final_thm_rkhs}
Suppose \Cref{ass:polish_spaces,ass:kernel_characteristic,ass:y_bounded,ass:well_posed1,ass:source1,ass:well_posed2,ass:source2} hold. 
Fix $\zeta>0$ and choose $\lambda_1=m_1^{\frac{1}{c_1+1}}$ and $m_1 =m_2^{\frac{\zeta(c_1+1)}{(c_1-1)}}$.
\begin{enumerate}
\setlength{\itemindent}{-0.1in}
    \item If $\zeta\le \frac{b_2(c_2+1)}{b_2 c_2+1}$, choose $\lambda_2={m_2}^{-\frac{\zeta}{c_2+1}}$. Then $\|\widehat{\eta}_{AXW}-\eta_{AXW}\|^2_{\cH_{\A\X\W}}=O_p\left(m_2^{-\frac{\zeta (c_2-1)}{c_2+1}}\right)$.
    \item If $\zeta \ge \frac{b_2(c_2+1)}{b_2 c_2+1}$, choose $\lambda_2={m_2}^{-\frac{b_2}{b_2 c_2+1}}$. Then $\|\widehat{\eta}_{AXW}-\eta_{AXW}\|^2_{\cH_{\A\X\W}}=O_p\left(m_2^{-\frac{b_2 (c_2-1)}{b_2 c_2+1}}\right)$.
\end{enumerate}
\end{thm}

\textbf{Proof of \Cref{thm:kpv_final_thm_rkhs}}.
\begin{proof}
Ignoring constants in \Cref{cor:kpv_consistency}, we have:
\begin{align*}
    \tS_{-1}&=O\left(\frac{r_C(\delta,m_1,c_1)^2}{\lambda_2^2}\right),\\
    &\tS_0=O\left(\frac{r_C(\delta,m_1,c_1)^2}{\lambda_2^2}.\left(\frac{1}{m_2^2\lambda_2^2}+\frac{1}{m_2\lambda_2^{1+1/b_2}}+\frac{1}{m_2^2\lambda_2^{3-c_2}}+\frac{1}{m_2\lambda_2^{2-c_2}}+\lambda_2^{c_2-1}+1\right)\right)\\
    &\mathcal{B}(\lambda_2)=O(\lambda_2^{c_2-1}),\quad \tS_1=O\left(\frac{1}{m_2^2\lambda_2^2}+\frac{1}{m_2\lambda_2^{1+1/b_2}}\right),\quad \tS_2=O\left(\frac{1}{m_2^2\lambda_2^{3-c_2}}+\frac{1}{m_2\lambda_2^{2-c_2}}\right).
\end{align*}
Following the same reasoning as in the proof of \Cref{thm:kpv_consistency}, this  leads to:
\begin{equation*}
    \|\widehat{\eta}_{AXW}-\eta_{AXW}\|^2_{\cH_{\A\X\W}}=O\left( \frac{1}{m_2^{2+\zeta}\lambda_2^4} +\frac{1}{m_2^{1+\zeta}\lambda_2^{3+1/b_2}}+\frac{1}{m_2^{\zeta}\lambda_2^2}
    +\lambda_2^{c_2-1}+\frac{1}{m_2^2\lambda_2^2}+\frac{1}{m_2\lambda_2^{1+1/b_2}}\right).
\end{equation*}
The final result results from \citet[Theorem 5]{szabo2016learning}. It consists in matching pairs of terms in the above equation and dividing by $\lambda_2$ to obtain the final rate.
\end{proof}

\subsubsection{\textbf{Proof details for \Cref{thm:kpv_consistency}}}


First introduce $\widetilde{\eta}_{AXW}$ as the minimizer of the empirical risk of stage 2, when plugging the true $\mu_{W|a,x,z}$ (instead of its estimate from Stage 1):
\begin{equation}\label{eq:erm2bis}
    \widetilde{\eta}_{AXW}=\argmin_{\eta\in \cH_{\A\X\W}} \widetilde{L}(\eta) , 
    \text{ where }\widetilde{L}(\eta)=\frac{1}{m_2} \sum_{j=1}^{m_2}\left( \ty_j- \eta[ \phia(\ta_j,\tx_j)\otimes \mu_{W|\ta_j,\tx_j,\tz_j} ] \right)^2 +\lambda_2 \Vert\eta\Vert^2_{\cH_{\A\X\W}}.
\end{equation}
Similarly to $\widehat{\eta}_{AXW}$, it has a closed form solution given below (see \citet[Section D.1]{grunewalder2012modelling}).
\begin{thm}\label{thm:closed_form2bis}
For any $\lambda_2>0$, the solutions of \eqref{eq:erm2bis}, exists, is unique, and is given by:
\begin{align*}
    \widetilde{\eta}_{AXW}&=(\boldsymbol{T}_2 +\lambda_2)^{-1} g_2, \text{ where }\boldsymbol{T}_2 = \frac{1}{m_2} \sum_{j=1}^{m_2} \left[ \mu_{W|\ta_j,\tx_j,\tz_j}\otimes \phi(\ta_j,\tx_j)\right]\otimes \left[ \mu_{W|\ta_j,\tx_j,\tz_j}\otimes \phi(\ta_j,\tx_j)\right]\\
&\text{and } g_2 =\frac{1}{m_2}\sum_{j=1}^{m_2} \left[ \mu_{W|\ta_j,\tx_j,\tz_j}\otimes \phi(\ta_j,\tx_j)\right]\ty_j.
\end{align*}
\end{thm}

Define also $\eta_{AXW}^{\lambda_2}$ as the minimizer of the population version of \eqref{eq:erm2bis}:
\begin{equation}\label{eq:pop2bis}
    \eta_{AXW}^{\lambda_2}=\argmin_{\eta \in \cH_{\A\X\W}} L^{\lambda_2}(\eta) ,\\ \text{ where }L^{\lambda_2}(\eta)= \E_{AXYZ}\left\{ Y- \eta[ \phia(A,X)\otimes \mu_{W|a,x,z} ] \right\}^2 +\lambda_2 \Vert\eta\Vert^2_{\cH_{\A\X\W}}.
\end{equation}

The excess risk for the KPV estimator can be decomposed in five terms as stated in the following proposition.

\begin{proposition}\label{prop:risk_decomposition}
The excess risk of the Stage 2 estimator can be bounded by five terms:
\begin{equation*}
    \tilde{R}(\widehat{\eta}_{AXW})-\tilde{R}(\eta_{AXW})\le 5\left[ S_{-1} + S_0 + \A(\lambda_2) + S_1+S_2 \right]
\end{equation*}
where 
\begin{align*}
    &S_{-1} = \|\sqrt{T_2} \circ (\boldsymbol{\widehat{T}_2} + \lambda_2)^{-1}(\widehat{g}_2 - g_2)\|^2_{\cH_{\A\X\W}}, \quad S_0 = \|\sqrt{T_2} \circ (\boldsymbol{\widehat{T}_2} + \lambda_2)^{-1}
\circ (\boldsymbol{T_2} - \boldsymbol{\widehat{T}_2})\widetilde{\eta}_{AXW}\|^2_{\cH_{\A\X\W}}\\
   & S_1 = \|\sqrt{T_2}\circ (\boldsymbol{T_2} +\lambda_2)^{-1}(g_2 -\boldsymbol{T}_2 \eta_{AXW})\|^2_{\cH_{\A\X\W}},\quad S_2 = \|\sqrt{T_2} \circ (\boldsymbol{T_2} + \lambda_2)^{-1}\circ (T_2-\boldsymbol{T_2})(\eta_{AXW}^{\lambda_2} - \eta_{AXW})\|^2_{\cH_{\A\X\W}}\\
    &\text{ and the residual } \A(\lambda_2)= \|\sqrt{T_2} (\eta^{\lambda_2}_{AXW} -\eta_{AXW})\|^2_{\cH_{\A\X\W}}.
\end{align*}
\end{proposition}
\begin{proof} The excess risk can be decomposed as:
\begin{align}\label{eq:risk_decomposition}
   \hspace{-0.3cm} \tilde{R}(\widehat{\eta}_{AXW})-\tilde{R}(\eta_{AXW})&=\| \sqrt{T_2}(\widehat{\eta}_{AXW} -\eta_{AXW})\|^2_{\cH_{\A\X\W}}\nonumber\\
   &=\| \sqrt{T_2}\left[(\widehat{\eta}_{AXW} - \widetilde{\eta}_{AXW}) + (\widetilde{\eta}_{AXW} -  \eta_{AXW}^{\lambda_2})+ (\eta_{AXW}^{\lambda_2} -\eta_{AXW})\right]\|^2_{\cH_{\A\X\W}} 
\end{align}
Using the operator identity $A^{-1}-B^{-1}= A^{-1}(B-A)B^{-1}$ and \Cref{thm:closed_form2}, the first term in \eqref{eq:risk_decomposition} can be bounded by $5(S_{-1}+S_0)$, the second one by $5(S_1+S_2)$ and the last one by $5\A(\lambda_2)$ (see \citet[Section A.1.8]{szabo2015two}). The factor 5 comes from the inequality $(\sum_{i=1}^{n}a_i)^2\le n \sum_{i=1}^n a_i^2$.
\end{proof}

We next give the analogous RKHS norm result. 

\begin{corollary}
\label{cor:rkh_norm_decomposition}
The error in RKHS norm of the Stage 2 estimator can be bounded by five terms:
\begin{equation*}
    \|\widehat{\eta}_{AXW}-\eta_{AXW}\|^2_{\cH_{\A\X\W}}\le 5\left[ \tS_{-1} + \tS_0 + \mathcal{B}(\lambda_2) + \tS_1+\tS_2 \right]
\end{equation*}
where 
\begin{align*}
    &\tS_{-1} = \|(\boldsymbol{\widehat{T}_2} + \lambda_2)^{-1}(\widehat{g}_2 - g_2)\|^2_{\cH_{\A\X\W}}, \quad \tS_0 = \|(\boldsymbol{\widehat{T}_2} + \lambda_2)^{-1}
\circ (\boldsymbol{T_2} - \boldsymbol{\widehat{T}_2})\widetilde{\eta}_{AXW}\|^2_{\cH_{\A\X\W}}\\
   & \tS_1 = \| (\boldsymbol{T_2} +\lambda_2)^{-1}(g_2 -\boldsymbol{T}_2 \eta_{AXW})\|^2_{\cH_{\A\X\W}},\quad \tS_2 = \| (\boldsymbol{T_2} + \lambda_2)^{-1}\circ (T_2-\boldsymbol{T_2})(\eta_{AXW}^{\lambda_2} - \eta_{AXW})\|^2_{\cH_{\A\X\W}}\\
    &\text{ and the residual } \mathcal{B}(\lambda_2)= \|\eta^{\lambda_2}_{AXW} -\eta_{AXW}\|^2_{\cH_{\A\X\W}}.
\end{align*}
\end{corollary}

The first two terms $S_{-1},S_0$  in \Cref{prop:risk_decomposition} (likewise $\tS_{-1},\tS_0$ in \Cref{cor:rkh_norm_decomposition}) characterize the estimation error due to Stage 1; the middle term $\cA(\lambda_2)$  (or $\cB(\lambda_2)$ in \Cref{cor:rkh_norm_decomposition}) characterizes the regularization bias; while the two last terms $S_1,S_2$ (or $\tS_1,\tS_2$ in \Cref{cor:rkh_norm_decomposition}) characterize the estimation error from Stage 2. 
The goal is now to bound each term of \Cref{prop:risk_decomposition}  (or  \Cref{cor:rkh_norm_decomposition}) separately. For the three last terms from Stage 2, we can benefit from the minimax rates and results for ridge regression \citep{caponnetto2007optimal}, see  \Cref{prop:bound_operator_quantities,prop:bound_S1_S2,cor:bound_tS1_tS2}. Stage 1 requires intermediate results (\Cref{prop:bound_g_T,prop:bound_T2_T2hat,prop:bound_S_minus1_0} for the proof of \Cref{thm:kpv_final_thm}, and \Cref{cor:boundOperatorNormInvRKHS,cor:bound_tS_minus1_0} for the proof of \Cref{prop:rate_kpv_ce}). 

We first have the following bounds that characterize the relation between $\eta^{\lambda_2}_{AXW}$ and $\eta_{AXW}$.

\begin{proposition}\label{prop:bound_operator_quantities}
Suppose \Cref{ass:source2} holds, which means that $\rho \in \cP(\gamma_2,b_2,c_2)$ and that the eigenvalues $(l_k)_{k\in \mathbb{N}^*}$ of $T_2$ satisfy $\alpha_2\le l_k  k^{-b_2}\le \beta_2$. Then,
 the residual $\cA(\lambda_2)$, the reconstruction error $\cB(\lambda_2)$, and the effective dimension $\cN(\lambda_2)$ are defined and bounded as follows:
 \begin{align*}
     \cA(\lambda_2) = \|\sqrt{T_2} (\eta^{\lambda_2}_{AXW} -\eta_{AXW})\|^2_{\cH_{\A\X\W}}\le \gamma_2 \lambda_2^{c_2},& \qquad
     \cB(\lambda_2)=\| \eta^{\lambda_2}_{AXW} -\eta_{AXW}\|^2_{\cH_{\A\X\W}}\le \gamma_2 \lambda_2^{c_2-1},\\ \cN(\lambda_2)=Tr\left[(T_2+\lambda_2)^{-1}\circ T_2 \right]&\le \beta_2^{\frac{1}{b_2}} \frac{\pi/b_2}{sin(\pi/b_2)}\lambda_2^{-\frac{1}{b_2}}.
 \end{align*}
\end{proposition}
The bounds on $\cA(\lambda_2)$, $\cB(\lambda_2)$ follow from \citet[Proposition 3]{caponnetto2007optimal}, while the bound on $\cN(\lambda_2)$ follows from \citet{sutherland2017fixing}. The residual $\cA(\lambda_2)$ and reconstruction error $\cB(\lambda_2)$, which depend on $\rho$, control the complexity of $\eta_{AXW}$. 
The effective dimension $\cN(\lambda_2)$ measures the complexity of the hypothesis space $\cH_{\A\X\W}$ with respect to $\rho_{\Hw\times\A\times\X}$.

\begin{proposition}\label{prop:bound_S1_S2}\citep[Step 2 and 3 of Theorem 4]{caponnetto2007optimal}
Assume  \Cref{ass:y_bounded} and \Cref{ass:well_posed2} hold. Assume also that $\lambda_2\le \| T_2\|_{\cL{\cH_{\A\X\W}}}$ and $m_2\ge \frac{2C_{\epsilon}\mathcal{N}(\lambda_2)}{\lambda_2}$. Then, we can bound $S_1$ and $S_2$ from \Cref{prop:risk_decomposition} as follows w.p. $1-2\epsilon/3$:
\begin{equation*}
    S_1\le 32 \ln^2(6/\epsilon) \left[\frac{(c_Y + \| \eta_{AXW}\|_{\cH_{\A\X\W}})^2(4+m_2\lambda_2\cN(\lambda_2))}{m_2^2\lambda_2}\right],\quad
    S_2\le 8 \ln^2(6/\epsilon)\left[ \frac{4\cB(\lambda_2)+m_2\cA(\lambda_2)}{m_2^2\lambda_2}\right].
\end{equation*}
\end{proposition}
\begin{corollary}\label{cor:bound_tS1_tS2}
Suppose the assumptions of \Cref{prop:bound_S1_S2} hold.  Then, we can bound $\tS_1$ and $\tS_2$ from \Cref{cor:rkh_norm_decomposition} as follows w.p. $1-2\epsilon/3$:
\begin{equation*}
    \tS_1\le 32 \ln^2(6/\epsilon) \left[\frac{(c_Y + \| \eta_{AXW}\|_{\cH_{\A\X\W}})^2(4+m_2\lambda_2\cN(\lambda_2))}{m_2^2\lambda_2^2}\right],\quad
    \tS_2\le 32 \ln^2(6/\epsilon)\left[ \frac{4\cB(\lambda_2)+m_2\cA(\lambda_2)}{m_2^2\lambda_2^2}\right].
\end{equation*}
\end{corollary}
\begin{proof}
Both resuls in \Cref{cor:bound_tS1_tS2} are  minor changes to the relevant proofs of \citet{caponnetto2007optimal}. Using the notation of the present paper: for  the term $\tS_1$, the left hand side of \citep[eq. 47]{caponnetto2007optimal} loses the leading $\sqrt{T_2}$, and the right hand bound becomes $2/\sqrt{\lambda_2}$. For the term $\tS_2$, the left hand side of \citep[eq. 39]{caponnetto2007optimal} loses the leading $\sqrt{T_2}$, and the right hand becomes $2/\lambda_2$. The reasoning is the same as in our bounds for $\tS_{-1}$ and $\tS_{0}$: see in particular \Cref{cor:boundOperatorNormInvRKHS}. 
\end{proof}
The following bounds are obtained easily by using the bounds from \Cref{thm:kpv_s1_bound} on the difference between the estimated conditional mean embeddings of Stage 1 and the true one.

\begin{proposition} \label{prop:bound_g_T}
Assume the assumptions of \Cref{thm:kpv_s1_bound} hold and define $\lambda_1$ accordingly. Suppose also that \Cref{ass:y_bounded,ass:well_posed2} hold. Then, w.p. \(\,1 - \delta\,\):
\begin{equation}
    \Vert \widehat{g_2} - g_2 \Vert^2_{\cH_{\A\X\W}} \leq
     \kappa^{10} r_C(\delta,m_1,c_1)^2  c_Y^2,\quad\text{ and }\quad
    \Vert \boldsymbol{T}_2- \boldsymbol{\widehat{T}}_2 \Vert^2_{\cL(\cH_{\A\X\W})} \le 4\kappa^{10}r_C(\delta,m_1,c_1)^2.
\end{equation}
\end{proposition}
\begin{proof}
Using \Cref{ass:y_bounded}, \Cref{thm:kpv_s1_bound}, and  $(\sum_{i=1}^{n}a_i)^2\le n \sum_{i=1}^n a_i^2$, we have :
\begin{multline*}
 \Vert \widehat{g_2} - g_2 \Vert^2_{\cH_{\A\X\W}} \le \frac{1}{m_2}\sum_{j=1}^{m_2} \|\left[ (\widehat{\mu}_{W|\ta_j,\tx_j,\tz_j}-\mu_{W|\ta_j,\tx_j,\tz_j})\otimes \phi(\ta_j,\tx_j)\right]\ty_j\|^2_{\cH_{\A\X\W}}\\ \le\frac{1}{m_2}\sum_{j=1}^{m_2} \| (\widehat{\mu}_{W|\ta_j,\tx_j,\tz_j}-\mu_{W|\ta_j,\tx_j,\tz_j})\|^2_{\Hw} \|\phi(\ta_j,\tx_j)\|^2_{\cH_{\A\X}}\ty_j^2 \le \kappa^{10} r_C(\delta,m_1,c_1)^2  c_Y^2.
\end{multline*}
On the other hand,  using $(\sum_{i=1}^{n}a_i)^2\le n \sum_{i=1}^n a_i^2$ and the identity $(a+b)^2\le 2a^2+2b^2$, we have:
\begin{align*}
  &\Vert \boldsymbol{T}_2- \boldsymbol{\widehat{T}}_2 \Vert^2_{\cL(\cH_{\A\X\W})}\\ 
  &\le \frac{2}{m_2}\sum_{j=1}^{m_2} \| \left[ (\mu_{W|\ta_j,\tx_j,\tz_j}-\widehat{\mu}_{W|\ta_j,\tx_j,\tz_j})\otimes \phi(\ta_j,\tx_j)\right]\otimes \left[ \mu_{W|\ta_j,\tx_j,\tz_j}\otimes \phi(\ta_j,\tx_j)\right]\|^2_{\cL(\cH_{\A\X\W})}\\
  &+\frac{2}{m_2}\sum_{j=1}^{m_2}\| \left[\widehat{\mu}_{W|\ta_j,\tx_j,\tz_j}\otimes \phi(\ta_j,\tx_j)\right]\otimes \left[  (\mu_{W|\ta_j,\tx_j,\tz_j}-\widehat{\mu}_{W|\ta_j,\tx_j,\tz_j})\otimes \phi(\ta_j,\tx_j)\right]\|^2_{\cL(\cH_{\A\X\W})}\\
    &\le \frac{2}{m_2} \sum_{j=1}^{m_2}\| \widehat{\mu}_{W|\ta_j,\tx_j,\tz_j}-\mu_{W|\ta_j,\tx_j,\tz_j}\|^2_{\Hw}\|\mu_{W|\ta_j,\tx_j,\tz_j}\|^2_{\Hw} \|\phi(\ta_j,\tx_j)\|^2_{\cH_{\A\X}}\\
    &+ \frac{2}{m_2} \sum_{j=1}^{m_2}\| \widehat{\mu}_{W|\ta_j,\tx_j,\tz_j}-\mu_{W|\ta_j,\tx_j,\tz_j}\|^2_{\Hw}\|\widehat{\mu}_{W|\ta_j,\tx_j,\tz_j}\|^2_{\Hw} \|\phi(\ta_j,\tx_j)\|^2_{\cH_{\A\X}}
    \\
    &\le 4 \kappa^{10}r_C(\delta,m_1,c_1)^2. \qedhere
\end{align*}
\end{proof}

\begin{proposition}\label{prop:bound_T2_T2hat}
Assume the assumptions of \Cref{thm:kpv_s1_bound} hold and define $\lambda_1$ accordingly. Let $C_{\epsilon}=96\ln^2(6/\epsilon)$. Suppose also that \Cref{ass:y_bounded,ass:well_posed2} hold. Finally, assume $\lambda_2\le \|T_2\|_{\cL(\cH_{\A\X\W})}$ and that :
\begin{equation*}
    m_2\ge  \frac{2C_{\epsilon}\mathcal{N}(\lambda_2)}{\lambda_2},\qquad m_1\ge \bar{m}(\delta, c_1)) , 
    :=\left[ \frac{8\kappa\sqrt{\gamma_1}(c_1+1)}{4^{\frac{1}{c_1+1}}\lambda_2} \right]^{2\frac{c_1+1}{c_1-1}} \left(\frac{4\kappa^3(\kappa+\kappa^3\|C_{W|A,X,Z} \|_{\cH_{\Gamma}})\ln(2/\delta)}{\sqrt{\gamma_1}(c_1-1)}\right)^{2}.
\end{equation*}
Then, w.p. $1-\frac{\epsilon}{3}-\delta$, we have:
\begin{equation*}
\|\sqrt{T_2} \circ (\boldsymbol{\widehat{T}_2} + \lambda_2)^{-1}\|_{\cL(\cH_{\A\X\W})}\le \frac{2}{\sqrt{\lambda_2}}.
\end{equation*}
\end{proposition}

\begin{proof}
We follow the proof of \citet[Proposition 39]{singh2019kernel}. Using the Neumann series of $I-(T_2-\boldsymbol{\widehat{T}_2})(T_2+\lambda_2)^{-1}$, we have:
\begin{equation*}
    \|\sqrt{T_2} \circ (\boldsymbol{\widehat{T}_2} + \lambda_2)^{-1}\|_{\cL(\cH_{\A\X\W})}\le \|\sqrt{T_2} \circ (T_2 + \lambda_2)^{-1}\|_{\cL(\cH_{\A\X\W})} \sum_{k=0}^{\infty}\|(T_2 -\boldsymbol{\widehat{T}_2}) \circ (T_2 + \lambda_2)^{-1}\|^k_{\cL(\cH_{\A\X\W})}.
\end{equation*}
We first deal with the first term on the r.h.s. Observe that by definition of the operator norm,
\begin{equation*}
    \|\sqrt{T_2} \circ (T_2 + \lambda_2)^{-1}\|_{\cL(\cH_{\A\X\W})}=\sup_{l \in (l_k)_{k\in \mathbb{N}^*}}\frac{\sqrt{l}}{l+\lambda_2}\le \frac{1}{2\sqrt{\lambda_2}},
\end{equation*}
where the last inequality results from arithmetic-geometric mean inequality ($\sqrt{l\lambda_2}\le (l+\lambda_2)/2$). We now deal with the second term on the r.h.s. First, we apply the triangle inequality :
\begin{equation*}
    \|(T_2 -\boldsymbol{\widehat{T}_2}) \circ (T_2 + \lambda_2)^{-1}\|_{\cL(\cH_{\A\X\W})}\le \|(T_2 -\boldsymbol{T_2}) \circ (T_2 + \lambda_2)^{-1}\|_{\cL(\cH_{\A\X\W})}+ \|(\boldsymbol{T}_2 -\boldsymbol{\widehat{T}_2}) \circ (T_2 + \lambda_2)^{-1}\|_{\cL(\cH_{\A\X\W})}.
\end{equation*}
Since $\| (T_2 +\lambda_2)^{-1}\|_{\cL(\cH_{\A\X\W})}\le 1/\lambda_2$, by \Cref{prop:bound_T2_T2hat} the second term is easily bounded w.p. $1-\delta$ as : \begin{equation*}
    \|(\boldsymbol{T}_2 -\boldsymbol{\widehat{T}_2}) \circ (T_2 + \lambda_2)^{-1}\|_{\cL(\cH_{\A\X\W})}\le\frac{ \|\boldsymbol{T}_2 -\boldsymbol{\widehat{T}_2}\|_{\cL(\cH_{\A\X\W})}}{\lambda_2}\le \frac{\kappa^{5}r_C(\delta,m_1,c_1)}{\lambda_2}.
\end{equation*}
For a  $\lambda_2$, $m_1$ can be chosen so that $\kappa^{5}r_C(\delta,m_1,c_1)/\lambda_2\le 1/4$, which legitimates the use of the Neumann series at the beginning of the proof. This is actually given by setting $m_1\ge \bar{m}(\delta,c_1)$.
By \citet[Step 2.1, Theorem 4]{caponnetto2007optimal}, the first term is bounded with probability $1-\frac{\epsilon}{3}$ by:
\begin{equation*}
    \|(T_2 -\boldsymbol{T_2}) \circ (T_2 + \lambda_2)^{-1}\|_{\cL(\cH_{\A\X\W})}\le \frac{1}{2}.
\end{equation*}
for $m_2\ge\frac{2C_{\epsilon}\mathcal{N}(\lambda_2)}{\lambda_2}$. 
Hence, we can conclude that for $m_1\ge \bar{m}(\delta,c_1)$ and $m_2\ge \frac{2C_{\epsilon}\mathcal{N}(\lambda_2)}{\lambda_2}$, we have w.p. $1-\frac{\epsilon}{3}-\delta$:
\begin{equation*}
    \|(T_2 -\boldsymbol{\widehat{T}_2}) \circ (T_2 + \lambda_2)^{-1}\|_{\cL(\cH_{\A\X\W})}\le \frac{1}{2}+\frac{1}{4}=\frac{3}{4}\Longrightarrow \|\sqrt{T_2} \circ (\boldsymbol{\widehat{T}_2} + \lambda_2)^{-1}\|_{\cL(\cH_{\A\X\W})}\le \frac{1}{2\sqrt{\lambda_2}}\frac{1}{1-\frac{3}{4}}=\frac{2}{\sqrt{\lambda_2}}.\qedhere
\end{equation*}
\end{proof}

\begin{corollary}\label{cor:boundOperatorNormInvRKHS}
Suppose the assumptions of \Cref{prop:bound_T2_T2hat} hold. Then, w.p. $1-\frac{\epsilon}{3}-\delta$, we have:
\begin{equation*}
\|(\boldsymbol{\widehat{T}_2} + \lambda_2)^{-1}\|_{\cL(\cH_{\A\X\W})}\le \frac{4}{\lambda_2}.
\end{equation*}
\end{corollary}

\begin{proof}
     Using the Neumann series of $I-(T_2-\boldsymbol{\widehat{T}_2})(T_2+\lambda_2)^{-1}$, we have:
\begin{equation*}
    \| (\boldsymbol{\widehat{T}_2} + \lambda_2)^{-1}\|_{\cL(\cH_{\A\X\W})}\le \| (T_2 + \lambda_2)^{-1}\|_{\cL(\cH_{\A\X\W})} \sum_{k=0}^{\infty}\|(T_2 -\boldsymbol{\widehat{T}_2}) \circ (T_2 + \lambda_2)^{-1}\|^k_{\cL(\cH_{\A\X\W})}.
\end{equation*}
We first deal with the first term on the r.h.s. Observe that by definition of the operator norm,
\begin{equation*}
    \|(T_2 + \lambda_2)^{-1}\|_{\cL(\cH_{\A\X\W})}=\sup_{l \in (l_k)_{k\in \mathbb{N}^*}}\frac{1}{l+\lambda_2}\le \frac{1}{\lambda_2}.
\end{equation*}
The second term on the r.h.s. is bounded as in the proof of \Cref{prop:bound_T2_T2hat}. Hence, we can conclude that for $m_1\ge \bar{m}(\delta,c_1)$ and $m_2\ge \frac{2C_{\epsilon}\mathcal{N}(\lambda_2)}{\lambda_2}$, we have w.p. $1-\frac{\epsilon}{3}-\delta$:
\begin{equation*}
    \|(T_2 -\boldsymbol{\widehat{T}_2}) \circ (T_2 + \lambda_2)^{-1}\|_{\cL(\cH_{\A\X\W})}\le \frac{1}{2}+\frac{1}{4}=\frac{3}{4}\Longrightarrow \| (\boldsymbol{\widehat{T}_2} + \lambda_2)^{-1}\|_{\cL(\cH_{\A\X\W})}\le \frac{1}{\lambda_2}\frac{1}{1-\frac{3}{4}}=\frac{4}{\lambda_2}.\qedhere
\end{equation*}
\end{proof}

We now bound each term separately. 

\begin{proposition}\label{prop:bound_S_minus1_0} Assume the conditions of \Cref{prop:bound_g_T,prop:bound_T2_T2hat} hold. We can bound $S_{-1}$ and $S_0$ from \Cref{prop:risk_decomposition} w.p. $1-\frac{\epsilon}{3}-\delta$ as follows:
\begin{equation*}
 S_{-1}\le \frac{4}{\lambda_2} \kappa^{10} r_C(\delta,m_1,c_1)^2  c_Y^2,\quad 
 S_0 \le \frac{4}{\lambda_2}\kappa^{10}r_C(\delta,m_1,c_1)^2 \|\widetilde{\eta}_{AXW}\|^2_{\cH_{\A\X\W}}.
\end{equation*}
\end{proposition}
\begin{proof}
Using \Cref{prop:bound_g_T} and \Cref{prop:bound_T2_T2hat}, we have:
\begin{equation*}
    S_{-1}\le  \|\sqrt{T_2} \circ (\boldsymbol{\widehat{T_2}} + \lambda_2)^{-1}\|^2_{\cL(\cH_{\A\X\W})}\|\widehat{g}_2 - g_2\|^2_{\cH_{\A\X\W}}\le \frac{4}{\lambda_2} \kappa^{10} r_C(\delta,m_1,c_1)^2  c_Y^2,
\end{equation*}
and similarly we have:
\begin{equation*}
    S_{0}\le  \|\sqrt{T_2} \circ (\boldsymbol{\widehat{T}_2} + \lambda_2)^{-1}\|^2_{\cL(\cH_{\A\X\W})}
\|\boldsymbol{T_2} - \boldsymbol{\widehat{T}_2}\|^2_{\cL(\cH_{\A\X\W})} \|\widetilde{\eta}_{AXW}\|^2_{\cH_{\A\X\W}}\le \frac{4}{\lambda_2}\kappa^{10}r_C(\delta,m_1,c_1)^2 \|\widetilde{\eta}_{AXW}\|^2_{\cH_{\A\X\W}}.\qedhere
\end{equation*}
\end{proof}

\begin{corollary}
\label{cor:bound_tS_minus1_0} Assume the conditions of \Cref{prop:bound_g_T,prop:bound_T2_T2hat} hold. We can bound $\tS_{-1}$ and $\tS_0$ from \Cref{cor:rkh_norm_decomposition} w.p. $1-\frac{\epsilon}{3}-\delta$ by applying \Cref{cor:boundOperatorNormInvRKHS}, which yields
\begin{equation*}
 \tS_{-1}\le \frac{16}{\lambda_2^2} \kappa^{10} r_C(\delta,m_1,c_1)^2  c_Y^2,\quad 
 \tS_0 \le \frac{16}{\lambda_2^2}\kappa^{10}r_C(\delta,m_1,c_1)^2 \|\widetilde{\eta}_{AXW}\|^2_{\cH_{\A\X\W}}.
\end{equation*}
\end{corollary}

\begin{proposition}\label{prop:bound_eta_tilde}
Let $C_{\epsilon}=96\ln^2(6/\epsilon)$ and suppose that $m_2\ge \frac{2C_{\epsilon}\cN(\lambda_2)}{\lambda_2}$ and that $\lambda_2\le \|T_2\|_{\cL(\cH_{\A\X\W})}$. Then, w.p. $1-2\epsilon/3$ 
\begin{align*}
    &\|\widetilde{\eta}_{AXW}\|^2_{\cH_{\A\X\W}}
    \\
    &\le 4\left( \frac{32\ln^2(6/\epsilon)}{\lambda_2} \left[\frac{(c_Y + \| \eta_{AXW}\|_{\cH_{\A\X\W}})^2(4+m_2\lambda_2\cN(\lambda_2))}{m_2^2\lambda_2}\right]+
    \frac{32 \ln^2(6/\epsilon)}{\lambda_2}\left[ \frac{4\cB(\lambda_2)+m_2\cA(\lambda_2)}{m_2^2\lambda_2}\right] +  \cB(\lambda_2)+ \|\eta_{AXW}\|^2_{\cH_{\A\X\W}} \right).
\end{align*}
\end{proposition}
\begin{proof}
Using the triangle inequality, we have:
\begin{align*}
     \|\widetilde{\eta}_{AXW}\|_{\cH_{\A\X\W}}&\le   \|\widetilde{\eta}_{AXW} - \eta_{AXW}^{\lambda_2}\|_{\cH_{\A\X\W}}+  \|\eta_{AXW}^{\lambda_2}-\eta _{AXW}\|_{\cH_{\A\X\W}}+ \|\eta_{AXW}\|_{\cH_{\A\X\W}}\\
     &=  \|\widetilde{\eta}_{AXW} - \eta_{AXW}^{\lambda_2}\|_{\cH_{\A\X\W}}+\sqrt{\cB(\lambda_2)}+ \|\eta_{AXW}\|_{\cH_{\A\X\W}}\\
     &\le \sqrt{\tS_1}+\sqrt{\tS_2}+\sqrt{\cB(\lambda_2)}+ \|\eta_{AXW}\|_{\cH_{\A\X\W}}
\end{align*}
via \Cref{cor:bound_tS1_tS2}. The final step applies $(\sum_{i=1}^{n}a_i)^2\le n \sum_{i=1}^n a_i^2$.
\end{proof}

\subsection{Proof of \Cref{prop:rate_kpv_ce}}\label{sec:proof_rate_kpv_ce}

We prove the pointwise covergence result for the average causal effect $\beta(a)$. Note that the proof appearing in the ICML 2021 proceedings used an incorrect norm (the $L_2$ norm results from \Cref{thm:kpv_consistency}, rather than the RKHS norm results from \Cref{thm:kpv_final_thm_rkhs}). Consequently there were factors $\lambda_2^{-1}$ missing from some terms, and the convergence rate reported was  faster than the correct rate. The present proof adds the additional $\lambda_2^{-1}$ factors and corrects the error in the rates. 

Let $\widehat{\mu}_{XW}=\frac{1}{n_t}\sum_{i=1}^{n_t}[\phix(x_i)\otimes\phi(w_i)]$ and $\mu_{XW}=\E_{XW}[\phix(X)\otimes\phiw(W)]$.
By \citet[Proposition 1]{Tolstikhin17}, we have w.p. $1-\delta$:
\begin{equation*}
    \|\widehat{\mu}_{XW}-\mu_{XW}\|_{\cH_{\X\W}}\le r_{\mu}(n_t,\delta)\le \frac{4\kappa^2\ln(2/\delta)}{n_t}:=r_{\mu}(n_t,\delta).
\end{equation*} 
Moreover, by  \Cref{cor:kpv_consistency}, we have w.p. $1-\epsilon-\delta$ 
\begin{equation*}
    \|\widehat{\eta}_{AXW}-\eta_{AXW}\|_{\cH_{\A\X\W}}\le \tilde{r}_H(\delta,m_1,c_1,\epsilon, m_2, b_2,c_2).
\end{equation*}
We use the following decomposition for the causal effect :
\begin{align*}
   \widehat{\beta}(a)-\beta(a)&=   \widehat{\eta}_{AXW}[\widehat{\mu}_{XW}\otimes \phia(a)] - \eta_{AXW}[\mu_{XW}\otimes\phia(a)]\\
   &=   \widehat{\eta}_{AXW}[(\widehat{\mu}_{XW}-\mu_{XW}\otimes \phia(a)]+( \widehat{\eta}_{AXW}-\eta_{AXW})[\mu_{XW}\otimes\phia(a)]\\
   &= (\widehat{\eta}_{AXW}-\eta_{AXW})[(\widehat{\mu}_{XW}-\mu_{XW})\otimes \phia(a)]
   +\eta_{AXW}[(\widehat{\mu}_{XW}-\E_{XW}[\phix(X)\otimes\phiw(W)])\otimes \phia(a)]\\
   &+( \widehat{\eta}_{AXW}-\eta_{AXW})[\mu_{XW}\otimes\phia(a)].
\end{align*}
Therefore, w.p. $1-\epsilon-\delta$, by \Cref{thm:kpv_final_thm_rkhs}, $\|\widehat{\eta}_{AXW}-\eta_{AXW}\|_{\cH_{\A\X\W}}=0(m^{-\alpha})$ with $\alpha \in \left\{\frac{\zeta (c_2-1)}{c_2+1},\frac{b_2(c_2-1)}{b_2c_2+1}\right\}$ and :
\begin{align*}
    |\widehat{\beta}(a)-\beta(a)|&\le \| \widehat{\eta}_{AXW}-\eta_{AXW}\|_{\cH_{\A\X\W}}\|\widehat{\mu}_{XW}-\mu_{XW}\|_{\cH_{\X\W}}\|\phia(a)\|_{\Ha}
    +\| \eta_{AXW}\|_{\cH_{\A\X\W}}\|\widehat{\mu}_{XW}-\mu_{XW}\|_{\cH_{\X\W}} \|\phia(a)\|_{\Ha}\\
    &+\|\widehat{\eta}_{AXW}-\eta_{AXW}\|_{\cH_{\A\X\W}}\|\mu_{XW}\otimes\phia(a)\|_{\cH_{\A\X\W}}\\
    &\le \kappa. \tilde{r}_H(\delta,m_1,c_1,\epsilon, m_2, b_2,c_2) r_{\mu}(n_t,\delta) + \kappa \|\eta_{AXW}\|_{\cH_{\A\X\W}}r_{\mu}(n_t,\delta) + \kappa^3 \tilde{r}_H(\delta,m_1,c_1,\epsilon, m_2, b_2,c_2)\\
    &=O(n_t^{-\frac{1}{2}}+m^{-\alpha}).
\end{align*}

\section{Proxy Maximum Moment Restriction}
\label{app:C-PMMR}

In this section, we propose a novel approach to solve the proximal causal learning using the maximum moment restriction (MMR) framework \citep{Muandet20:KCM}. 
It is based on that proposed by \citet{zhang2020maximum} for the IV setting. 
On the other hand, we adapt it to the proxy setting, with a novel interpretation for $h$.
This is inspired by \citet{miao2018identifying} and \citet{tchetgen2020introduction}, but in their formulations $h$ is defined to be the solution of an ill-posed inverse problem, whereas we view $h$ as a regression function for $y$, which is more interpretable, as we will detail below.

\subsection{Maximum Moment Restriction for Proxy Setting}
\label{sec:pmmr_approach}

\textbf{Notations.} (i) Let $\mathcal{X}$ denote a measurable space. (ii) Let $X$ denote a random variable taking values in $\mathcal{X}$.

Notice that $\{Y, A, X, W\}$ are random variables under the generating process governed by Figure \ref{fig:overall}. 
Let $h \in \Omega(\mathcal{A} \times \mathcal{W} \times \mathcal{X})$ be a measurable function on $\mathcal{A} \times \mathcal{W} \times \mathcal{X}$. Therefore, $h(A, W, X)$ is a function of random variables, which is a random variable itself. 

\paragraph{Proof of Lemma \ref{lem:proxy-mmr}.} The proposed method is based on \Cref{lem:proxy-mmr}, which shows that any function $h\in \Omega(\mathcal{A} \times \mathcal{W} \times \mathcal{X})$ that is the solution to \Cref{eq:integral-eq} must also satisfy the conditional moment restriction (CMR), and vice versa. 
\begin{proof}
Let $\varepsilon$ be a random variable representing the residual of $h(A, W, X)$ with respect to $Y$:
\begin{equation}\label{eq:residual}
    \varepsilon := Y - h(A,W,X).
\end{equation}
Suppose that $h$ is the solution to \eqref{eq:integral-eq}.
Then, taking the conditional expectation of \eqref{eq:residual} conditioned on $A, Z, X$ yields
\begin{align*}
    \mathbb{E}[\varepsilon|A, Z, X] &= \mathbb{E}[Y|A, Z, X] - \mathbb{E}[h(A, W, X)|A, X, Z] \\
    &= \mathbb{E}[Y|A, Z, X] - \int_{\mathcal{W}} h(A, w, X) f(w|A, Z, X) dw \\
    &= 0.
\end{align*}
In the last term of the second equality, we take expectation over $W$ because it is the only variable not being held fixed by conditioning.
The last equality holds because $h$ is the solution to \eqref{eq:integral-eq} by definition.
\end{proof}

Note that we have derived the condition typical in additive noise instrumental variable (IV) models \citep{Hartford17:DIV, Dikkala20:Minimax, Bennett19:DeepGMM, zhang2020maximum, Muandet19:DualIV}. 
A more general term for this type of conditions is called \textit{conditional moment restrictions} (CMR) \citep{Newey199316EE}. This interpretation allows us to approach the problem of learning $h$ from a different perspective. That is to say, we look for $h$ for which the conditional moment restriction is zero. This contrasts with the typical two-stage approach of learning $h$, for which the objective is to find $h$ such that $\mathbb{E}_{\mathit{AXZY}}[(Y-\mathbb{E}_{W}[h(A,W,X)|A,Z,X])^2]$ is minimized. 



\textbf{Connection to IV.}
Typical formulation of IV models assumes the following structural model:
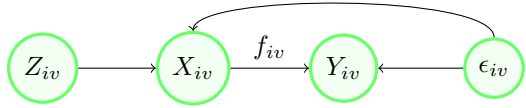
\begin{figure}[h!]
    \centering
    \begin{tikzpicture}[roundnode/.style={circle, draw=green!60, fill=green!5, very thick, minimum size=7mm},
squarednode/.style={rectangle, draw=red!60, fill=red!5, very thick, minimum size=5mm},]
    \node[roundnode] (x) at (0,0) {$X_{iv}$};
    \node[roundnode] (y) at (2,0) {$Y_{iv}$};
    \node[roundnode] (z) at (-2,0) {$Z_{iv}$};
    \node[roundnode] (eps) at (4,0) {$\epsilon_{iv}$};
    
    \draw[->] (x) -- (y) node[above, midway] {$f_{iv}$};
    \draw[->] (z) -- (x);
    \draw[->] (eps) -- (y);
    \draw[->] (eps) .. controls +(up:10mm) and +(up:10mm).. (x);
    \end{tikzpicture}

    \caption{DAG of an instrumental variable model}
    \label{fig:iv_graph}
\end{figure}

where in particular $Z_{iv} \indep \epsilon_{iv}$. Additionally, an additive noise model for generating $Y_{iv}$ is typically assumed, and the noise is assumed to have zero mean. In mathematical terms, these amount to
\begin{equation*}
    Y_{iv} = f_{iv}(X_{iv}) + \epsilon_{iv}, \hspace{0.5cm}\mathbb{E}[\epsilon_{iv}] = 0, \hspace{0.5cm}\mathbb{E}[\epsilon_{iv}|Z_{iv}] = 0 \hspace{0.25cm} a.s.
\end{equation*}

\begin{remark}
We make two comparisons between the IV setting and our proxy setting. 
\begin{enumerate}
    \item In the IV setting, $Y_{iv}=f_{iv}(X_{iv}) + \epsilon_{iv}$ is proposed as the \textit{structural} equation for $Y_{iv}$; in contrast, in our proxy setting $Y=h(A,X,W) + \epsilon$ is \textit{not} a structural equation, as it does not remain invariant under interventions. For readers unfamiliar with the concept of structural equations, we refer to \citet{pearl2009causality}. We thus offer the interpretation that $Y = h(A, W, X) + \epsilon$ is a \textit{regression} equation whose noise term has mean zero when conditioning on $A, Z, X$.
    \item In the IV setting, authors make the assumption of additive noise models for the structural equation of $Y_{iv}$, and $f_{iv}$ is then the causal effect of $X_{iv}$ on $Y_{iv}$, i.e. $f_{iv}(X_{iv}) = \mathbb{E}[Y_{iv}|do(X_{iv})]$. In our setting, the solution $h$ of the integral equation (\ref{eq:integral-eq}) directly gives us the causal effect, hence no additive noise assumption is made and our approach is entirely nonparametric.
    \item Lemma \ref{lem:proxy-mmr} establishes the connection between a class of problems that can be formulated in terms of an integral equation like \eqref{eq:integral-eq} and those that satisfy the CMR. Hence, we believe this result can be applied more broadly to problems that share similar structure to our setting.
\end{enumerate}
\end{remark}

\textbf{Proof of Lemma \ref{lem:pmmr_closed_form}}

\begin{proof}
Since $g \in \mathcal{H}_{\mathcal{AZX}}$, we may write $g(A,Z,X) = \langle g, k((A,Z,X), \cdot) \rangle$, thus we have
\begin{eqnarray*}
R_k(h) &=& \sup_{g \in \mathcal{H}_{\mathcal{AZX}}, \|g\|\leq 1} (\mathbb{E}[(Y-h(A,W,X))\langle g, k((A,Z,X), \cdot) \rangle])^2\\
&=& \sup_{g \in \mathcal{H}_{\mathcal{AZX}}, \|g\|\leq 1} (\mathbb{E}[\langle g,(Y-h(A,W,X)) k((A,Z,X), \cdot) \rangle])^2\\
&=& \sup_{g \in \mathcal{H}_{\mathcal{AZX}}, \|g\|\leq 1} (\langle g, \mathbb{E}[(Y-h(A,W,X)) k((A,Z,X), \cdot)] \rangle)^2\\
&=& \left\|\mathbb{E}[(Y-h(A,W,X))k((A,Z,X), \cdot)] \right\|^2_{\mathcal{H}_{\mathcal{AZX}}}.
\end{eqnarray*}
The second equality is due to linearity of an inner product, and we remark that it still holds despite $h$ and $g$ sharing variables $A$ and $X$, because $(Y- h(A,W,X)) \in \mathbb{R}$ as opposed to $\mathcal{H}_{\mathcal{AWX}}$. 
The last equality is due to the fact that $\mathcal{H}_{\mathcal{AZX}}$ is a vector space, and $\mathbb{E}[(Y-h(A,W,X))k((A,Z,X), \cdot)] \in \mathcal{H}_{\mathcal{AZX}}$ by assumption. 

Then, 
\begin{eqnarray*}
R_k(h) &=& \langle\mathbb{E}[(Y-h(A,W,X)) k((A,Z,X), \cdot), \mathbb{E}[(Y-h(A,W,X)) k((A,Z,X), \cdot) \rangle\\
&=& \mathbb{E}[\langle (Y-h(A,W,X)) k((A,Z,X), \cdot),(Y'-h(A',W',X')) k((A',Z',X'), \cdot) \rangle]\\
&=& \mathbb{E}[(Y-h(A,W,X))(Y'-h(A',W',X') k((A,Z,X),(A',Z',X') ))],
\end{eqnarray*}
as required.
\end{proof}


\subsection{Analytical Solution for PMMR}


Suppose further that $h$ also lies in an RKHS $\mathcal{H}_{\mathcal{A}\mathcal{Z}\mathcal{X}}$ endowed with the kernel function $l$. 
Then, we can use the representer theorem \citep{Scholkopf01:Representer} to derive a close-form solution for $h$. 
We note that the risk functional $R_k$ is different from standard least squares risk since it involves independent data samples as well as the kernel function $k$.
Nevertheless, the empirical risk still applies to data samples $\{y_i,a_i,w_i,x_i,z_i\}_{i=1}^n$, so the representer theorem still apply on RKHS features $\{l((a_i,w_i,x_i),\cdot)\}_{i=1}^n$.
This is to say, that by the representer theorem, 
\begin{equation*}
    \hat{h}(a,w,x) = \sum_{i=1}^n \alpha_i l((a_i, w_i, x_i), (a,w,x)),
\end{equation*}
for some $(\alpha_1,\ldots,\alpha_n) \in \mathbb{R}^n$.
Hence, we may rewrite the optimization problem as
\begin{equation}\label{eq:dual-mmr}
    \hat{\bm{\alpha}} = \argmin_{\bm{\alpha} \in \mathbb{R}^n} \; (\mathbf{y} - L \bm{\alpha} )^\top W (\mathbf{y} - L \bm{\alpha}) + \lambda \bm{\alpha}^\top L \bm{\alpha}
\end{equation}
where $L_{ij} = l((a_i, w_i, x_i), (a_j,w_j,x_j))$ and $W_{ij} = k\left((a_i,z_{i},x_i), (a_j,z_{j},x_j)\right)$.
The solution to \eqref{eq:dual-mmr} can be found by solving the first-order stationary condition, resulting in the closed-form expression:
\begin{equation*}
    \hat{\bm{\alpha}} = (LWL + \lambda L)^{-1}L W \mathbf{y}.
\end{equation*}

It can be shown that $\mathcal{H}_{\mathcal{X}_1 \times \cdots \times \mathcal{X}_m}$ is isometrically isomorphic to $\mathcal{H}_{\mathcal{X}_1} \otimes \cdots \otimes \mathcal{H}_{\mathcal{X}_m}$. 
In the latter, the kernel of the outer-product RKHS can be decomposed into the product of the kernels of the children RKHSes:
\begin{equation*}
    k(\mathbf{x}, \mathbf{x'}) = k_1(x_1,x'_1)k_2(x_2,x'_2)\cdots k_m(x_m, x'_m).
\end{equation*}
Hence, we may use an alternative closed-form formulation of $h$ with the product kernels
\begin{equation*}
    \hat{h}(a,w,x) = \sum_{i=1}^n \hat{\bm{\alpha}}_i l_{\mathcal{A}}(a_i,a)l_{\mathcal{W}}(w_i,w)l_{\mathcal{X}}(x_i,x)
\end{equation*}

\subsubsection{\textbf{Applying the representation theorem to PMMR}} \label{apprendix: representer_MMR}

First, we quote the representation theorem.

\begin{thm}
Consider a positive-definite real-valued kernel $k: \mathcal{X} \times \mathcal{X} \rightarrow \mathbb{R}$ on a non-empty set $\mathcal{X}$ with a corresponding reproducing kernel Hilbert space $H_{k}$. 
Let there be given
\begin{itemize}
    \item a training sample $\left(x_{1}, y_{1}\right), \ldots,\left(x_{n}, y_{n}\right) \in \mathcal{X} \times \mathbb{R}$,
    \item a strictly increasing real-valued function $g:[0, \infty) \rightarrow \mathbb{R}$, and
    \item an arbitrary error function $E:\left(\mathcal{X} \times \mathbb{R}^{2}\right)^{n} \rightarrow \mathbb{R} \cup\{\infty\}$,
\end{itemize}
which together define the following regularized empirical risk functional on $H_{k}$ :
$$
f \mapsto E\left(\left(x_{1}, y_{1}, f\left(x_{1}\right)\right), \ldots,\left(x_{n}, y_{n}, f\left(x_{n}\right)\right)\right)+g(\|f\|)
$$
Then, any minimizer of the empirical risk
$$
f^{*}=\operatorname{argmin}_{f \in H_{k}}\left\{E\left(\left(x_{1}, y_{1}, f\left(x_{1}\right)\right), \ldots,\left(x_{n}, y_{n}, f\left(x_{n}\right)\right)\right)+g(\|f\|)\right\}, \quad(*)
$$
admits a representation of the form:
$$
f^{*}(\cdot)=\sum_{i=1}^{n} \alpha_{i} k\left(\cdot, x_{i}\right)
$$
where $\alpha_{i} \in \mathbb{R}$ for all $1 \leq i \leq n$.
\end{thm}

In our case, we have
\begin{eqnarray*}
    E: (\mathcal{X} \times \mathbb{R}^2)^n &\rightarrow& \mathbb{R} \cup \{\infty\}\\
    \{((a_i,w_i,x_i,z_i),h(a_i,w_i,x_i), y_i)\}_{i=1}^n &\mapsto& \sum_{i, j=1}^{n} \frac{\left(y_{i}-h\left(a_{i}, w_{i}, x_{i}\right)\right)\left(y_{j}-h\left(a_{j}, w_{j}, x_{j}\right)\right) k\left(\left(a_{i}, z_{i}, x_{i}\right),\left(a_{j}, z_{j}, x_{j}\right)\right)}{n^{2}}
\end{eqnarray*}
so the representer theorem gives us $h(a, w, x)=\sum_{i=1}^{n} \alpha_i l\left(\left(a_{i}, w_{i}, x_{i}\right),(a, w, x)\right)$.


\subsection{\textbf{PMMR Algorithm}}
PMMR algorithm to estimate $h$ and derive causal effect is summarized below:
\IncMargin{1em}
\begin{algorithm}[h]\label{Alg:MMR_alg}\caption{Algorithm 1. PMMR}
\SetKwInOut{Input}{input}\SetKwInOut{Output}{output}
\Input{1. Train data $\{z_i^t, w_i^t, a_i^t, y_i^t, x_i^t\}_{i=1}^n$. 2. Kernel functions $l$ for $\mathcal{H}_{\mathcal{AWX}}$ and $k$ for $\mathcal{H}_{\mathcal{AZX}}$ with bandwidths $\sigma_{k}$ and $\sigma_l$ respectively. 3. Regularisation parameter $\lambda$. 4. Nystr{\"o}m approximation size $M$.}
\Output{$\hat{h}(a,w,x)$} 
\tcc{Write $\mathbf{x}$ for the matrix containing $x_i$ in the $i$th row.}

For all $1\leq i \leq n$, $1 \leq j \leq n$, $K_{ij} \gets k((a_i, z_i, x_i), (a_j, z_j, x_j))$\;

For all $1\leq i \leq n$, $1 \leq j \leq n$, $L_{ij} \gets l((a_i, w_i, x_i), (a_j, w_j, x_j))$\;

Do Nystr{\"o}m approximation for $K/{n^2}$, decomposing into $K/{n^2} = \Tilde{U}\Tilde{V}\Tilde{U}^T$ \;

$\hat{\alpha} \gets \lambda^{-1}[I - \Tilde{U}(\lambda^{-1} \Tilde{U}^T L \Tilde{U} + \Tilde{V}^{-1})^{-1} \Tilde{U}^T \lambda^{-1}L] \Tilde{U} \Tilde{V} \Tilde{U}^T \mathbf{y}$\;

$\hat{h}(a,w,x) \gets l((a,w,x), (\mathbf{a}^t, \mathbf{w}^t, \mathbf{x}^t))$\;

\caption{PMMR Algorithm}
\end{algorithm}

\subsection{Consistency and Convergence Rates}


In this section, we provide a consistency result of the causal estimate as well as the convergence rate of the PMMR solution.
For this, we will need the consistency result of the kernel mean embedding.


\begin{lemma}[Proposition A.1, \citet{Tolstikhin17}] \label{lemma: tolstikhin}
In the following, the authors present a general result whose special cases  establishes the convergence rate of $n^{-1/2}$ for $\|\mu_k(P_n) - \mu_k(P)\|_{\mathcal{F}}$ when $\mathcal{F} = \mathcal{H}_k$ and $\mathcal{F} = L^2(\mathcal{R}^d)$. They denote $P_n(X) := \frac{1}{n}\sum_{i=1}^n \delta_{X_i}$.

Let $(X_i)_{i=1}^n$ be random samples drawn $i.i.d.$ from $P$ defined on a separable topological space $\mathcal{X}$. Suppose $g: \mathcal{X} \rightarrow H$ is continuous and 
\begin{equation}
    \sup_{x \in \mathcal{X}} \|g(x)\|_{H}^2 < C_k < \infty
\end{equation}
where $H$ is a separable Hilbert space of real-valued functions. 
Then, for any $0 < \delta \leq 1$ with probability at least $1- \delta$ we have
\begin{equation}
    \norm{\int_{\X} g(x) dP_n(x) - \int_{\X} g(x) dP(x)}_{H} \leq \sqrt{\frac{C_k}{n}} + \sqrt{\frac{2C_k \log (1/\delta)}{n}} = r(n,\delta).
\end{equation}
\end{lemma}
\subsubsection{PMMR Consistency}
\begin{definition} For clarity, we define the following variables.
\begin{itemize}
    \item $h_0(a,x,w)$ is a solution to (\ref{eq:integral-eq}).
    \item $\hat{h}_n$ is the solution of a learning algorithm with sample size $n$.
    \item $\beta(a) := \E_{\mathit{WX}}[h_0(a,W, X)] = \E[Y|do(a)]$.
    \item $\hat{\beta}_n(a) := \E_{\mathit{WX}}[\hat{h}_n(a,W, X)]$.
    \item $\hat{\beta}_n^m(a) := \frac{1}{m}\sum_{i=1}^m \hat{h}_n(a, w_i, x_i)$ with $\{w_i, x_i\}_{i=1}^m \sim_{i.i.d.} \mathcal{P}_{\mathcal{WX}}$. 
    \item $\hat{\beta}(a)$: where $m$ and $n$ are clear in context, we abuse the notation to write $\hat{\beta}(a)$ to denote the estimator for $\beta(a)$ from our algorithm.
    \item $\mu_X$ denotes the kernel mean embedding of a random variable $X$.
    \item $\hat{\mu}^m_X$ denotes the empirical estimate of $\mu_X$, given by $\hat{\mu}^m_X = \frac{1}{m}\sum_{i=1}^m [\phi(x_{i} \sim \mathcal{P}_{X})]$. Where clear from context, we omit the superscript $m$, and just use $\hat{\mu}_X$ to denote finite-sample estimator of $\mu_X$.
\end{itemize}
\end{definition}
\begin{lemma}[Causal consistency] \label{lemma3mmr}
If $\hat{h}_n \xrightarrow{P} h_0$, then $\hat{\beta}_n^m (a) \xrightarrow{P} \beta(a) $ as $m,n \rightarrow \infty$.
\end{lemma}
\begin{proof}
For brevity, in the proof that follows, we write $\mu:=\mu_{XW}$ and $\hat{\mu}^m := \hat{\mu}^m_{XW}$.

Since $\beta(a) = \langle h, \mu(\mathcal{P}_{\mathcal{W}\mathcal{X}}) \otimes \phi(a) \rangle$ and $\hat{\beta}^m_n(a) = \langle \hat{h}_{n}, \hat{\mu}^m(\mathcal{P}_{\mathcal{W}\mathcal{X}}) \otimes \phi(a) \rangle$, we can verify that
\begin{align}
    \hat{\beta}_n^m(a) - \beta(a) 
    &= \langle \hat{h}_n, \phi(a) \otimes \hat{\mu}^m\rangle - \langle h_0, \phi(a) \otimes \mu \rangle\\
    &= \langle \hat{h}_n, \phi(a) \otimes \hat{\mu}^m\rangle - \langle \hat{h}_n, \phi(a) \otimes \mu\rangle + \langle \hat{h}_n , \phi(a) \otimes \mu \rangle - \langle h_0, \phi(a) \otimes \mu \rangle \\
    &= \langle \hat{h}_n, \phi(a) \otimes (\hat{\mu}^m - \mu)\rangle + \langle \hat{h}_n - h_0, \phi(a) \otimes \mu \rangle 
\end{align}

Thus, by the Cauchy-Schwartz inequality, we have for all $a$,

\begin{equation}
    |\hat{\beta}_n^m(a) - \beta(a)| \leq \|\hat{h}\|_{\mathcal{H}_{\mathcal{A} \mathcal{W} \mathcal{X}}}\|\phi(a)\|_{\Ha}\|\hat{\mu}^m - \mu\|_{\mathcal{H}_{\A\W}} + \|\hat{h}_n - h_0\|_{\mathcal{H}_{\mathcal{A}\mathcal{W} \mathcal{X}}}\|\phi(a)\|_{\Ha}\|\mu\|_{\mathcal{H}_{\A \W}} \label{eq: causal_consistency_cauchy}  
\end{equation}

From Lemma \ref{lemma: tolstikhin}, by setting $g$ to be the feature map on $\A \times \W$, we have
\begin{equation}
    \norm{\hat{\mu}^m - \mu} \leq \sqrt{\frac{C_k}{m}} + \sqrt{\frac{2C_k \log(1/\delta)}{m}} \eqdef r(m, \delta)
\end{equation}
with probability at least $1-\delta$.

Moreover, we know that $\hat{h}_n \xrightarrow{P} h_0$. This is to say, for any $\epsilon, \delta$, $\exists N$ s.t.
\begin{equation}
    \|\hat{h}_n - h_0\|_{\mathcal{H}_{\mathcal{A} \mathcal{W}\mathcal{X}}} \leq \epsilon, \hspace{0.5cm} \forall n \geq N
\end{equation}
with probability at least $1-\delta$.

Therefore, reflecting on (\ref{eq: causal_consistency_cauchy}) we observe that $\|\mu\|_{\mathcal{H}_{\A\W}}$ is bounded because we assume bounded kernels,  $\|\hat{h}\|_{\mathcal{H}_{\mathcal{A} \mathcal{W} \mathcal{X}}} \xrightarrow{P} \|h_0\|_{\mathcal{H}_{\mathcal{A}\mathcal{W} \mathcal{X}}}$ by assumption, and $\|h_0\|_{\mathcal{H}_{\mathcal{A} \mathcal{W}\mathcal{X}}}$ is constant, $\|\hat{h}_n - h_0\|_{\mathcal{H}_{\mathcal{A} \mathcal{W}\mathcal{X}}}$ and  $\|\hat{\mu}^m - \mu\|_{\mathcal{H}_{\A \W}}$ uniformly converge to zero in probability, which we have just shown. 

Therefore, $\sup_{a \in \mathcal{A}} \;\{\hat{\beta}_n^m (a) -  \beta(a)\} \xrightarrow{P} 0$.
\end{proof}

\begin{lemma} \label{lemma4mmr}
Suppose $R_k$ has at least one minimiser in $\mathcal{H_{AWX}}$ and $\mathcal{P_{AWX}}$ is a finite Borel measure with full support, i.e., $\mathrm{supp}[\mathcal{P_{AWX}}] = \mathcal{A \cross W \cross X}$. 
Then, $R_k$ has a unique minimiser in $\mathcal{H_{AWX}}$ if and only if the following condition holds:

$(\ast)$ \hspace{1cm}  $\forall g \in \mathcal{H_{AWX}}$, $\mathbb{E}_{\mathit{AWX}}[g(A, W, X)|A,Z,X] = 0$  $\mathcal{P_{AZX}}$-almost surely if and only if $g(a, w, x) = 0$ $\mathcal{P_{AWX}}-$almost surely. 
\end{lemma}

\begin{proof}
To prove that $R_k$ has a unique minimiser in $\mathcal{H_{AXW}}$, we need i) A minimiser to $R_k$ exists in $\mathcal{H_{AWX}}$ ii) It is unique. By Assumption \ref{ass:completeness_2} and \citet[Appendix, Conditions (v)-(vii)]{miao2018identifying}, a minimiser exists in $\ltwoawx$ - we further require $R_k$ has a minimiser in $\mathcal{H_{AWX}}$ by assumption.
We still need to show uniqueness. 

($\implies$) Suppose that there exist two different functions $h_1$ and $h_2$ that minimise $R_k$. 
Then, it follows from \citet[Theorem 1]{zhang2020maximum} that $\mathbb{E}[Y -h_1 | A,Z,X] = \mathbb{E}[Y - h_2|A, Z,X] = 0$ $\mathcal{P_{AZX}}-$almost surely. 
This means that $\mathbb{E}_{\mathit{AWX}}[h_1(A,W,X) - h_2(A,W,X)|A,Z,X] = 0 $ $\mathcal{P_{AZX}}-$almost surely. 
Suppose ($\ast$) is true, we must have $g(a,w,x)=h_1(a,w,x) - h_2(a,w,x) = 0$ $\mathcal{P_{AWX}}-$almost surely.
As a result, $g$ is the zero function in $\ltwoawx$ and for any other functions $f \in \ltwoawx$, $\langle g, f \rangle_{\ltwoawx} = 0$.

Now, we describe briefly the integral operator representation of our kernel function $l$ and consequently a representation of the RKHS inner product. A more detailed discussion can be found in, e.g., \citet{rkhs_theory}. 

\textbf{Integral operator of kernel on $\mathcal{H_{AWX}}$.} Let $l: (\mathcal{A \cross W \cross X})^2 \rightarrow \mathbb{R}$ be the kernel function on $\mathcal{A \cross W \cross X}$. We define an operator $S_l: \ltwoawx \rightarrow \mathcal{C}(\mathcal{A \cross W \cross X})$, where $\mathcal{C}(\mathcal{A \cross W \cross X})$ is the space of continuous functions on $\mathcal{A \cross W \cross X}$, as
\begin{equation}
    (S_l f)((a,w,x)) = \int l((a,w,x), (a',w',x')) f((a',w',x')) \,d\mathcal{P_{AWX}}((a',w',x')), \quad f\in \ltwoawx 
\end{equation}
\noindent where $S_l$ can be shown to be well-defined \citep{rkhs_theory}, and $T_l = I_l \circ S_l$ its composition with the inclusion $I_l: \mathcal{C(A \cross W \cross X)} \xhookrightarrow{} \ltwoawx$. $T_l$ is said to be the integral operator of kernel $l$. 

It can be shown that the symmetry of $l$ implies the integral operator is self-adjoint; the positive definiteness of $l$ implies that $T_l$ is a positive operator, i.e., all eigenvalues are non-negative; continuity of $l$ imples $T_l$ is compact by the Arzela-Ascoli theorem.
Then, by the Spectral theorem \citep[Theorem 49]{rkhs_theory}, any compact, self-adjoint operator can be diagonalised in an appropriate orthonormal basis.

\textbf{Relating the RKHS norm with the $\mathcal{L}^2-$norm.} Further supposing that $\mathcal{P_{AWX}}$ has full support, i.e., $\mathrm{supp}[\mathcal{P_{AWX}}] = \mathcal{A \cross W \cross X}$, then Mercer's theorem says that for a continuous kernel $l$ on a compact metric space with a finite Borel measure of full support, we can decompose the kernel function $l$ using its at most countable set $J$ of \textit{strictly positive} eigenvalues $\{\lambda_j\}_{j \in J}$ and eigenfunctions $\{e_j\}_{j \in J}$. 

\begin{equation}
    l((a,w,x), (a',w',x')) = \sum_{j \in J} \lambda_j e_j((a,w,x)) e_j((a',w',x')) 
\end{equation}
where the convergence is uniform on $\mathcal{(A \cross W \cross X)}^2$ and absolute on each $(a,w,x), (a',w',x') \in \mathcal{A \cross W \cross X}$. See \citet[Section 6.2]{rkhs_theory} for further details.

Then, we may construct the RKHS $\mathcal{H_{AWX}}$ based on the integral operator $T_l$ and its associated eigenfunctions $\{e_j\}_{j\in J}$, which depend on the underlying measure $\mathcal{P_{AZX}}$, as
\begin{equation}
    \mathcal{H_{A W X}} = \left\{f = \sum_{j \in J} a_j e_j \hspace{0.5cm} \left\{ \frac{a_j}{\sqrt{\lambda_j}} \right\} \in l^2(J) \right\}
\end{equation}
\noindent with an inner product $\langle \sum_{j \in J} a_j e_j, \sum_{j\in J} b_j e_j\rangle_{\mathcal{H_{AWX}}} = \sum_{j \in J} \frac{a_j b_j}{\lambda_j}$. 
Note that $\left\{ a_j/\sqrt{\lambda_j} \right\} \in l^2(J)$ implies that $f \in \ltwoawx$.
Thus, $a_j = \langle f, e_j \rangle_{\ltwoawx}$. 

Now, recall that $g(a,w,x) = h_1(a,w,x) - h_2(a,w,x)$ is a zero function in $\ltwoawx$, which also means that $\|g\|_{\mathcal{H_{AWX}}} = \sqrt{\langle g, g\rangle}_{\mathcal{H_{AWX}}} = 0$. 
Therefore, $h_1(a,w,x)=h_2(a,w,x)$ for all $(a, w, x)\in\mathcal{A}\times\mathcal{W}\times\mathcal{X}$ as norm convergence in RKHS implies pointwise convergence \citep[pp. 119]{steinwart2008support}.
By contradiction, the minimizer of $R_k$ must be unique.

($\impliedby$) Suppose ($\ast$) does not hold, i.e., $(A,Z,X)$ is not complete for $(A, W, X)$, then there exists $g \in \mathcal{H_{AWX}}$ such that $g \neq 0$ and $\E_{\mathit{AWX}}[g(A,W,X)|A,Z,X] = 0$, $\mathcal{P_{AZX}}-$almost surely. Then, for any minimizer $h$ of $R_k$ (if it exists), $h + Cg$ for some constant $C$ is also a minimizer, so it cannot be unique.
\end{proof}

\begin{thm}[Causal consistency of PMMR] \label{thm: mmr_consistency}
Assume $\mathcal{H}$ is a real-RKHS, $k: (\A \times \Z \times \X)^2 \rightarrow \R$ is bounded, $\Omega{(h)}$ is convex, $\lambda \xrightarrow{P} 0$. 
Moreover, assume $Z$ is complete for $W$, i.e., for all $g \in \mathcal{L}^2[\mathcal{P}_{\mathcal{W}}]$, $\E[g(W)|Z] = 0$, $\mathcal{P}_{\mathcal{Z}}-$almost surely if and only if $g(W) = 0$, $\mathcal{P}_{\mathcal{W}}-$almost surely. 
Then, $\hat{h}_n \xrightarrow{P} h_0$.
\end{thm}
\begin{proof}

Given $\Omega(h)$ is convex in $h$, we prove consistency based on \citet[Theorem 2.7]{NEWEY19942111}, which requires (i) $R_k(h)$ is uniquely minimized at $h_0$; (ii) $\hat{R}_V(h) + \lambda \Omega(h)$ is convex; (iii) $\hat{R}_V(h) + \lambda \Omega(h) \xrightarrow{P} R_k(h)$ for all $h \in \mathcal{H}$.

Since $\mathcal{H}$ is a real-RKHS, which is a vector space, it is convex because for any $x, y \in \mathcal{H}$, $a \in [0,1]$, $ax + (1-a)y \in \mathcal{H}$ by closure of vector spaces. 
Since $\mathcal{H}$ is convex, $R_k$ is convex \citep[Theorem 5]{zhang2020maximum}.  
By assumption, $A,Z,X$ is complete for $W$. 
Then, by Lemma \ref{lemma4mmr}, $R_k$ is minimized at $h_0$. 
Since $\mathcal{H}$ is open, $h_0$ is in the interior of $\mathcal{H}$.

Since $\hat{R}_V(h) = \norm{\frac{1}{n} \sum_{i=1}^n (y_i - h(a_i, w_i, x_i))k((a_i, z_i, x_i), \cdot)}_{\mathcal{H}_k}^2$, by the law of large numbers, we have that $\frac{1}{n} \sum_{i=1}^n (y_i - h(a_i, w_i, x_i))k((a_i,z_i, x_i), \cdot) \xrightarrow{P} \E[(Y - h(A,W,X))k((A,Z,X), \cdot)]$. Then $\hat{R}_V(h)\xrightarrow{P} R_k(h)$ for all $h \in \mathcal{H}$ by the Continuous Mapping Theorem \cite{mann1943} since $\|\cdot\|_{\mathcal{H}_k}$ is continuous. 
As $\lambda \xrightarrow{P} 0$, $\hat{R}_V(h) + \lambda \Omega(h) \xrightarrow{P} R_k(h)$ by Slutsky's Theorem \citep[Lemma 2.8]{vandervaart2000}. 
Since $\Omega(h)$ is convex, $\hat{R}_V(h) + \lambda \Omega(h)$ is convex since addition preserves convexity.
Thus, by \citet[Theorem 2.7]{NEWEY19942111}, $\hat{h}_n \xrightarrow{P} h_0$. 
\end{proof}

\begin{corollary}
Assume $\mathcal{H}$ is a real-RKHS, $k: (\A \times \Z \times \X)^2 \rightarrow \mathbb{R}$ is bounded, $\Omega(h)$ is convex, and $\lambda \xrightarrow{P} 0$. Moreover, assume $(A, Z, X)$ is complete for $W$, then the causal effect estimate $\hat{\beta}_n^m \xrightarrow{P} 0$ as $m,n \rightarrow \infty$.
\end{corollary}
\begin{proof}
By Theorem \ref{thm: mmr_consistency}, the conditions guarantee that $\hat{h}_n \xrightarrow{P} h_0 $. 
Then, by Lemma \ref{lemma3mmr} $\hat{\beta}_n^m(A) \xrightarrow{P} \beta(A)$.
\end{proof}

\subsubsection{PMMR Convergence Rate}

To provide the convergence rate of PMMR, we will first provide an alternative interpretation of PMMR as a linear ill-posed inverse problem in the RKHS \citep{Nashed74:IntegralEq,Carrasco07:LIP}. 
Let $\phi(a,x,w) := k((a,x,w),\cdot)$ and $\varphi(a,x,z) := k((a,x,z),\cdot)$ be the canonical feature maps.
Then, the unregularized PMMR objective can be expressed as
\begin{eqnarray*}
    R_k(h) &=& \left\| \mathbb{E}[(Y - h(A,X,W))\varphi(A,X,Z)] \right\|_{\mathcal{H}_{\mathit{AXZ}}}^2 \\
    &=& \left\| \mathbb{E}[Y\varphi(A,X,Z)] - \mathbb{E}[h(A,X,W)\varphi(A,X,Z)] \right\|_{\mathcal{H}_{\mathit{AXZ}}}^2 \\
    &=& \left\| g - Th \right\|^2_{\mathcal{H}_{\mathit{AXZ}}},
\end{eqnarray*}
where
\begin{equation}\label{eq:g-T-def}
    g := \int Y\varphi(A,X,Z) \,d\rho(A,X,Y,Z), \quad
    Th := \int h(A,X,W)\varphi(A,X,Z)\, d\rho(A,X,W,Z).
\end{equation}
Here $\rho(A,X,Y,Z)$ and $\rho(A,X,W,Z)$ are the restrictions of $\rho(A,X,W,Y,Z)$ to $\mathcal{A}\times\mathcal{X}\times\mathcal{Y}\times\mathcal{Z}$ and $\mathcal{A}\times\mathcal{X}\times\mathcal{W}\times\mathcal{Z}$, respectively.
By Assumptions \ref{ass:y_bounded} and \ref{ass:kernel_characteristic}, $g\in\mathcal{H}_{\mathit{AXZ}}$ and $T$ is a bounded linear operator from $\mathcal{H}_{\mathit{AXW}}$ to $\mathcal{H}_{\mathit{AXZ}}$. Let $T^*: \mathcal{H}_{\mathit{AXZ}} \to \mathcal{H}_{\mathit{AXW}}$ be an adjoint operator of $T$ such that $\langle Tu,v\rangle_{\mathcal{H}_{\mathit{AXZ}}} = \langle u,T^*v\rangle_{\mathcal{H}_{\mathit{AXW}}}$ for all $u\in\mathcal{H}_{\mathit{AXW}}$ and $v\in\mathcal{H}_{\mathit{AXZ}}$.

Based on the above formulation, we can rewrite the PMMR regularized objective and its empirical estimate as follow:
\begin{equation}\label{eq:mmr-obj-emp}
    R_{\lambda}(h) = \| g - Th \|^2_{\mathcal{H}_{\mathit{AXZ}}} 
    + \lambda\|h\|_{\mathcal{H}_{\mathit{AXW}}}^2, \qquad 
    \widehat{R}_{\lambda}(h) = \| \hat{g} - \widehat{T}h \|^2_{\mathcal{H}_{\mathit{AXZ}}} 
    + \lambda\|h\|_{\mathcal{H}_{\mathit{AXW}}}^2,
\end{equation}
where $\hat{g}$ and $\widehat{T}$ are the empirical estimates of $g$ and $T$ based on the i.i.d. sample $(a_i,x_i,w_i,y_i,z_i)_{i=1}^n$ from $\rho(A,X,W,Y,Z)$:
\begin{equation}\label{eq:g-T-emp}
    \hat{g} := \frac{1}{n}\sum_{i=1}^n y_i\varphi(a_i,x_i,z_i), \quad
    \widehat{T}h := \frac{1}{n}\sum_{i=1}^n h(a_i,x_i,w_i)\varphi(a_i,x_i,z_i).
\end{equation}
Likewise, we denote by $\widehat{T}^*$ an adjoint operator of $\widehat{T}$, i.e., for $f\in\mathcal{H}_{\mathit{AXZ}}$,
\begin{equation}\label{eq:adjoint-T}
    T^*f := \int f(A,X,Z)\phi(A,X,W)\, d\rho(A,X,W,Z), \quad 
    \widehat{T}^*f := \frac{1}{n}\sum_{i=1}^n f(a_i,x_i,z_i)\phi(a_i,x_i,w_i).
\end{equation}

\paragraph{Cross-covariance operator.}
We can view the operator $T$ as an element of the product RKHS $\mathcal{H}_{\mathit{AXW}}\otimes\mathcal{H}_{\mathit{AXZ}}$, i.e., for $h\in\mathcal{H}_{\mathit{AXW}}$,
\begin{eqnarray*}
Th &=& \int h(A,X,W)\varphi(A,X,Z)\, d\rho(A,X,W,Z) \\
&=& \int \langle h, \phi(A,X,W)\rangle_{\mathcal{H}_{\mathit{AXW}}}\varphi(A,X,Z)\, d\rho(A,X,W,Z) \\
&=& \int \left[\phi(A,X,W)\otimes\varphi(A,X,Z)\right]h\, d\rho(A,X,W,Z) \\
&=& \left[\int \phi(A,X,W)\otimes\varphi(A,X,Z)\, d\rho(A,X,W,Z)\right]h.
\end{eqnarray*}
Thus, $T = \mathbb{E}[\phi(A,X,W)\otimes\varphi(A,X,Z)] \in \mathcal{H}_{\mathit{AXW}}\otimes\mathcal{H}_{\mathit{AXZ}}$ and is a (uncentered) \emph{cross-covariance} operator mapping from $\mathcal{H}_{\mathit{AXW}}$ to $\mathcal{H}_{\mathit{AXZ}}$ \citep{Bak73:CC,Fukumizu04:DRS}.
Likewise, $T^* = \mathbb{E}[\varphi(A,X,Z)\otimes\phi(A,X,W)] \in \mathcal{H}_{\mathit{AXZ}}\otimes\mathcal{H}_{\mathit{AXW}}$.
The cross-covariance operator $T$ is Hilbert-Schmidt, and $\|T\| \leq \|T\|_{\text{HS}} = \|T\|_{\mathcal{H}_{\mathit{AXW}}\otimes\mathcal{H}_{\mathit{AXZ}}}$ where $\|\cdot\|_{\text{HS}}$ denotes a Hilbert-Schmidt norm \citep[Lemma 3]{Fukumizu06:KCCA}.
The latter equality holds because the space of Hilbert-Schmidt operators $\text{HS}(\mathcal{H}_1,\mathcal{H}_2)$ forms Hilbert space which are isomorphic to the product space $\mathcal{H}_1\otimes\mathcal{H}_2$ given by the product kernel.

\paragraph{PMMR solutions.} Based on \eqref{eq:mmr-obj-emp}, we can define the PMMR solutions in the population limit and in the finite sample regime respectively as
\begin{eqnarray}
    h_{\lambda} &:=& {\arg\min}_{h\in\mathcal{H}_{\mathit{AXW}}}\; R_{\lambda}(h) = (T^*T + \lambda I)^{-1}T^*g \label{eq:h_pop} \\
    \hat{h}_{\lambda} &:=& {\arg\min}_{h\in\mathcal{H}_{\mathit{AXW}}}\; \widehat{R}_{\lambda}(h) = (\widehat{T}^*\widehat{T} + \lambda I)^{-1}\widehat{T}^*\hat{g} \label{eq:h_emp}
\end{eqnarray}
The solution \eqref{eq:h_pop} is obtained by noting that $R_{\lambda}(h) = \langle h, T^*Th + \lambda h - 2T^*g \rangle_{\mathcal{H}_{\mathit{AXW}}} + \|g\|^2_{\mathcal{H}_{\mathit{AXZ}}}$ whose Frechet derivative is zero only if $(T^*T + \lambda I)h = T^*g$.
The solution in \eqref{eq:h_emp} can be obtained in a similar way.
Let $h_0$ be the solution that uniquely minimizes the unregularized risk $R(h)$.
Then, we can decompose the estimation bias into two parts:
\begin{equation}\label{eq:mmr-decomposition}
\hat{h}_{\lambda} - h_0 = (\hat{h}_{\lambda} - h_{\lambda}) + (h_{\lambda} - h_0).
\end{equation}
The first part $\hat{h}_{\lambda} - h_{\lambda}$ corresponds to an estimation error of the regularized solution $h_{\lambda}$, whereas the second part $h_{\lambda} - h_0$ is the regularization bias.
Hence, we can obtain the convergence rate of $\hat{h}_{\lambda}$ by first characterizing the rates of the regularization bias and estimation error separately, and then choosing the regularization parameter $\lambda$ such that both rates coincide.


\subsubsection*{Characterizing the Regularization Bias}

To control the regularization bias, we impose a regularity condition on the true unknown $h_0$.
Following \citet{Carrasco07:LIP}, we assume that $h_0$ belong to a regularity space $H_\gamma = (T^*T)^\gamma$ for some positive $\gamma$. 
The following is a restatement of \citet[Def. 3.4]{Carrasco07:LIP}; see, also \citet{smale2007} for a similar condition.

\begin{definition}[$\gamma$-regularity space]
The $\gamma$-regularity space of the compact operator $T$ is defined for all $\gamma > 0$, as the RKHS associated with $(T^*T)^\gamma$. That is,
\begin{equation}
    H_\gamma = \left\{ h \in \mathcal{N}(T)^{\perp} \quad \text{such that}\quad \sum_{j=1}^\infty \frac{\langle h, \phi_j\rangle}{\alpha_j^{2\gamma}} < \infty\right\}
\end{equation}
with the inner product
\begin{equation}
    \langle f,g \rangle_\gamma = \sum_{j=1}^\infty \frac{\langle f, \phi_j\rangle\langle g, \phi_j\rangle}{\alpha_j^{2\gamma}}
\end{equation}
for $f,g \in H_\gamma$.
\end{definition}

In what follows, we will make the following assumption on $h_0$.
\begin{assumption}\label{ass:mmr-regularity}
    $h_0 \in H_{\gamma}$ for $\gamma \in (0,2]$.
\end{assumption}

\begin{proposition}[Regularization bias]\label{prop:mmr-reg-bias}
Let $T:\mathcal{H}_{\mathit{AXW}}\to\mathcal{H}_{\mathit{AXZ}}$ be an injective compact operator. Then, if Assumption \ref{ass:mmr-regularity} holds and $h_\lambda$ is defined by \eqref{eq:h_pop}, we have
\begin{equation}
\|h_{\lambda} - h_0\|^2_{\mathcal{H}_{\mathit{AXW}}} = \mathcal{O}(\lambda^{\min(\gamma,2)}).
\end{equation}
\end{proposition}

\begin{proof}
\citet[Proposition 3.12]{Carrasco07:LIP}
\end{proof}

\subsubsection*{Characterizing the Estimation Error}

\begin{proposition}[Estimation error]\label{prop:mmr-estimation-error}
Let $h_{\lambda} = (T^*T + \lambda I)^{-1}T^*g$ be the regularized solution given by \eqref{eq:h_pop} and $\hat{h}_{\lambda} = (\widehat{T}^*\widehat{T} + \lambda I)^{-1}\widehat{T}^*\hat{g}$, then
\begin{equation*}
    \|\hat{h}_{\lambda} - h_{\lambda}\|_{\mathcal{H}_{\mathit{AXW}}} \leq d(\lambda)\|\widehat{T}^*\hat{g} - \widehat{T}^*\widehat{T}h_0\| + d(\lambda)\|\widehat{T}^*\widehat{T} - T^*T \|\left\|h_0 - h_{\lambda}\right\|_{\mathcal{H}_{\mathit{AXW}}}.
\end{equation*}
where $d(\lambda) := \|\widehat{\Gamma}_{\lambda}\|=\| (\widehat{T}^*\widehat{T} + \lambda I)^{-1} \|$.
\end{proposition}

\begin{proof}
To simplify the notation, we will use $\Gamma_\lambda := (T^*T + \lambda I)^{-1}$ and $\widehat{\Gamma}_\lambda := (\widehat{T}^*\widehat{T} + \lambda I)^{-1}$ throughout the proof. 
First, we have
\begin{equation}\label{eq:mmr-intermediate-1}
    \hat{h}_{\lambda} - h_{\lambda} = \widehat{\Gamma}_{\lambda}\widehat{T}^*\hat{g} - \Gamma_{\lambda} T^*g
    = \widehat{\Gamma}_{\lambda}\widehat{T}^*\left(\hat{g} - \widehat{T}h_0\right) + \underbrace{\widehat{\Gamma}_{\lambda}\widehat{T}^*\widehat{T}h_0  - \Gamma_{\lambda}T^*Th_0}_{(\star)}.
\end{equation}
Then, we can write $(\star)$ as
\begin{eqnarray}\label{eq:mmr-intermediate-2}
    \widehat{\Gamma}_{\lambda}\widehat{T}^*\widehat{T}h_0  - \Gamma_{\lambda}T^*Th_0 
    &=&  \widehat{\Gamma}_{\lambda}(\widehat{T}^*\widehat{T} - T^*T )h_0  + \widehat{\Gamma}_{\lambda}T^*Th_0
    + \Gamma_{\lambda}T^*Th_0 \nonumber \\
    &=&  \widehat{\Gamma}_{\lambda}(\widehat{T}^*\widehat{T} - T^*T )h_0 
    + (\widehat{\Gamma}_{\lambda} - \Gamma_{\lambda})T^*Th_0 \nonumber \\
    &\stackrel{(a)}{=}& \widehat{\Gamma}_{\lambda}(\widehat{T}^*\widehat{T} - T^*T)h_0 
    + \widehat{\Gamma}_{\lambda}( T^*T - \widehat{T}^*\widehat{T})\Gamma_{\lambda}T^*Th_0 \nonumber \\
    &\stackrel{(b)}{=}& \widehat{\Gamma}_{\lambda}(\widehat{T}^*\widehat{T} - T^*T )h_0 
    + \widehat{\Gamma}_{\lambda}( T^*T - \widehat{T}^*\widehat{T})h_{\lambda} \nonumber \\
    &=& \widehat{\Gamma}_{\lambda}(\widehat{T}^*\widehat{T} - T^*T )(h_0 - h_{\lambda}),
\end{eqnarray}
where we applied the identity $A^{-1} - B^{-1} = A^{-1}(B - A)B^{-1}$ to $\widehat{\Gamma}_{\lambda} - \Gamma_{\lambda}$ to get $(a)$, and $(b)$ holds because $h_{\lambda} = \Gamma_{\lambda}T^*Th_0$.
Combining \eqref{eq:mmr-intermediate-1} and \eqref{eq:mmr-intermediate-2} yields
\begin{eqnarray*}
    \hat{h}_{\lambda} - h_{\lambda} &=& \widehat{\Gamma}_{\lambda}\widehat{T}^*(\hat{g} - \widehat{T}h_0) + \widehat{\Gamma}_{\lambda}(\widehat{T}^*\widehat{T} - T^*T)(h_0 - h_{\lambda}).
\end{eqnarray*}
Consequently, we have
\begin{eqnarray*}
    \|\hat{h}_{\lambda} - h_{\lambda}\|_{\mathcal{H}_{\mathit{AXW}}} &\leq& d(\lambda)\|\widehat{T}^*\hat{g} - \widehat{T}^*\widehat{T}h_0\| + d(\lambda)\|\widehat{T}^*\widehat{T} - T^*T \|\left\|h_0 - h_{\lambda}\right\|_{\mathcal{H}_{\mathit{AXW}}},
\end{eqnarray*}
where $d(\lambda) := \|\widehat{\Gamma}_{\lambda}\| =\| (\widehat{T}^*\widehat{T} + \lambda I)^{-1} \|$ as required.
\end{proof}

By Proposition \ref{prop:mmr-reg-bias}, Proposition \ref{prop:mmr-estimation-error}, and \eqref{eq:mmr-decomposition}, we can see that the rate of convergence of the estimation bias $\|\hat{h}_{\lambda} - h_0\|$ depends on the following quantities:
(i) A sequence of regularization parameters $\lambda$ which will govern the rate of convergence of the regularization bias $\|h_\lambda - h_0\|$.
(ii) The rate of convergence to infinity of $d(\lambda)$.
(iii) The rates of convergence of $\|\widehat{T}^*\widehat{T} - T^*T \|$ and $\|\widehat{T}^*\hat{g} - \widehat{T}^*\widehat{T}h_0\|$ which are governed by the estimation of $T$ and $g$.
In the next section, we provide the rates for these intermediate quantities.

\subsubsection*{Rates of Intermediate Quantities}

Since we will deal with random variables taking values in Hilbert spaces, we need the following concentration inequality.

\begin{lemma}[Bennett inequality in Hilbert space]\label{lem:bennett}
Let $\mathcal{H}$ be a Hilbert space and $\xi$ be a random variable with values in $\mathcal{H}$. Assume that $\|\xi\| \leq M < \infty$ almost surely. Denote $\sigma^2(\xi) = \mathbb{E}[\|\xi\|^2]$. Let $\{\xi_i\}_{i=1}^n$ be independent random drawers of a random variable $\xi$. Then, with probability at least $1-\delta$,
\begin{equation*}
    \left\|\frac{1}{n}\sum_{i=1}^n[\xi_i - \mathbb{E}[\xi_i]]\right\| \leq \frac{2M\log(2/\delta)}{n} + \sqrt{\frac{2\sigma^2(\xi)\log(2/\delta)}{n}}.
\end{equation*}
\end{lemma}

\begin{lemma}[Consistency of $\hat{g}$, $\widehat{T}$, and $\widehat{T}^*$]\label{lem:A1}
Suppose that Assumptions \ref{ass:y_bounded} and \ref{ass:kernel_characteristic} holds. Let $\sigma_g^2$ and $\sigma^2_T$ be defined by
\begin{equation*}
    \sigma^2_g := \mathbb{E}[\|Y\varphi(A,X,Z)\|^2], \quad 
    \sigma^2_T := \mathbb{E}[\|\phi(A,X,W)\|^2\|\varphi(A,X,Z)\|^2].
\end{equation*}
Then, each of the following statements holds true with probability at least $1-\delta$:
\begin{eqnarray*}
     \|\hat{g} - g\|  &\leq& \frac{2c_Y\kappa^3\log(2/\delta)}{n} + \sqrt{\frac{2\sigma^2_g\log(2/\delta)}{n}} \\
     \|\widehat{T} - T\| &\leq& \frac{2\kappa^6\log(2/\delta)}{n} + \sqrt{\frac{2\sigma^2_T\log(2/\delta)}{n}} \\
     \|\widehat{T}^* - T^*\| &\leq& \frac{2\kappa^6\log(2/\delta)}{n} + \sqrt{\frac{2\sigma^2_T\log(2/\delta)}{n}}
\end{eqnarray*}
\end{lemma}

\begin{proof}
Let $\xi_g(a,x,y,z) := y\varphi(a,x,z)$. It follows from Assumptions \ref{ass:y_bounded} and \ref{ass:kernel_characteristic} that
\begin{equation*}
    \|\xi_g(a,x,y,z) \| \leq |y|\|\varphi(a,x,z)\| = |y|\sqrt{k(a,a)k(x,x)k(z,z)} \leq c_Y\kappa^3.
\end{equation*}
Hence, we have 
\begin{equation*}
    \hat{g} = \frac{1}{n}\sum_{i=1}^n\xi_g(a_i,x_i,y_i,z_i), \quad 
    g = \mathbb{E}[\xi_g(A,X,Y,Z)].
\end{equation*}
If $\sigma^2_g = \mathbb{E}[\|\xi\|^2] = \mathbb{E}[\|Y\varphi(A,X,Z)\|^2]$, it follows from Lemma \ref{lem:bennett} that
\begin{equation*}
    \|\hat{g} - g\|  \leq \frac{2c_Y\kappa^3\log(2/\delta)}{n} + \sqrt{\frac{2\sigma^2_g\log(2/\delta)}{n}}
\end{equation*}
with probability at least $1-\delta$.
Next, to bound $\|\widehat{T} - T\|$, recall that we can express $\widehat{T}$ and $T$ as elements of $\mathcal{H}_{\mathit{AXW}}\otimes\mathcal{H}_{\mathit{AXZ}}$ as follows:
\begin{equation*}
    \widehat{T} = \frac{1}{n}\sum_{i=1}^n\phi(a_i,x_i,w_i)\otimes\varphi(a_i,x_i,z_i), \quad 
    T = \int \phi(A,X,W)\otimes\varphi(A,X,Z)\,d\rho(A,X,W,Z).
\end{equation*}
Let $\xi_T(a,x,w,z) := \phi(a,x,w)\otimes\varphi(a,x,z)\in\mathcal{H}_{\mathit{AXW}}\otimes\mathcal{H}_{\mathit{AXZ}}$. Then, by Assumption \ref{ass:kernel_characteristic},
\begin{equation}
    \|\xi_T(a,x,w,z)\| = \|\phi(a,x,w)\|\|\varphi(a,x,z)\| 
    \leq \sqrt{k(a,a)k(x,x)k(w,w)}\sqrt{k(a,a)k(x,x)k(z,z)} \leq \kappa^6.
\end{equation}
As a result, we can express $\widehat{T}$ and $T$ as
\begin{equation}
    \widehat{T} = \frac{1}{n}\sum_{i=1}^n\xi_T(a_i,x_i,w_i,z_i), \quad 
    T = \mathbb{E}[\xi_T(A,X,W,Z)].
\end{equation}
Letting $\sigma^2_T := \mathbb{E}[\|\xi_T\|^2] = \mathbb{E}[\|\phi(A,X,W)\|^2\|\varphi(A,X,Z)\|^2]$ and applying Lemma \ref{lem:bennett} yields with probability at least $1-\delta$
\begin{equation}
    \|\widehat{T} - T\| \leq \|\widehat{T} - T\|_{\mathcal{H}_{\mathit{AXW}}\otimes\mathcal{H}_{\mathit{AXZ}}} \leq \frac{2\kappa^6\log(2/\delta)}{n} + \sqrt{\frac{2\sigma^2_T\log(2/\delta)}{n}}.
\end{equation}
The bound on $\|\widehat{T}^* - T^*\|$ can be obtained using similar proof techniques, so we omit it for brevity.
\end{proof}

\begin{lemma}\label{lem:A2}
$\|\widehat{T}^*\widehat{T} - T^*T\| = \mathcal{O}(1/\sqrt{n})$.
\end{lemma}

\begin{proof}
First, we have
\begin{eqnarray*}
    \|\widehat{T}^*\widehat{T} - T^*T\| &=& \|\widehat{T}^*\widehat{T} - \widehat{T}^*T + \widehat{T}^*T - T^*T\| \\
    &\leq& \|\widehat{T}^*\widehat{T} - \widehat{T}^*T\| + \|\widehat{T}^*T - T^*T\| \\
    &=& \|\widehat{T}^*(\widehat{T} - T)\| + \|(\widehat{T}^* - T^*)T\|\\
    &\le&\|(\widehat{T}^*-T^*)(\widehat{T} - T)\| + \|T^*(\widehat{T} - T)\|+ \|(\widehat{T}^* - T^*)T\|\\
    &\le& \|\widehat{T} - T\|^2 + 2\|T\| \|\widehat{T} - T\|.
\end{eqnarray*}
Hence, the rate of convergence of $\|\widehat{T}^*\widehat{T} - T^*T\|$ is dominated by the rate of $\|\widehat{T} - T\|$ which, according to Lemma \ref{lem:A1}, is in the order of $\mathcal{O}(1/\sqrt{n})$.
\end{proof}

\begin{lemma}\label{lem:A3}
$\|\widehat{T}^*\hat{g} - \widehat{T}^*\widehat{T}h_0 \| = \mathcal{O}(1/\sqrt{n})$.
\end{lemma}

\begin{proof}
First, we have
\begin{eqnarray*}
    \|\widehat{T}^*\hat{g} - \widehat{T}^*\widehat{T}h_0 \|_{\mathcal{H}_{\mathit{AXW}}} &=& \|(\widehat{T}^*\hat{g} - T^*Th_0) + (T^*Th_0 - \widehat{T}^*\widehat{T}h_0) \|_{\mathcal{H}_{\mathit{AXW}}} \\
    &=& \|(\widehat{T}^*\hat{g} - T^*g) + (T^*Th_0 - \widehat{T}^*\widehat{T}h_0) \|_{\mathcal{H}_{\mathit{AXW}}} \\
    &=& \|(\widehat{T}^*\hat{g} - \widehat{T}^*g) + (\widehat{T}^*g - T^*g) + (T^*Th_0 - \widehat{T}^*\widehat{T}h_0) \|_{\mathcal{H}_{\mathit{AXW}}} \\
    &\leq& \underbrace{\|\widehat{T}^*\hat{g} - \widehat{T}^*g\|_{\mathcal{H}_{\mathit{AXW}}}}_{(A)} + \underbrace{\|\widehat{T}^*g - T^*g\|_{\mathcal{H}_{\mathit{AXW}}}}_{(B)} + \underbrace{\|T^*Th_0 - \widehat{T}^*\widehat{T}h_0 \|_{\mathcal{H}_{\mathit{AXW}}}}_{(C)}.
\end{eqnarray*}
Next, we will bound each term separately.

\paragraph{Probabilistic bound on $(A)$.}
Since $\widehat{T}^*$ is a Hilbert-Schmidt operator in $\mathcal{H}_{\mathit{AXZ}}\otimes\mathcal{H}_{\mathit{AXW}}$, we have by Assumption \ref{ass:kernel_characteristic} that $\|\widehat{T}^*\| \leq \|\widehat{T}^*\|_{\text{HS}} \leq \kappa^3$. Consequently, 
$\|\widehat{T}^*\hat{g} - \widehat{T}^*g\|_{\mathcal{H}_{\mathit{AXW}}} = \|\widehat{T}^*(\hat{g} - g)\|_{\mathcal{H}_{\mathit{AXW}}} \leq \|\widehat{T}^*\|\|\hat{g} - g\|_{\mathcal{H}_{\mathit{AXZ}}} \leq \kappa^3\|\hat{g} - g\|_{\mathcal{H}_{\mathit{AXZ}}}$.
By Lamma \ref{lem:A1}, we have with probability at least $1-\delta$,
\begin{equation}
    (A) \leq \frac{2c_Y\log(2/\delta)}{n} + \frac{1}{\kappa^3}\sqrt{\frac{2\sigma^2_g\log(2/\delta)}{n}}.
\end{equation}
That is, $(A) = \mathcal{O}(1/\sqrt{n})$.

\paragraph{Probabilistic bound on $(B)$.} 
Using Lemma \ref{lem:A1}, we have $\|\widehat{T}^*g - T^*g\| \leq \|\widehat{T}^*-T^*\|\|g\|_{\mathcal{H}_{\mathit{AXZ}}} = \mathcal{O}(1/\sqrt{n})$. 

\paragraph{Probabilistic bound on $(C)$.} 
$\|T^*Th_0 - \widehat{T}^*\widehat{T}h_0 \|_{\mathcal{H}_{\mathit{AXW}}} \leq \|T^*T - \widehat{T}^*\widehat{T}\|\|h_0 \|_{\mathcal{H}_{\mathit{AXW}}} = \mathcal{O}(1/\sqrt{n})$ by Lemma \ref{lem:A2}.

Since $(A)$, $(B)$, and $(C)$ are all in the order of $\mathcal{O}(1/\sqrt{n})$, $\|\widehat{T}^*\hat{g} - \widehat{T}^*\widehat{T}h_0 \| = \mathcal{O}(1/\sqrt{n})$ as required.
\end{proof}

\paragraph{Probabilistic bound on $\|\widehat{\Gamma}\|$.}

Assume $\lambda \le \|T^*T\|$ and $n\ge 2C_{\epsilon}\kappa\mathcal{N}(\lambda)\lambda^{-1}$. Then, with probability at least $1-\epsilon/3$, $\| \widehat{\Gamma}\|\le 1/\lambda$.

\begin{proof}
Assume
\begin{equation}\label{eq:one_half}
    \|(T^*T - \widehat{T}^*\widehat{T})(T^* T + \lambda I)^{-1}\|\le \frac{1}{2}.
\end{equation}
Using the Neumann series of $I-(T^* T -\widehat{T}^*\widehat{T}  )(T^*T+\lambda)^{-1}$, we have
\begin{align*}
(\widehat{T}^*\widehat{T} + \lambda I)^{-1} &= (T^* T + \lambda I)^{-1}(I-(T^*T - \widehat{T}^*\widehat{T})(T^* T + \lambda I)^{-1})^{-1}\\
&=(T^* T + \lambda I)^{-1} \sum_{k=0}^{\infty} ((T^*T - \widehat{T}^*\widehat{T})(T^* T + \lambda I)^{-1})^k.
\end{align*}
Hence,
\begin{align*}
\|(\widehat{T}^*\widehat{T} + \lambda I)^{-1}\| 
&=\|(T^* T + \lambda I)^{-1}\| \sum_{k=0}^{\infty} \|(T^*T - \widehat{T}^*\widehat{T})(T^* T + \lambda I)^{-1}\|^k\\
&\le \|(T^* T + \lambda I)^{-1}\| \frac{1}{1-\|(T^*T - \widehat{T}^*\widehat{T})(T^* T + \lambda I)^{-1}\|}\\
&\le 2\|(T^* T + \lambda I)^{-1}\|.
\end{align*}
where the last inequality results from \eqref{eq:one_half}. 
On the other hand, by the spectral theorem,
\begin{equation*}
    \|(T^* T + \lambda I)^{-1}\|=\sup_{l \in (l_k)_{k=0}^{\infty}}\frac{1}{l+\lambda} \le \frac{1}{\lambda},
\end{equation*}
where $(l_k)_{k=0}^{\infty}$ are the eigenvalues of $T^*T$.
We now prove \eqref{eq:one_half}. We have
\begin{equation*}
    \|(T^*T - \widehat{T}^*\widehat{T})(T^* T + \lambda I)^{-1}\|\le  \|(T^*T - \widehat{T}^*\widehat{T})\| \|(T^* T + \lambda I)^{-1}\|\le \frac{\| (T^*T - \widehat{T}^*\widehat{T})\|}{\lambda}
\end{equation*}
The last term is smaller than $\mathcal{O}(1/\sqrt{n})$ with high probability.
\end{proof}
\subsubsection*{Final Step}

We have shown that $\|\widehat{T}^*\widehat{T} - T^*T\| = \mathcal{O}(1/\sqrt{n})$ and $\|\widehat{T}^*\hat{g} - \widehat{T}^*\widehat{T}h_0 \| = \mathcal{O}(1/\sqrt{n})$, and are now in a position to provide the rate of convergence of the estimation bias $\|\hat{h}_{\lambda} - h_0\|$.

\subsection{Proof of Theorem \ref{thm:pmmr_final_thm}}
\textbf{Theorem statement.}
Suppose that $h_0\in\mathcal{H}_\gamma$ for some $\gamma > 0$ and the conditions of Lemma \ref{lem:A1}, \ref{lem:A2}, and \ref{lem:A3} hold. 
If $n^{\frac{1}{2}-\frac{1}{2}\min\left(\frac{2}{\gamma+2},\frac{1}{2}\right)}$ is bounded away from zero, and  
$\lambda = n^{-\frac{1}{2}\min\left(\frac{2}{\gamma+2},\frac{1}{2}\right)}$,
then
\begin{equation}
    \| \hat{h}_{\lambda} - h_0 \| = \mathcal{O}\left(n^{-\frac{1}{2}\min\left(\frac{\gamma}{\gamma+2},\frac{1}{2}\right)}\right).
\end{equation}

\begin{proof}
Suppose that $\|\widehat{T}^*\widehat{T} - T^*T\| = \mathcal{O}(1/\alpha_n)$ and $\|\widehat{T}^*\hat{g} - \widehat{T}^*\widehat{T}h_0 \| = \mathcal{O}(1/\beta_n)$. 
Then, it follows from Proposition \ref{prop:mmr-estimation-error} and \citet[Proposition 4.1]{Carrasco07:LIP} that 
\begin{equation}
    \|\hat{h}_{\lambda} - h_0\| = \mathcal{O}\left(\frac{1}{\lambda\beta_n} + \left(\frac{1}{\lambda\alpha_n} + 1\right)\|h_\lambda - h_0\|\right).
\end{equation}
Hence, $\lambda\beta_n$ must go to infinity as least as fast as $\|h_\lambda - h_0\|^{-1}$. That is, for $h_0\in\mathcal{H}_\gamma$, Proposition \ref{prop:mmr-reg-bias} implies that
\begin{equation}
    \lambda^2\beta_n^2 \geq \lambda^{-\min(\gamma,2)} \Rightarrow \lambda \geq \beta_n^{-\max\left(\frac{2}{\gamma+2},\frac{1}{2}\right)}. 
\end{equation}
Thus, to get the fastest possible rate, we will choose $\lambda = \beta_n^{-\max\left(\frac{2}{\gamma+2},\frac{1}{2}\right)}$. Consequently, the rate of convergence of $\|\hat{h}_\lambda - h_0\|$ and $\|h_\lambda - h_0\|$ will coincide if and only if $\alpha_n\beta_n^{-\max\left(\frac{2}{\gamma+2},\frac{1}{2}\right)}$ is bounded away from zero.
Finally, by Lemma \ref{lem:A2} and Lemma \ref{lem:A3}, we substitute $\alpha_n = \sqrt{n}$ and $\beta_n = \sqrt{n}$ to get the stated result.
\end{proof}

\textbf{Proof of Proposition \ref{prop:rate_pmmr_ce}}
\begin{proof}
We can adapt Lemma \ref{lemma3mmr} easily to see that, setting $m=n_t$ and for simplicity of notation writing $\hat{\beta} = \hat{\beta}_{n}^(n_t)$,
\begin{align}
    |\hat{\beta}(a) - \beta(a)| &\leq \|\hat{h}_{\lambda}\|_{\mathcal{H}_{\mathcal{A} \mathcal{W} \mathcal{X}}}\|\phi(a)\|_{\Ha}\|\hat{\mu}^{n_t} - \mu\|_{\mathcal{H}_{\A\W}} + \|\hat{h}_{\lambda} - h_0\|_{\mathcal{H}_{\mathcal{A}\mathcal{W} \mathcal{X}}}\|\phi(a)\|_{\Ha}\|\mu\|_{\mathcal{H}_{\A \W}} \nonumber\\
    &= \mathcal{O}(\|\hat{\mu}^{n_t} - \mu\|_{\mathcal{H}_{\A\W}}) + \mathcal{O}(\|\hat{h}_{\lambda} - h_0\|_{\mathcal{H}_{\mathcal{A}\mathcal{W} \mathcal{X}}}) \label{eq: final_causal_consistency}  
\end{align}

From Lemma \ref{lemma: tolstikhin}, by setting $g$ to be the feature map on $\A \times \W$, we have
\begin{equation}
    \norm{\hat{\mu}^{n_t} - \mu} \leq \sqrt{\frac{C_k}{n_t}} + \sqrt{\frac{2C_k \log(1/\delta)}{n_t}} = \mathcal{O}(n_t^{-\frac{1}{2}})
\end{equation}

By Theorem \ref{thm:pmmr_final_thm} we have 
\begin{equation}
    \| \hat{h}_{\lambda} - h_0 \| = \mathcal{O}\left(n^{-\frac{1}{2}\min\left(\frac{\gamma}{\gamma+2},\frac{1}{2}\right)}\right).
\end{equation}

Thus, collecting rates of both terms in \eqref{eq: final_causal_consistency} we get 
\begin{equation}
    |\hat{\beta}(a) - \beta(a)| = \mathcal{O}(n_t^{-\frac{1}{2}} + n^{-\frac{1}{2}\min\left(\frac{\gamma}{\gamma+2},\frac{1}{2}\right)})
\end{equation}

\end{proof}

\section{Experiments}\label{app:D-experiments}
\subsection{Data}

\begin{figure*}
    \centering
    {{\includegraphics[width=.6\textwidth]{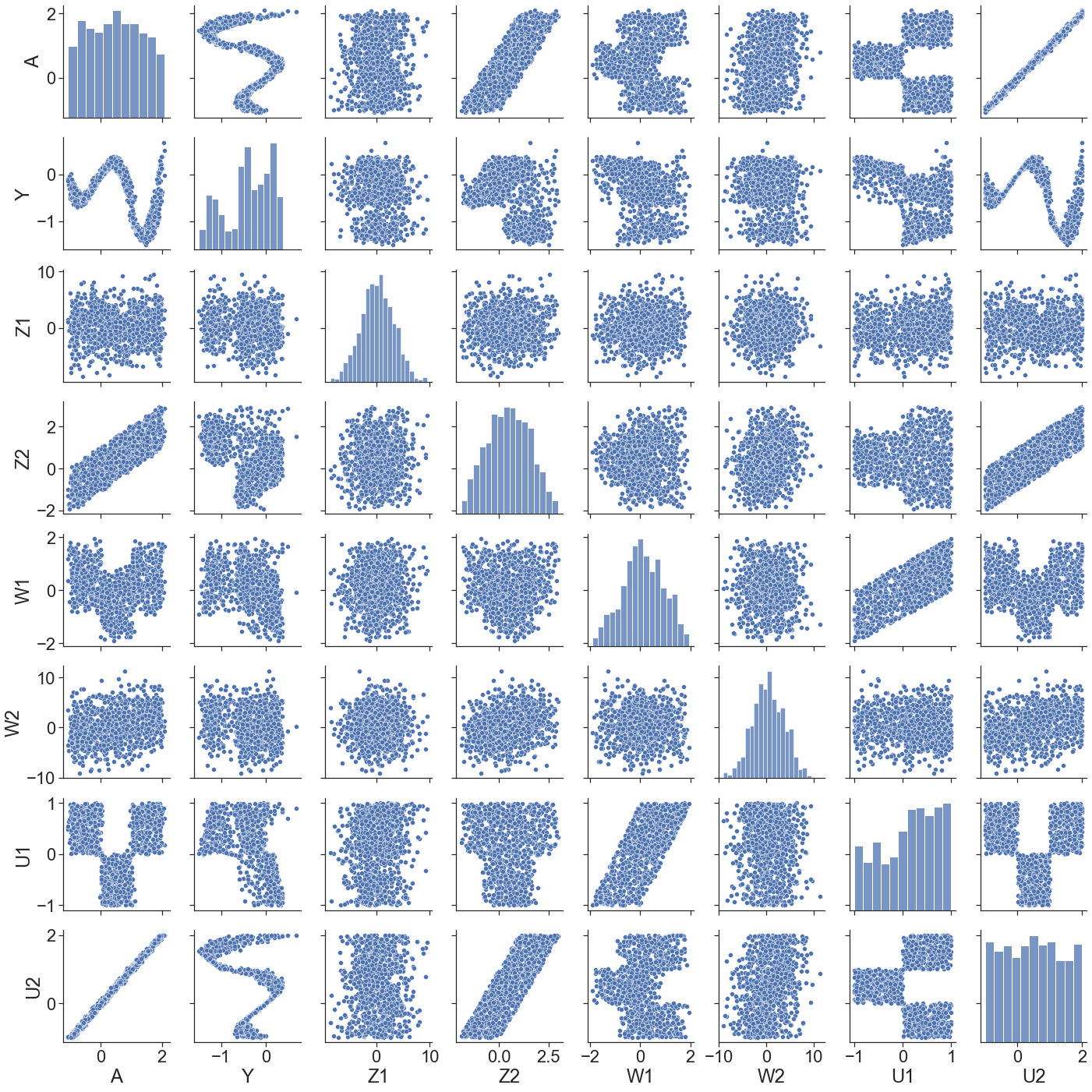}}}%
    \caption{Synthetic generative model, sample size=1000}
    \label{fig:synthetic_generative}%
\end{figure*}

\subsubsection{Real world data}
\textbf{Disclaimer: We have applied our proposed methodologies on real world datasets to demonstrate performance of our methods. The results should only be interpreted within framework of assessing methodologies.}

For the Abortion and Criminality data \cite{Woody2020EstimatingHE}, the treatment variable is \textit{effective abortion rate}, the outcome variable is \textit{murder rate}, and the covariates are
\textit{prisoner population per capita}, \textit{state unemployment rate}, \textit{income per capita}, \textit{state poverty rate},  \textit{beer consumption per capita}, \textit{presence of concealed weapons law}, \textit{police employment rate per capita}, and \textit{generosity to Aid to Families with Dependent Children}. How we selected the proxy variables are described below. We mask the rest of the variables as unobserved confounders.

For the Education case study \cite{deaner2021proxy, fruehwirth2016timing}, we are interested in the effect of grade retention on long-term cognitive outcome, measured in terms of a reading and maths score when the subject is aged around 11. In particular, our treatment variables are discrete, with levels at $0$ (no retention), $1$ (kindergarten retention) and $2$ (early elementary school retention). Following \cite{deaner2021proxy}, we use as proxy variables Kindergarten test scores ($W$) and early or late elementary school test scores ($Z$). Like in the Abortion and Criminality data, we mask the rest of the variables as unobserved confounders.

To construct the \emph{True Average Causal Effect} for real world datasets, where we do not have access to the full generative model to infer $\E(y|do(a))$, we have developed an empirical model to learn the \emph{latent variable} for each dataset. Specifically, we followed the procedure below to model the latent confounder. 
\begin{enumerate}
    \item We identified the potential candidates for proxies $W$ and $Z$ by stratifying variables based on the domain knowledge and correlation with $y$ and $a$.
    For Criminology case study ("Legalized abortion and crime"), we followed  \cite{Woody2020EstimatingHE} to identify proxy variables and categorize them as $W$ and $Z$. 
    For the Educational case study ("Grade retention and Cognitive outcome"), we selected proxies as proposed by  \cite{deaner2021proxy}. By this, we constructed a multi-dimensional proxy variables $W$ and $Z$ for each example. 
    \item We have included all other covariates as common endogenous confounders in generative model, i.e. $X$.  
    
    In Criminology case, as proposed by \cite{Woody2020EstimatingHE}, we added a set of exogenous common confounders to the model. In contrast with  endogenous confounders, the common latent confounder ($U$) is not a parent of the exogenous confounders/covariates. 
    \item Assuming a generative model consistent with graph in \cref{fig:overall}, we learned parameters of this generative model from data. Specifically, we assumed a graph $\mathcal{G}$ consistent with \cref{fig:overall} and learned the \emph{Structural causal model (SCM)}, \(E(V_i|pa(V_i))=f_i(pa(V_i)),\quad \forall i \in \mathcal{G}\), for each endogenous variable. We fit a \emph{ generalized additive model} for each experiment to learn parameters of the generative model.
    \item  To learn the generative distribution of the unmeasured confounder, we fit a \emph{Gaussian Mixture Model} on noise term of SCMs, learned at the previous stage. That is, we assumed the latent confounder $U$ (multidimensional confoudner unaccounted for in previous step) manifest as correlated noises of SCMs. We learned the parameters of a Gaussian Mixture model representing this latent variable.
    \item We proceed to generate samples \(\{(a,x,z,w,y)_i\}^n_{i=1}\) from the generative model for \(\mathcal{G}\) learnt in previous steps (n=10000).
    \item The True Average Causal Effect at a given $A=a$ is estimated by fixing $A$ at $a$ and averaging the $Y$ samples sampled from the fixed $A$ and the rest of its parents.
\end{enumerate} 

\subsection{Hyperparameters selection}
For both KPV and PMMR, we employ Gaussian kernel \eqref{eq:kernel} for continuous variables, as it is a continuous, bounded, and characteristic kernel and meets all assumptions required to guarantee consistency of the solution at population level. 
\begin{equation}
    k_{x_i,x_j}=exp\{-\frac{\| x_i-x_j \|^2}{2\sigma^2}\}\label{eq:kernel}
\end{equation}
See \cite{sriperumbudur2011universality} for survey of properties of these kernels. For multidimensional inputs, we use the product of scalar kernels for each  dimension as the kernel of the input. 
In both  KPV and PMMR settings, we deal with two categories of hyper-parameters: (1) Kernel's length-scale ($\sigma$), and (2) regularization hyper-parameters.

\subsubsection{\textbf{Hyperparameter selection procedure (KPV).}}\label{sub:D21}
\textit{Kernel's length-scale}. A convenient heuristic is to set the length-scale equal to the median inter-point distances of all points in sample with size $n$. that is, \(\sigma:=Med (\vert x_i-x_j\vert_{\cH})\quad \forall\, i,j \in n\). 
We initiated the length-scale hyperparameter according to this heuristic for every input (and every dimension of multidimensional inputs). We, subsequently, chose the optimal length-scale from a narrow range around this level to allow for narrower/wider kernels to be considered.

\textit{Regularization hyper-parameters}. For the regularization parameters, for both Stage 1 and Stage 2, we use the leave-one-out cross validation method and follow the procedure proposed in \citep[Algorithm.  H1]{singh2020kernel} to find the optimal regularization hyper-parameter. In particular, we constructed $H_\lambda$ and  $\Tilde{H}_\lambda$ for Stage 1 as:\[H_{\lambda_1} = I- \mathcal{K}_{AXZ} (\mathcal{K}_{AXZ}+m_1\lambda_1)^{-1}, \qquad \Tilde{H}_{\lambda_1} = diag(H_{\lambda_1}), \quad \mathcal{K}_{AXZ}:=K_{AA} \odot K_{XX} \odot K_{ZZ}\]
and implemented a grid search over $\Lambda_1$ to find $\lambda_1$ as a minimizer of the closed form of validation loss \eqref{eq:stage1_reg}. 
\begin{eqnarray}\label{eq:stage1_reg}
\hat{\lambda}_1= \argmin_{\lambda_1 \in \Lambda_1} \frac{1}{m_1}\vert \Tilde{H}^{-1}_{\lambda_1} H_{\lambda_1} K_{WW} H_{\lambda_1} \Tilde{H}^{-1}_{\lambda_1}\vert_2,\quad \Lambda \in \R 
\end{eqnarray}
For Stage 2:\[H_{\lambda_2} = I- A \left( m_2\lambda_2 + \Sigma \right)^{-1}, \qquad \Tilde{H}_{\lambda_2} = diag(H_{\lambda_2})\]
where $A:=\Gamma_{(\tA,\tX,\tZ)} \overline{\otimes} I_{m_2\times m_2}$ and $\Sigma$ is defined as \eqref{eq:sigma}. We implemented a grid search over $\Lambda_2$ to find $\lambda_2$ as a minimizer of the closed form of validation loss \eqref{eq:stage2_reg}.
\begin{eqnarray}\label{eq:stage2_reg}
\hat{\lambda}_2= \argmin_{\lambda_2 \in \Lambda_2} \frac{1}{m_2}\vert \Tilde{H}^{-1}_{\lambda_2} H_{\lambda_2} y\vert^2_2,\quad \Lambda_2 \in \R 
\end{eqnarray}

Note that in our setting, we assumed that the optimal hyperparameters of the first and second stages can be selected independently. In reality, however, the hyperparameter selected in first stage, has a direct effect on second stage loss and consequently, the optimal value of the hyperparameter in second stage. 
\subsubsection{\textbf{Hyperparameter selection procedure (PMMR).}} 

\textit{Kernel's length-scale}. We select $\sigma_{l}$ and $\sigma_{k}$ using the median interdistance heuristic on the joint kernels $l: (A \times X \times W)^2 \rightarrow \mathbb{R}$ and $k: (A \times X \times Z)^2 \rightarrow \mathbb{R}$. 

\textit{Regularization hyper-parameters}. For the regularization parameter $\lambda$, we let $b_{l}^2 = (\lambda n^2)^{-1} \implies \lambda = \frac{1}{(b_ln)^2}$. For all training sizes $n$, we fixed the range of $b_ln$ to be $[2, 450]$, which translate to a range in $\lambda$ of $[4.9 \times 10^{-6}, 0.25]$, and we do grid search with a grid size of $50$. 

The metric we use for hyperparameter selection is the empirical estimate of the $V-statistic$, that is, $\widehat{R}_V$. We select the hyperparameter $\lambda$ which minimizes $\hat{R}_V$ over a held-out validation set.

\subsection{Results}
\subsubsection{Abortion \& Criminality}
The unobserved confounding variables ($U$) are selected as "income per capita", "police employment rate per capita", "state unemployment rate" and "state poverty rate"; the outcome inducing proxies ($W$) are selected as "prisoner population per capita", "prescence of concealed weapons law", "beer consumption per capita". We calculate their Canonical Correlation, obtaining an absolute correlation value $(|r_{CCA}|)$ of $0.48$, suggesting strong correlation between $W$ and $U$.


\section{A Connection between Two-stage Procedure and Maximum Moment Restrictions for the Proxy Setting}\label{sec:connection_proof}
Note that $R$ and $\tilde{R}$, true loss for PMMR and KPV methods, respectively, are both positive quantities.

\begin{lemma}
  A minimizer of $\tilde{R}$ is a minimizer of $R$; and  vice-versa. This minimize is unique. 
\end{lemma}
\begin{proof}
  For any $h, h' \in L^2_{P_{AXW}}$, by developing the squares and  using the law of iterated expectation, we have :
\begin{align*}
    \tilde{R}(h) - \tilde{R}(h') &= \mathbb{E}_{AXYZ}[(Y - \mathbb{E}[h(A,X,W)\,|\,A,X,Z])^2] - \mathbb{E}_{AXYZ}[(Y - \mathbb{E}[h'((A,X,W)\,|\,A,X,Z])^2]\\
    &= 2 \E_{AXYZ}[ Y\E[h'(A,X,W)-h(A,X,W)|A,X,Z ]] + \E_{AXZ}[ \E[h(A,X,W)|A,X,Z]^2] \\
    &\qquad    - \E_{AXZ}[ \E[h'(A,X,W)|A,X,Z]^2]\\
    & = 2 \E_{AXZ}[ \E[Y|A,X,Z]\E[h'(A,X,W)-h(A,X,W)|A,X,Z ]] \\
    & \qquad + \E_{AXZ}[ \E[h(A,X,W)|A,X,Z]^2] - \E_{AXZ}[ \E[h'(A,X,W)|A,X,Z]^2]\\
    &= R(h) - R(h').
\end{align*}

Assuming $\exists h,h'\in L^2_{P_{A,X,W}}$ such that $\E[Y|A,X,Z]=\E[h(A,X,W)|A,X,Z]$, according to the preceding computations we have:
\begin{align*}
    \tilde{R}(h) - \tilde{R}(h')
    &= R(h) - R(h') \\
     &= 2 \E_{AXZ}[ \E[Y|A,X,Z]\E[h'(A,X,W)-h(A,X,W)|A,X,Z ]] \\
     &\qquad +\E_{AXZ}[ \E[h(A,X,W)|A,X,Z]^2] - \E_{AXZ}[ \E[h'(A,X,W)|A,X,Z]^2]\\
    &= \E_{AXZ}[ \E[h'(A,X,W)|A,X,Z]^2] - 2\E_{AXZ}[ \E[h(A,X,W)|A,X,Z]\E[h'(A,X,W)|A,X,Z]]\\
    &\qquad +\E_{AXZ}[ \E[h(A,X,W)|A,X,Z]^2]\\
    &= \E_{AXZ}[ (\E[h'(A,X,W)|A,X,Z]- \E[h(A,X,W)|A,X,Z])^2].
\end{align*}
Taking $h'=h$ in the equation above shows that $h$ is a minimizer of $R$ and $\tilde{R}$ Hence, a unique minimizer of $R$ is a minimizer of $\tilde{R}$; and  vice-versa.
\end{proof}
2. By \Cref{lem:proxy-mmr}, $R(h)=0$ if and only if $h$ satisfies the conditional moment restriction (CMR): $ \mathbb{E}[Y - h(A,W,X)\,|\,A, Z, X] = 0$, $\mathbb{P}(A,Z,X)$-almost surely. We now show $R_k(h)=0$ if and only if $R(h)=0$. Firstly, by the law of iterated expectations,
\begin{align*}
    \mathbb{E}[(Y-h(A,W,X))k((A,Z,X), \cdot)]&=\E_{A,X,Z}[\E[(Y-h(A,W,X))k((A,Z,X),\cdot)|A,X,Z]]\\
    &=\E_{A,X,Z}[\E[(Y-h(A,W,X))|A,X,Z]k((A,Z,X),\cdot)].
\end{align*}
By \Cref{lem:pmmr_closed_form}, 
$R_k(h) = \left\|\mathbb{E}[(Y-h(A,W,X))k((A,Z,X), \cdot)] \right\|^2_{\mathcal{H}_{\mathcal{AZX}}}.$ Hence, if $h$ satisfies the CMR condition, then $R_k(h)=0$. We now assume that $R_k(h)=0$. We can write $R_k(h)$ as:
\begin{equation*}
    \iint g(a,x,z)k((a,x,z),(a',x',z')g(a',x',z')d(a,x,z)d(a,x,z)=0,
\end{equation*}
where we define $g(a,x,z)=\E_{AXWY}[Y-h(A,X,W)|a,x,z]d\rho(a,x,z)$. Since $k$ is ISPD by \Cref{ass:ispd}, this implies the CMR: $\mathbb{E}[Y - h(A,W,X)\,|\,A, Z, X] = 0$, $\mathbb{P}(A,Z,X)$-almost surely.

3. In KPV, the method is decomposed in two stages.

\textit{First stage.} Under the assumption that $\E[f(w)|A,X,Z=\cdot]$ is in $\cH_{\A\X\Z}$ for any $f \in \cH_{W}$, the conditional mean embedding  $\mu$ can be written $\mu_{W|a,x,z}=C_{W|A,X,Z} \phi(a,x,z)$ for any $(a,x,z)\in \A\times\X\times\Z$, where $C_{W|A,X,Z}:\cH_{\A\X\Z} \to \cH_{\W}$ is the conditional mean embedding operator is well-defined \cite{song2009hilbert}. Let $\cH_{\Gamma}$ the vector-valued RKHS of operators from  $\cH_{\A\X\Z}$ to $\Hw$. A crucial result is that the tensor product  $\cH_{\A\X\Z}\otimes \Hw$ is isomorphic to $\mathcal{L}^2(\cH_{\A\X\Z},\Hw)$ the space of Hilbert-Schmidt operators from  $\cH_{\A\X\Z}$ to $\Hw$. Hence, by choosing the vector-valued kernel $\Gamma$ with feature map : $(w,a,x,z)\mapsto [ \phia(a)\otimes \phix(x)\otimes \phiz(z)\otimes \phiw(w)]=\phia(a)\otimes \phix(x)\otimes \phiz(z) \ps{\phiw(w),\cdot}_{\Hw}$, we have $\cH_{\Gamma}=\mathcal{L}^2(\cH_{\A\X\Z},\Hw)$ and they share the same norm. We denote by $L^2(\A\times\X\times\Z,\rho_{\A\X\Z})$ the space of square integrable functions from $\A\times\X\times \Z$ to $\W$ with respect to measure $\rho_{\A\X\Z}$, where $\rho_{\A\X\Z}$ is the restriction of $\rho$ to $\A\times\X\times\Z$. Assuming $C_{W|A,X,Z} \in \cH_{\Gamma}$, it is the solution to the following risk minimization:
\begin{equation}\label{eq:kpv_1}
    C_{W|A,X,Z} = \argmin_{c \in \mathcal{H}_{\Gamma} } E(C) \quad \text{ where }\quad E(C)=\E_{AXZW}\left[\| \phi(W) - C \phi(a,x,z) \|^2_{\Hw}  \right]
\end{equation}

\textit{Second stage.} Under the assumptions of a characteristic kernel and that $h_0 \in \mathcal{H}_{AW}$, $\E[h(A,X,W)|A,X,Z]=\eta_{AXW}[\phia(a,x)\otimes\mu_{W|a,x,z}]$. The operator $\eta_{AXW}$ minimizes 
\begin{equation*}
    \eta_{AXW} = \argmin_{\eta \in \mathcal{H}_{\A\X\W}}\tilde{R}(\eta) \quad \text{ where } \quad \tilde{R}(\eta) =\E_{AXYZ}\left[ (Y - \eta_{AXW}[\phia(a,x)\otimes\mu_{W|a,x,z}])^2\right],
\end{equation*}
where $\mu_{W|a,x,z}=C_{W|A,X,Z} \phi(a,x,z)$ and $C_{W|A,X,Z}$ is the solution of \eqref{eq:kpv_1}. Hence, as long as the problem is well-posed, i.e $C_{W|A,X,Z}\in \cH_{\Gamma}$ and $h\in \cH_{\A\X\W}$, the KPV approach recovers $\E[h(A,X,W)|\cdot]$, with $\E[h(A,X,W)|A,X,Z]=\eta_{AXW}[\phia(a,x)\otimes\mu_{W|A,X,Z}]=\eta_{AXW}[\phia(a,x)\otimes C_{W|A,X,Z} \phi(A,X,Z)]$.

\end{document}